\documentclass[final,5p,times,twocolumn]{elsarticle}

\usepackage{caption}
\usepackage{subcaption}
\usepackage{algorithm}
\usepackage{physics}
\usepackage{booktabs}
\usepackage{tabularx}
\usepackage{makecell}
\usepackage{cellspace}
\usepackage{array} 
\usepackage{float}
\usepackage{xcolor}

\usepackage{amssymb}

\usepackage{pifont}
\newcommand{\cmark}{\ding{51}}%
\newcommand{\xmark}{\ding{55}}

\journal{arXiv}

\usepackage{filecontents}

\begin{document}
\begin{frontmatter}



\title{Federated Learning: A Cutting-Edge Survey of the Latest Advancements and Applications}


\author[a1]{Azim Akhtarshenas}
\author[a6]{Mohammad Ali Vahedifar}
\author[a2]{Navid Ayoobi}
\author[a3]{Behrouz Maham}
\author[a4]{Tohid Alizadeh}
\author[a5]{Sina Ebrahimi}
\author[a1]{David López-Pérez}

\affiliation[a1]{organization={Department of Telecommunication Engineering, Universitat Politecnica de Valencia},
            city={Valencia},
            country={Spain}}
\affiliation[a6]{organization={Department of Electrical and Computer Engineering, University of Tehran},
            city={Tehran},
            country={Iran}}

\affiliation[a2]{organization={Department of Computer Science, University of Houston},
            city={Houston},
            state={TX},
            country={USA}}

\affiliation[a3]{organization={Department of Electrical and Computer Engineering, Nazarbayev University},
            city={Astana},
            country={Kazakhstan}}

\affiliation[a4]{organization={Department of Robotics and Mechatronics, Nazarbayev University},
            city={Astana},
            country={Kazakhstan}}

\affiliation[a5]{organization={Centre for Future Transport and Cities (CFTC), Coventry University},
            city={Coventry},
            country={UK}}

\begin{abstract}
Robust machine learning (ML) models can be developed by leveraging large volumes of data and distributing the computational tasks across numerous devices or servers. Federated learning (FL) is a technique in the realm of ML that facilitates this goal by utilizing cloud infrastructure to enable collaborative model training among a network of decentralized devices. Beyond distributing the computational load, FL targets the resolution of privacy issues and the reduction of communication costs simultaneously. To protect user privacy, FL requires users to send model updates rather than transmitting large quantities of raw and potentially confidential data. Specifically, individuals train ML models locally using their own data and then upload the results in the form of weights and gradients to the cloud for aggregation into the global model. This strategy is also advantageous in environments with limited bandwidth or high communication costs, as it prevents the transmission of large data volumes. With the increasing volume of data and rising privacy concerns, alongside the emergence of large-scale ML models like Large Language Models (LLMs), FL presents itself as a timely and relevant solution. It is therefore essential to review current FL algorithms to guide future research that meets the rapidly evolving ML demands. This survey provides a comprehensive analysis and comparison of the most recent FL algorithms, evaluating them on various fronts including mathematical frameworks, privacy protection, resource allocation, and applications. Beyond summarizing existing FL methods, this survey identifies potential gaps, open areas, and future challenges based on the performance reports and algorithms used in recent studies. This survey enables researchers to readily identify existing limitations in the FL field for further exploration.
\end{abstract}




\end{frontmatter}


\section{Introduction}
\label{sec1}
\begin{table*}[htb]
  \caption{LIST of frequently-used ABBREVIATIONS}
  \label{tab:1}
\centering
\begin{tabular}{ | l  l |l l| } 

\hline

\centering
 \textbf{Abbreviation} &\textbf{Definition}\hspace{0cm}& \textbf{Abbreviation}\hspace{0cm} &\textbf{Definition} \\
  \hline  
 3GPP & Third Generation Partnership Project&IIoT  &Industrial Internet of Thing \\

 ADMM &Alternate direction multiplier method &IoMT &Internet of Medical Things \\

AIoT & Artificial Internet of Thing& IoT &Internet-of-Things \\

AE &Autoencoder&IoTSE&IoT search engine \\

 CRNN &Convolutional Recurrent Neural Network &IoV& Internet of Vehicles \\

DDPG &Deep deterministic policy gradient &MARL&Multi-agent reinforcement learning \\

DL &Deep learning &MDP& Markov decision process \\

DML &Deep Mutual learning & MEC&Mobile edge computing \\

DP &Differential privacy& MINLP& Mixed-integer non-linear programming \\

 DRL&Deep reinforcement learning& QL &  Q-learning  \\

E2E &End-to-End&QoE &   Quality of experience  \\

EC &Edge computing  & QoS& Quality of service \\

EHR &Electronic Healthcare Records & RAN &   Radio access network  \\

 ES &Edge server&RL &  Reinforcement Learning\\

FedAvg &Federated averaging & SA& Secure aggregation  \\

FDRL &Federated deep reinforcement learning & SGD&   Stochastic gradient descent  \\

 FL &Federated Learning& SNR &  Signal-to-Noise Ratio  \\

GBS &Ground base station& UAV &Unmanned Aerial Vehicle  \\

HFL & Hierarchical Federated Learning & UDEC&  Ultra-dense edge computing \\

 IBC&Identity-Based Cryptosystems & UE&  User eqquipment \\

  \hline
\end{tabular}

\end{table*}

Recent years have witnessed considerable changes and enhancements in wireless mobile communication systems.
In the nascent stages of decentralized communication networks, the inherent constraints of early devices, like limited battery life, inadequate storage, and computational capacity, necessitated the introduction of centralized cloud centers within 5G wireless communication systems.
These cloud centers were designed to aggregate, store, and analyze the entirety of the data generated by these devices.
However, over the forthcoming years, the extensive proliferation of smart devices and the exponential surge in data traffic have posed substantial challenges in centralized and decentralized systems for both 5G and preceding generations of wireless communication networks.
They encountered a significant latency in efficiently processing, storing, and transmitting the immense volume of mobile data, particularly in ultra-dense networks, owing to bandwidth constraints.
In order to alleviate this latency and reduce the communication burden, edge servers (ESs) were introduced as intermediate computational entities positioned between the central server and IoT devices.
These ESs are tasked with executing computational operations and storing data in proximity to the IoT clients \cite{kar2023offloading}.

Beyond the challenge posed by the substantial volume of data, the issue of data privacy, particularly within medical domains \cite{kairouz2021advances}, faces significant threats during data communication within IoT-based infrastructures.
6G, with its exceptional features, robust capabilities, and scalability, emerges as a fortunate solution to the concerns outlined above.
It envisions a transformative evolution in network characteristics, including improved performance, higher quality of service (QoS), and enhanced quality of experience (QoE) within IoT-based mobile networks \cite{saad2019vision,you2021towards}.
The integration of Artificial Intelligence (AI) and Machine Learning (ML) into 6G networks offers enhanced network management orchestration performance by autonomously addressing optimization challenges.
As an example, AI can enhance network power efficiency by dynamically activating and deactivating components in response to real-time operational conditions, thereby eliminating human-induced errors.
Within the realm of AI model training, FL emerges as a promising technique that facilitates distributed learning.
Employing a distributed framework, FL seeks to enable localized data training on users' devices, subsequently transmitting solely model parameters to the server to safeguard against the potential exposure of sensitive raw data to untrustworthy servers \cite{venkatasubramanian2023iot}.

\subsection{AI and 6G Networks} 
AI has firmly embedded itself as an indispensable facet of home life, industrial, and academic environments.
The pervasive influence of ML algorithms and techniques is discernible in the entirety of wireless networks, extending from smart cities to remote patient monitoring and smart robots.
The incorporation of ML methods into the IoT and edge infrastructures has empowered the next generation of communication networks to exhibit ultra-reliable characteristics with reduced latency.
ML models possess the capability to analyze link Signal-to-Noise Ratio (SNR), loss rate, delay, and packet loss in 6G and subsequent generations.
Additionally, ML plays a crucial role for enhancing network efficiency, responding to the escalating network burdens caused by the proliferation of interconnected devices resulting from recent technological advances in wireless networks.
In the context of traditional cloud-based machine learning services, the development process entails the central gathering of training datasets.
Nonetheless, this training methodology confronts two primary concerns:
1) expensive communication and energy costs, and 2) compromised data privacy.
In \cite{al2111edge}, the authors presented a comprehensive explanation of the foundations and supporting technologies of FL as a solution to the aforementioned issues.
They also introduced a recently developed approach to bringing ML to edge servers.
Baccour \textit{et al.} \cite{baccour2023zero} provided a unique platform architecture that deploys a zero-touch pervasive artificial intelligence (PAI) as a service (PAIaaS) in 6G networks, leveraging a smart system founded on blockchain technolog.
Their platform is designed to standardize the integration of PAI at every architectural level and unify the interfaces, with the goal of facilitating service deployment across diverse application and infrastructure domains, alleviating user concerns regarding cost, security, and resource allocation, and concurrently meeting the rigorous performance criteria of 6G networks.
Furthermore, they introduced a federated-learning-as-a-service use case as a proof-of-concept to assess the capacity of their suggested system.
Their model exhibits self-optimization and self-adaptation, aligning itself with the dynamic behaviors of 6G networks, thereby reducing users' perceived costs.
The authors in \cite{wang2023artificial} conducted a thorough analysis of AI-assisted 6G network slicing (NS) for network assurance and service provisioning. 
Their analysis includes an exploration of promising characteristics and AI-assisted approaches concerning core network, transport network, radio access network (RAN) slicing, management systems and slice extensions.
Additionally, they suggested an elastic bandwidth scaling technique based on Reinforcement Learning (RL) that provides significant advantages in terms of increasing the request fulfillment rate and adjusting to environmental changes.

AI model is disseminated from the cloud server to edge computing (EC) nodes, wherein task nodes perform local processing and remote processing by offloading AI duties to cloud servers or other EC nodes in a 6G network.
Li \textit{et al.} \cite{li2022ai} aimed to jointly optimize the resource allocation and computation offloading choices for each node.
This optimization task is approached by solving a mixed-integer non-linear programming (MINLP) problem in order to reduce the overall computing time and energy consumption of all task nodes and increase the inference accuracy of AI tasks.
They employed an alternate direction multiplier method (ADMM)-based approach to decompose this non-convex issue into manageable MINLP sub-problems.
Their proposed ADMM-based approach allows each task node to improve its computation mode and resource allocation by using local channel state information (CSI), which aligns well with the demands of large-scale networks.
Utilizing Augmented Reality and Virtual Reality technologies, the authors in \cite{gupta2022tactile} unveiled an intelligent touch-enabled system for B5G/6G and an IoT-based wireless communication network.
The core of touch technology, enriched by the incorporation of intelligence stemming from approaches like AI, ML, and deep learning (DL), is founded upon the tactile internet and NS.
For the intelligent touch-based wireless communication system, an architectural framework is introduced, featuring a layered structure and interfaces, alongside its comprehensive end-to-end (E2E) solution.
The forthcoming 6G network is envisioned to provide a diverse range of industries with the capacity to leverage AR/VR technology in applications within robotics and healthcare facilities, aimed at addressing numerous societal issues.
Their study concluded by offering a set of use cases for the integration of touch infrastructure into automation, robotics, and intelligent healthcare systems in order to contribute to the diagnosis and treatment of widespread Covid-19 infections.

\begin{table*}[htb]
\tiny
  \caption{Comparative analysis of existing surveys with the proposed survey on FL}
  \label{CAESPSF}
  \centering
  \resizebox{\linewidth}{!}{%
  \renewcommand{\arraystretch}{1.2}
  \begin{tabular}{cccccc}
\specialrule{0.12em}{0pt}{0pt}
    \thead{Reference}& \thead{Year}&\thead{Accuracy of \\Used References}&\thead{ Mathematical Analysis of \\FL Averaging Algorithms} &\thead{Challenges and \\ Future Directions} & \thead{Mostly Updated References \\ from 2020 and Beyond} \\
\specialrule{0.12em}{0pt}{0pt}
    	  \cite{kar2023offloading} & 2023   &\xmark &\xmark&\xmark&\xmark \\
\cite{venkatasubramanian2023iot} & 2023 &\cmark&\xmark&\cmark&\cmark\\
		\cite{al2111edge}&2023 &\xmark &\xmark&\xmark&\cmark\\

		\cite{gupta2022tactile}&2023 &\xmark&\xmark&\xmark &\xmark \\

		\cite{abdulrahman2020survey}&2021 &\xmark&\xmark& \xmark&\xmark \\

		\cite{khan2021federated}&2021 &\xmark&\xmark&\xmark &\xmark\\

		\cite{arisdakessian2022survey}&2022 &\xmark&\xmark&\xmark &\xmark \\
\specialrule{0.12em}{0pt}{0pt}
\end{tabular}
}
\end{table*}

The transition from the IoT to the Internet of Vehicles (IoV) is underway. Vehicles equipped with internet connectivity have the capability to perceive, communicate, assess, and make decisions.
The extensive collection of vehicle-related data facilitates the utilization of AI and DL to deliver enhanced services for Intelligent Transportation Systems.
However, AI/DL-based ITS applications require substantial computational resources, both during the training process and model deployment.
A viable solution is exploiting the vast processing capacity that could be obtained by combining the computational power present in individual vehicles and ITS infrastructure.
In \cite{phung2021onevfc}, the authors presented the concept of a tangible vehicular fog computing platform based on OneM2M, denoted as oneVFC.
The oneVFC standard gains advantages from oneM2M by enabling interoperability and establishing hierarchical resource organization.
OneVFC coordinats information flows and computational activities on vehicle fog nodes, maintaines dispersed resources, and reports outcomes to application users.
The paper elaborates on how oneVFC efficiently manages AI-driven applications that are running on various machines within a laboratory-scale model comprising Raspberry Pi modules and laptops.
Moreover, the paper demonstrates how oneVFC excells in significantly decreasing application processing time, especially in scenarios with elevated workloads.

\subsection{Intelligent IoT-based and Edge Networks}
IoT devices and associated clients contribute to generating a substantial data influx, necessitating both storage and analytical processes.
Addressing the intricacies of large data transmission, storage, computation, and resource optimization requires developing and implementing systematic solutions.

 The authors in \cite{xing2021edge} aimed to address the challenge of how IoT devices distribute their computational tasks among EC servers and on-chip computation units to maintain a balance between energy efficiency and data privacy at the physical layer. Initially, they formulated an optimization function for IoT devices, considering factors such as energy consumption, transmission delay, and privacy requirements. Subsequently, the authors analyzed the scenario of direct transmission and determined optimal transmit power levels with and without privacy considerations. Finally, they extended the model to relay transmission scenarios, particularly when EC servers are distant, and introduced a relay selection algorithm for IoT devices.

Data processing and structure optimization face significant demands due to the billions of data bytes generated at the network edge, where traditional processing algorithms are ineffective in terms of time, cost, and computational. 
Therefore, the combination of EC and AI, resulting in the development of edge intelligence, offers a promising solution.
In pursuing this objective, Deng \textit{et al.} \cite{deng2020edge} delineated a distinction between AI on the edge and AI for the edge so called \textit{intelligence-enabled edge computing}. The former entails leveraging AI technologies to provide more optimum solutions for EC issues, and the latter delves into the comprehensive execution of AI model development, encompassing model training and inference directly at the edge.
The study \cite{esmaeili2024reinforcement} introduces a set of distributed Load Balancing (LB) algorithms that utilize machine learning to overcome limitations of the previous LB algorithm, EVBLB, such as static time intervals for execution, exhaustive server information for neighbour selection, and central coordination for request dispatching. To improve control and scalability on edge servers, three efficient algorithms are proposed: Q-learning (QL), multi-arm bandit, and gradient bandit algorithms. QL predicts execution time by integrating rewards from prior executions, enhancing performance. MAB and GB prioritize optimal neighbour servers while adapting to dynamic changes in request rates, sizes, and server resources.

The authors in \cite{chen2020privacy} introduced a searchable scheme aimed at addressing challenges within the intelligent edge network. Initially, they presented the S-HashMap index structure to ensure efficient and secure data updates while enabling multi-keyword fuzzy ciphertext retrieval. Additionally, to assess similarity between the query vector and index nodes, the authors employed secure k-nearest neighbor (kNN) techniques to compute the Euclidean distance. Their proposed approach eliminates the need for predefined dictionaries and achieves efficient multi-keyword fuzzy search and index updates without complicating search processes.


When EC redistributes data and models to IoT devices, it introduces new security challenges. To address new security challenges in computing redistributes, the author in \cite{xue2022differential} introduced Acies, a privacy-preserving classification system based on differential privacy (DP) designed for EC. Acies safeguards classification models transferred to edge devices and is compatible with popular classifiers like Nearest Neighbourhood, Support Vector Machine, and Sparse Representation Classifier, along with various feature selection techniques. Their evaluations across different datasets demonstrate that classification models integrated with Acies can maintain privacy while retaining high utility. Acies ensures robust privacy even under reconstruction attacks, with minimal effects on classification accuracy.
The authors in \cite{qu2020privacy} offered a novel architecture for training ML models that enables intelligent EC. This architecture operates through two distinct phases: Firstly, a cooperative federated pre-training phase takes place between the cloud and edge server, drawing inspiration from federated learning (FL). This phase incorporates an incentive mechanism designed to ensure fair reward distribution based on the contributions of ESs towards pre-training the model. Secondly, a privacy-preserving model segmentation training phase occurs between the edge server and the device. This phase utilizes homomorphic encryption to enhance and safeguard models on end devices while offloading significant computational tasks to edge servers.

The authors in \cite{hafeezedge} initiated their exploration by introducing sampling and data reduction methodologies. These methodologies facilitate a decrease in the volume of data sent for cloud-based processing. Nevertheless, it is crucial to acknowledge that the use of smaller datasets in ML algorithms may involve potential compromises in accuracy.
An alternative and feasible strategy is to position ML algorithms in close proximity to data sources to minimize data transfer requirements. Within the EC paradigm, three primary modalities are employed to facilitate the execution of ML and data processing functions on intermediary nodes: device-edge, device-cloud, and edge-cloud interactions.
An assessment is conducted of these three cutting-edge procedures in conjunction with conventional methods, leading to a comprehensive discussion of their advantages and disadvantages. Furthermore, this paper \cite{hafeezedge} proposed a novel architecture, elucidating the potential application of EC within the Industrial Internet of Things (IIoT) for both data reduction and achieving successful predictive maintenance (PM).
PM stands as a pivotal IIoT technology designed to continually monitor the health of machinery, enabling the prediction of component failures before they occur.

Numerous applications in EC, like federated ML and multiplayer AR games, require distant clients to engage in cooperative endeavors via message exchanges to achieve common objectives. However, the effective deployment of such cooperative edge applications for optimizing system performance throughout an entire edge network remains a subject of uncertainty.
The authors in \cite{wang2020service} discussed a formal analysis of the issue.
To achieve a holistic system representation, they offered a variety of cost models by presenting an iterative technique called ITEM based on a comprehensive formulation. 
In each iteration, they built a graph to encapsulate all the costs and transformed the cost optimization issue into a graph-cut problem. 
By resolving a sequence of graph cuts employing available max-flow techniques, the lowest-cost shortcut is found.
They established the existence of a parameterized constant approximation ratio for ITEM.
Moreover, they developed an online method called OPTS that is based on optimally alternating between partial and complete placement updates, driven by insights from the optimum stopping theory.

By balancing QoS and energy efficiency, multi-access EC enables IoT applications to locate their services in the ESs of mobile networks. Prior initiatives have placed a primary emphasis on computational requisites, leaving the communication needs related to latency and bandwidth in the domain of IoT comparatively unaddressed. Additionally, the task of modeling urban smart things, elucidating their connectivity with multi-access EC networks, characterizing the multifaceted resource demands encompassing computation, communication, and IoT for application services, and modeling the federation of multiple MEC service providers in an urban environment poses a unique set of challenges for the smart city \cite{bansal2022urbanenqosplace}. 
In response to these research gaps, the authors have presented the following solutions:
i) The "UrbanEnQoSMDP" framework, tailored for optimizing service placement within the "Urban IoT-Federated Multi-access EC-Cloud" architecture to accommodate the computational, per-flow communication, and IoT requisites; 
ii) The "$\epsilon$-greedy with mask" policy, crafted for the systematic selection of suitable USTs in advance to ensure the fulfillment of IoT requirements;
and iii) "UrbanEnQoSPlace," a multi-action deep reinforcement learning (DRL) model that employs the outlined strategy to resolve the "UrbanEnQoSMDP" problem by concurrently considering all services of an application that were created by the Dueling Deep-Q Network.

\begin{figure*}[t]
	\centering
	\begin{subfigure}[b]{0.28\textwidth}
		\centering
		\includegraphics[width=\textwidth]{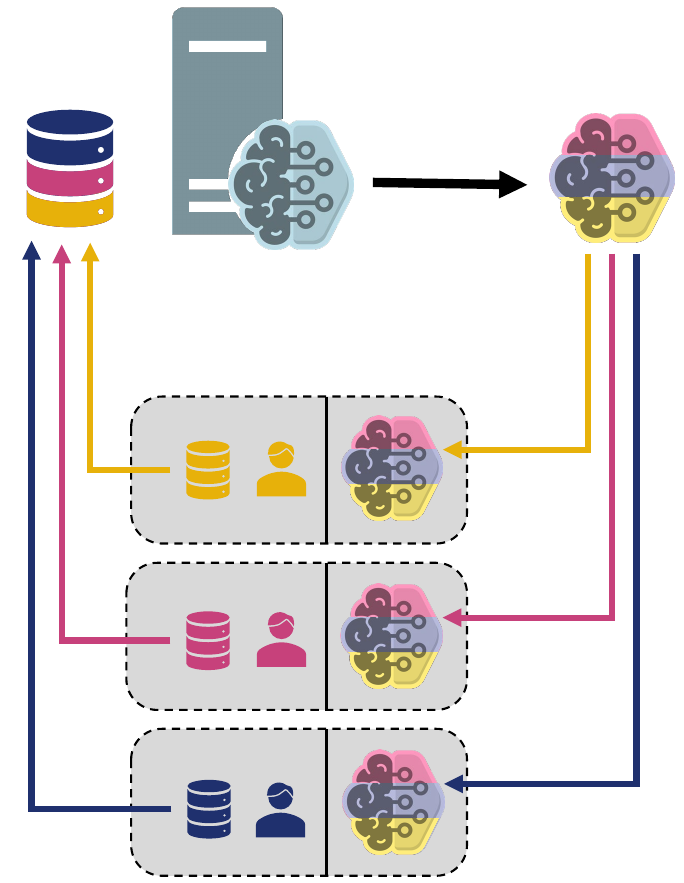}
		\caption{Centralized Learning}
		\label{fig:centralized}
	\end{subfigure}
	\hfill
	\begin{subfigure}[b]{0.4\textwidth}
		\centering
		\includegraphics[width=\textwidth]{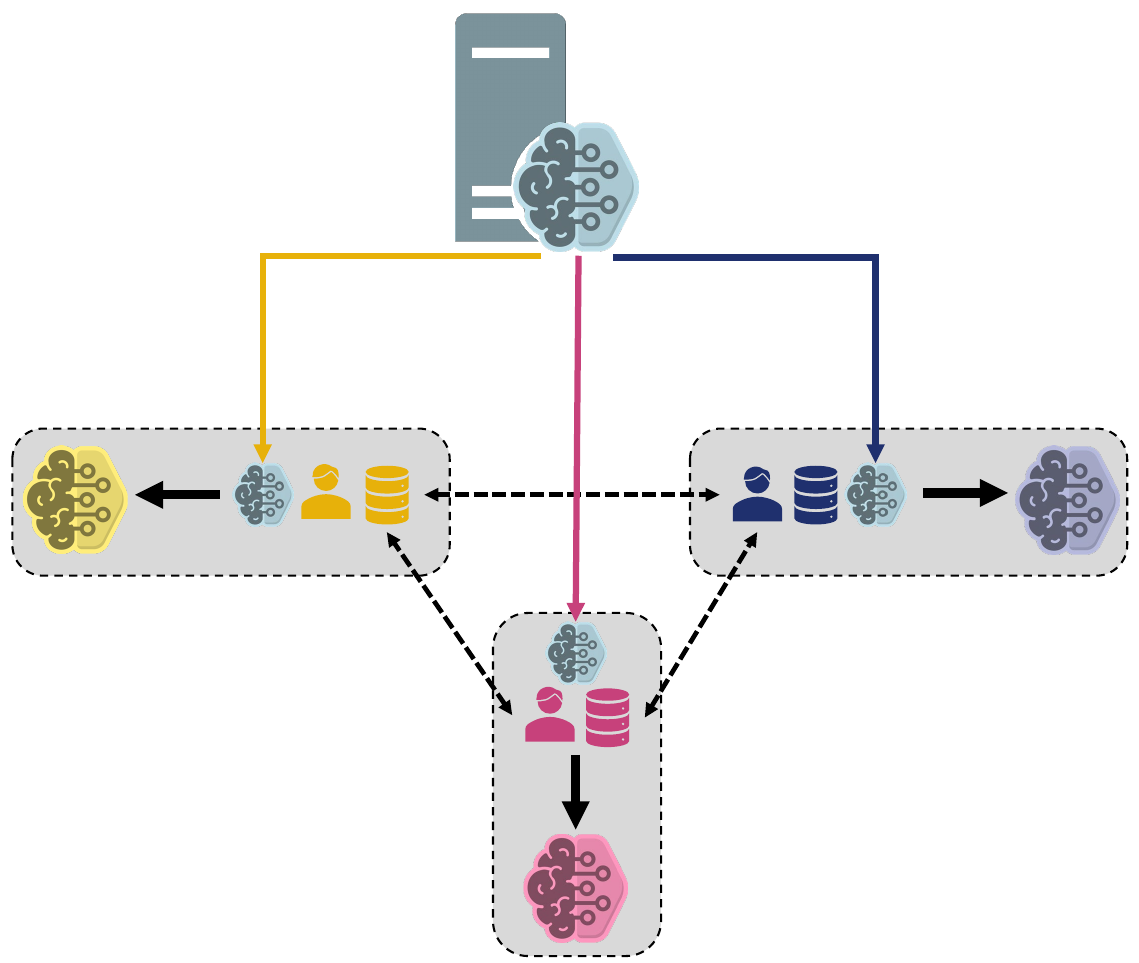}
		\caption{Decentralized Learning}
		\label{fig:decentralized}
	\end{subfigure}
	\hfill
	\begin{subfigure}[b]{0.31\textwidth}
		\centering
		\includegraphics[width=\textwidth]{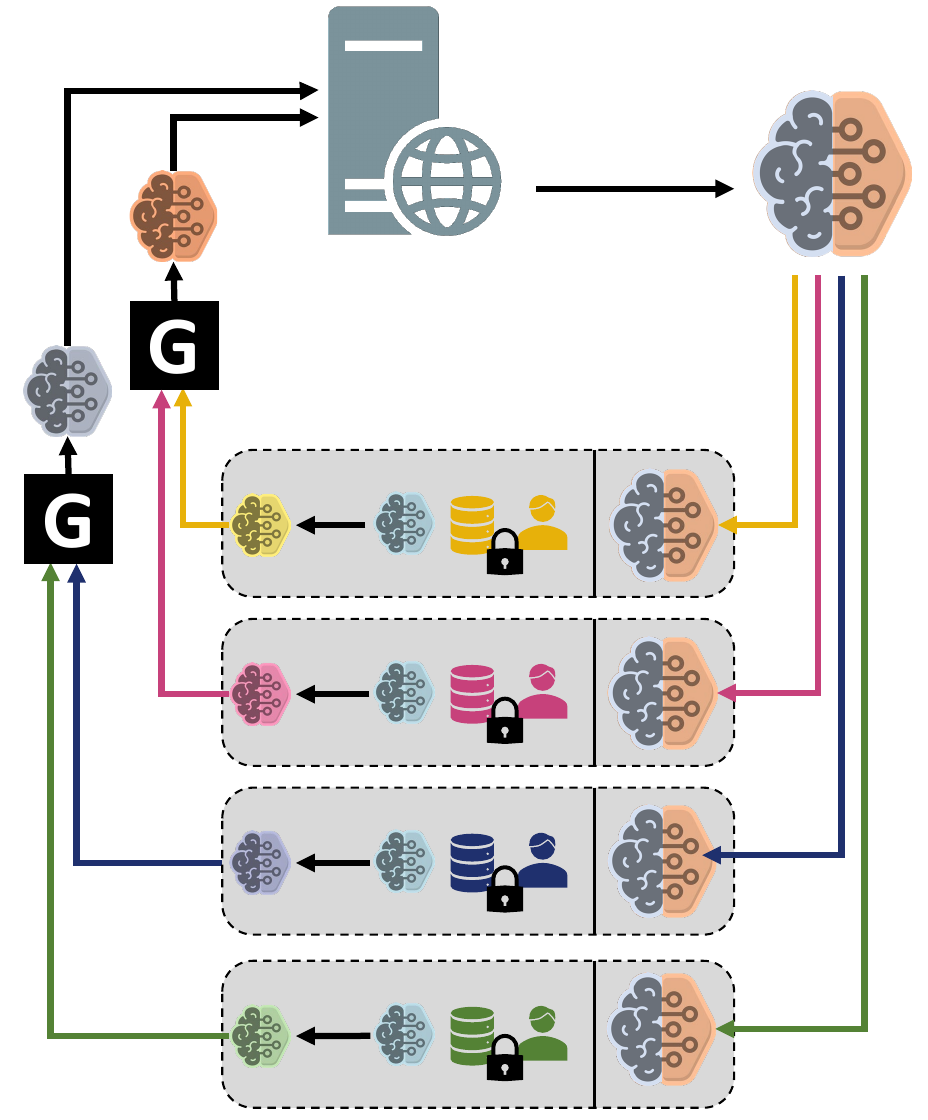}
		\caption{Federated Learning}
		\label{fig:federated}
	\end{subfigure}
	\caption{Comparison of three centralized learning, decentralized learning, and federated learning structures}
	\label{CDF}
\end{figure*}


\subsection{Motivation and Contributions}
FL's inception and its integration with privacy-preserving techniques has successfully persuaded a broad demographic, notably patients, to contribute their sensitive data for AI model training.
This is achieved by transmitting model parameters rather than raw data, alleviating concerns regarding potential privacy risks.
Furthermore, the advent of FL has inspired wireless communication researchers and data scientists, motivating them to bring their previously incomplete practical or theoretical models to completion and operationalization within both academic and industrial domains.
However, FL is currently in its early stages, it's extension has immense potential for our daily lives and various industrial sectors. 
Consequently, in order to foster its prudent development, an extensive volume of research and efforts must be dedicated to realizing substantial advancements in FL-assisted architectures.
In accordance with the graphical depiction provided in Figure 2 in \cite{khan2021federated}, the quantity of research publications in the past three years markedly surpasses that of publications up until the year 2020. In light of this accelerated pace of advancement, it is imperative to direct and structure the
 contemporary trends within this domain systematically.
In light of this accelerated pace of advancement, there arises an imperative to systematically direct and structure the contemporary trends within this domain.
 The mathematical analysis of FL averaging algorithms is paramount in developing and deploying reliable and scalable FL-based schemes.
It provides researchers with a systematic and technical means to calculate impactful parameters, subsequently guiding them in the design and development of their study model.
Ultimately, serving as a roadmap, the identification of current challenges and prospective directions plays a pivotal role in offering guidance to researchers and scholars and expediting the ongoing progression within this domain.
Considering all of these aspects, it is noteworthy that none of the prior surveys cited in Table \ref{CAESPSF} have provided exhaustive coverage of the aforementioned requirements, thereby highlighting a significant gap in this area.
Consequently, driven by recent indicators of advancement within the FL domain. To address extant deficiencies, we present this comprehensive survey as a means to systematically bridge this gap by incorporating of the latest advancements and applications in the FL field.
In alignment with this goal, our initial focus centers on a comprehensive examination of 5G technology and its vulnerabilities, followed by an exploration of the capabilities inherent in 6G networks.
We, then, delve into the intricacies of intelligent IoT, fog, and edge-based architectures.
Moving forward, our research trajectory involves an in-depth examination of the extant literature pertaining to centralized learning, decentralized learning, and, most notably, FL.

\subsubsection{Who benefits?}
FL offers assistance to diverse groups of individuals and researchers across a wide range of disciplines.
\begin{itemize}
\item \textbf{ML researchers} gain significant insights from this survey that allows them to review and select the most suitable methods among existing approaches. These methods enable the distribution of the computational load necessary for training complex ML models across various devices, all while ensuring privacy is maintained.

\item \textbf{Communication operators and service providers} benefit from this survey through a detailed overview of current algorithms that decrease data transmission and improve user privacy, resulting in reduced operational costs, enhanced customer trust, and adherence to data protection regulations.

\item \textbf{Aerial and non-terrestrial network researcher} benefit from this survey by achieving a detailed overview of FL in their field and readily evaluating the reported performance of existing methods. This enables them to identify current gaps and limitations in order to devise new optimization algorithms and collaboration methods to reduce the complexity of these network infrastructures.

\item \textbf{Healthcare systems and hospitals.} By examining this survey, healthcare systems and hospitals can evaluate the strength of FL algorithms in protecting confidential patient data, thereby guaranteeing data privacy while benefiting from large ML models trained on extensive datasets from numerous healthcare sources.

\item \textbf{Smart city and home companies} can utilize this survey to investigate current technologies in the field, identify existing drawbacks, and strategically invest their time and resources in developing novel methods and technologies to increase their revenue. 
\end{itemize}

\subsubsection{The contributions of Survey}
The main contributions of this survey are listed as follows:

\begin{itemize}

\item Offering mathematical analyses and algorithmic frameworks for a multitude of FL averaging techniques, encompassing a comprehensive range of methodologies and approaches, provides a robust foundation for understanding and improving FL processes. These analyses enable the evaluation of various strategies, enhancing their effectiveness and adaptability in different scenarios.

\item Delineating prospective research trajectories that involve identifying potential avenues for exploration and development within each distinct area of FL. Furthermore, addressing unresolved queries entails investigating outstanding questions and uncertainties that persist within the field, aiming to provide clarity and advance knowledge in these domains.

\item Expanding on the task of cataloging accuracy and AI/ML methods utilized by each referenced source, which involves meticulously documenting the specific techniques and algorithms employed for each aspect of FL under examination,. This comprehensive approach provides a detailed overview of the methodologies and strategies adopted across various research endeavors, contributing to a deeper understanding of the field's landscape and capabilities.
\item Compiling a selection of open research areas and future trends aims to help researchers in their future investigations by providing valuable insights into emerging topics. This compilation serves as a comprehensive guide, highlighting potential areas for exploration and innovation. Consequently, it facilitates the identification of key opportunities and challenges within various fields.
\end{itemize}
\subsection{Survey Outline}
The remainder of the survey is structured as follows:
Section \ref{sec:comover} delineates centralized, decentralized, and federated learning, and provides a comprehensive exposition of FL-based structures.
The security concerns associated with FL are discussed in Section \ref{sec:sec}.
Sections \ref{sec: resalofl} delves into the subject of resource allocation within FL systems.
The applications of FL are presented in Section \ref{sec:appfl}.
Section \ref{sec:scale} is dedicated to a comprehensive discussion on the scalability aspects of FL architectures.
Finally, Section \ref{sec:conc} brings the survey to its conclusion.
\section{Comprehensive Overview of FL} \label{sec:comover}
Understanding the distinctions between decentralized, centralized, and FL is crucial for selecting the most suitable approach based on specific requirements and constraints. Each method presents unique trade-offs in terms of privacy, scalability, and efficiency, shaping the future of ML paradigms. We will elaborate on each scenario below, providing detailed explanations.
\subsection{Centralized learning}
Centralized learning conducts training on a single server or cluster, simplifying coordination and resource allocation. It offers easy access to the entire dataset, promoting consistency and standardization in model training. This approach is advantageous for industries with ample computational resources and where data privacy concerns are manageable. However, it may raise concerns regarding data privacy and scalability, particularly with large datasets. For example, the authors in \cite{sun2023joint} explored resource allocation in centralized wireless cellular communication with multiple cells, users, and channels.  To address resource allocation in a centralized structure, they introduced an approach combining Deep Deterministic Policy Gradient (DDPG) reinforcement learning and unsupervised learning. Their algorithm constructs a DDPG-based channel allocation neural network and an unsupervised learning-based power control neural network. By using double experience replay, it trains these networks to adapt to dynamic wireless environments, optimizing energy efficiency and transmit rates. 

The integration of ML is becoming critical for the development of future networks and IoT applications, with input components of ML units (MLU) exhibiting varying degrees of relevancy for precise output determination. In \cite{gharouni2021relevance}, by assuming the transmission of attributes from multiple terminals, the resource allocation problem for an ML-based centralized control system was re-examined, emphasizing relevancy to enhance MLU performance. A heuristic greedy resource allocation algorithm was introduced, leveraging lookup tables created via a Kullback-Leibler divergence (KLD)-based method to assign quantization bits to input attributes. This algorithm is applied to a network of inverted pendulums on carts, utilizing MLUs as controllers. 

The operation and control of microgrids (MGs) continue to be a significant area of research. The authors in \cite{calloquispe2023centralized} examined frequency control in islanded MGs, emphasizing the importance of distributed energy resource integration and system stability. Normally, secondary control in MGs has depended on proportional-integral controllers. Therefore, the authors proposed a centralized secondary control strategy using a deep deterministic policy gradient (DDPG) agent based on RL. The results indicate that RL-based control algorithms successfully address the shortcomings of conventional control methods, improving the stability, reliability, and overall performance of MGs.
The authors in \cite{ren2022meta} introduced a gradient-based meta-reinforcement learning approach for centralized traffic signal control, which extracts meta-knowledge from multiple meta-training tasks to improve the efficiency of training on new tasks. To mitigate the curse of dimensionality in centralized control, they developed a specialized agent that employs the Divide and Conquer paradigm to decompose the search space. This combination of centralized control and meta-reinforcement learning is, to the best of our knowledge, unprecedented. The combination of gradient-based meta-learning and centralized control RL methods achieves robust generalization and scalability. 

The advent of 6G networks is expected to bring a surge in remote device deployment, necessitating more data processing. Utilizing satellite networks to transfer data to cloud servers for ML analysis is a promising solution. However, this approach may encounter bottlenecks due to lengthy transmissions on satellite channels. The authors in \cite{rodrigues2022deep} examined these challenges by  modelling and analyzing a service model, revealing the limitations of centralized learning over satellite networks. It assesses when centralized learning is feasible and when other techniques are needed due to transmission delays and overloaded channels.

\subsection{Decentralized learning}
Decentralized learning involves distributing the training process across multiple nodes or devices, prioritizing privacy and scalability. Each node trains on its local data, preserving privacy and reducing data transfer needs. This approach fosters robustness against node failures and network disruptions. Ideal for scenarios where data privacy is paramount, such as healthcare or finance. The challenge in decentralized learning is effectively coordinating decentralized learning while preserving data privacy and learning security across the board.
The authors in \cite{alsagheer2023decentralized} investigated decentralized ML governance, encompassing ML value chain management, community-based decision-making for the ML process , ownership and rights management of ML assets, decentralized identity for the ML community, decentralized ML finance, and risk management. Decentralized learning empowers edge users to collaboratively train models through device-to-device communication. However, existing approaches have been limited to wireless networks with reliable workers and fixed topologies. To tackle this, the authors in \cite{jeong2022asynchronous} introduced an asynchronous, decentralized stochastic gradient descent (DSGD) algorithm that is resilient to computation and communication failures at the wireless network edge. Theoretical analysis of their model performance includes a non-asymptotic convergence guarantee. Multi-agent Decentralized Learning provides scalability, allowing agents to learn from their local datasets. However, challenges arise from dataset heterogeneity, communication graph structure, and quantifying prediction uncertainty. To address them, the authors in \cite{9983539} introduced "G-Fedfilt," a new aggregation rule inspired by graph signal processing (GSP) and based on graph filtering. This proposed aggregator facilitates a structured flow of information by leveraging the topology of the graph. The authors in \cite{alshammari2024baygo} introduced BayGO, a fully decentralized Bayesian learning approach with local averaging. Within BayGO, agents learn posterior distributions over model parameters locally and share this information with neighbors. An aggregation rule ensures optimality and consensus. The theoretical analysis provided by the authors establishes the convergence rate of agents' posterior distributions, considering network structure and information heterogeneity. Agents optimize this rate by considering their posterior distributions and network structure, resulting in a sparse graph configuration that enhances communication efficiency.

D2D-assisted decentralized learning is a key solution for mobile devices to collaboratively train AI networks without a centralized parameter server. However, dense network connections lead to high learning latency and energy consumption. Link selection and aggregation weight also significantly impact learning performance. To address these challenges, the authors in \cite{liu2023communication} proposed joint mechanisms for computing power adjustment, wireless resource allocation, link selection, and aggregation weight adaptation. They analyzed learning performance metrics and formulated an optimization problem to minimize total learning cost, considering learning latency and energy consumption. The convergence speed of D2D-aided decentralized optimization algorithms is heavily influenced by network connectivity, with denser topologies resulting in faster convergence. However, the limited communication range of wireless nodes in real-world mesh networks reduces local connectivity, degrading the efficiency of decentralized learning protocols and potentially making them impractical. In \cite{zecchin2022uav}, the authors explored the use of unmanned aerial vehicles (UAVs) as flying relays to enhance decentralized learning procedures. They proposed an optimized UAV trajectory, defined as a sequence of waypoints that the UAV visits sequentially to transfer intelligence across sparsely connected groups of users. The authors in \cite{liang2023secure} introduced a hierarchical decentralized learning framework for IoV networks, incorporating multiple fog nodes and mobile vehicles. They proposed a network-level masking mechanism and optimized the consensus matrix for efficient signaling in IoV. The network-level masking removes the pairing demands for inter-fog handovers of mobile vehicles and is proven to be cancelled through distributed consensus.
The authors in \cite{lv2023blockchain} explored algorithms and tools to facilitate device collaboration in constructing personalized DL models for pervasive computing applications in decentralized structures. They first elaborated whether individual devices owned by different users learn robust models personalized to their user's experiences while safeguarding data privacy. This approach, called Opportunistic Collaborative Learning (OppCL), led to the development of a resilient learning algorithm that mitigates the challenges raised by unexpected mobility patterns. Moreover, OppCL was enhanced from homogeneous networks to heterogeneous ones using the lossy compression techniques for DL models. \\
However, there may be some concerns with consistency and synchronization due to more complex coordination and communication among nodes in ultra-dense networks.

\subsection{Federated Learning}
FL merges decentralized and centralized approaches, enabling collaborative model training across distributed devices while preserving privacy. In FL, data is decentralized, generated, and retained on the individual machines or clients where it originates. Each device possesses its own distinct set of local data. It facilitates privacy-preserving updates by transmitting model parameters instead of raw data. FL periodically aggregates these updates on a central server to refine a global model, offering adaptive personalization without compromising privacy. Suitable for applications where data privacy is critical, such as personalized recommendations or predictive maintenance. FL ensures superior security and privacy by maintaining decentralized data and aggregating model changes in a manner that prioritizes privacy protection \cite{abdulrahman2020survey, khan2021federated}. FL enables devices to retain their data while contributing to model training through the sharing of model parameters and weights. In contrast, centralized learning gathers data from various sources. 
To update parameters, FL sends them to a single central server for global model training Fig. \ref{CDF}.

The transition from centralized (cloud-based) to distributed on-device learning (edge server) has led to the emergence of a new paradigm known as FL. The objective of this approach is to retain collected data on local devices and servers for training local models, safeguarding the privacy of data and confidential information. FL systematically addresses storage and communication costs while ensuring a notable degree of privacy at the client level. At first glance, we can summarize the benefits of FL as follows:

\begin{table*}[t]
\footnotesize
  \caption{References on Applicable FL-based Structures}
  \label{AFLS}
  \setlength{\aboverulesep}{0.05cm}
  \setlength{\belowrulesep}{0.05cm}
  \renewcommand{\arraystretch}{1.2}
\begin{tabularx}{\textwidth}{cccc>{\raggedright\arraybackslash}m{11.7cm}}
    \specialrule{0.12em}{0pt}{0pt}
    Reference& Year&Acc$>90\%$&AI/ML approach & \multicolumn{1}{c}{Open Areas and Future Challenges and Directions}\\
    \specialrule{0.12em}{0pt}{0pt}
\cite{huang2022accelerating}&2021&\cmark&TOFEL&\textbf{\textbf{---}}\\ \specialrule{0.0005em}{0pt}{0pt}
\cite{xu2021accelerating} & 2021 &\cmark&DNN/FL-PQSU& Testing FL-PQSU with new models/datasets and IoT devices, and alternative compression processes applicable to FL will be investigated. \\ \specialrule{0.0005em}{0pt}{0pt}
\cite{aouedi2022intrusion}&2022&\xmark&Semi-supervised&\textbf{\textbf{---}}\\ \specialrule{0.0005em}{0pt}{0pt}
\cite{geng2021adaptive}&2021&\cmark&CNN&\textbf{---}\\ \specialrule{0.0005em}{0pt}{0pt}
\cite{li2022data}&2022&\xmark&FEDGS&\textbf{---}\\\specialrule{0.0005em}{0pt}{0pt}
\cite{zhang2020efficient} & 2021&\xmark&DNN& Exploring the effect of hyperparameters ($\alpha, \beta, and \: \gamma$) to find the optimal Big.Little branch designs. \\\specialrule{0.0005em}{0pt}{0pt}
\cite{yang2021tree} &2021 &\cmark&DML& Enhancing E-Tree efficiency by additionally adjusting the number of layers and aggregation frequency with RL enhanced. \\\specialrule{0.0005em}{0pt}{0pt}
\cite{gao2021federated}&2021&\xmark&FS&\textbf{---}\\\specialrule{0.0005em}{0pt}{0pt}
\cite{chiu2020semisupervised}&2021&\cmark&E2E-FL&\textbf{---}\\\specialrule{0.0005em}{0pt}{0pt}
\cite{math2021reliable}&2020&\xmark&semisupervised&\textbf{---}\\\specialrule{0.0005em}{0pt}{0pt}
\cite{kim2021cooperative}&2021 &\textbf{---}&\textbf{---}& Developing a resilient control scheme for fuzzy-logic-based cooperative game theory and applying high-order control to a dynamic system. \\\specialrule{0.0005em}{0pt}{0pt}
\cite{orescanin2021federated}& 2021 &\xmark&FFT& Replacing the router with another edge vehicle and involving secondary clients equipped with sensors to create an E2E edge device network. \\\specialrule{0.0005em}{0pt}{0pt}
\cite{li2021federated}& 2021 &\textbf{---}&DQN& Using public infrastructure updates, multiple data migrations, and fault tolerance to fulfill the needs of vehicular edge networks. \\\specialrule{0.0005em}{0pt}{0pt}
\cite{sun2022long}& 2022 &\textbf{---}&FedTDLearning& Studying a unique approach that combines self-supervised learning and deep RL to facilitate the convergence challenge in some areas.
\\\specialrule{0.0005em}{0pt}{0pt}
\cite{xia2022pervasivefl}&2022&\xmark&NN/DML&\textbf{---}\\\specialrule{0.0005em}{0pt}{0pt}
 \specialrule{0.12em}{0pt}{0pt}
\end{tabularx}
\end{table*}

\begin{itemize}
 \setlength{\itemsep}{1.5pt} 
	\item Offline operability
	\item Enhanced latency performance
	\item Enhanced accuracy
	\item Privacy fortification
	\item Prolonged battery life
	\item Localized model training
\end{itemize}

FL is a decentralized approach that prioritizes privacy by retaining raw data on devices and employing local ML training. This strategy effectively reduces data transfer overhead. The accumulated knowledge is subsequently aggregated and disseminated among participants through a federation of the learned and shared models on a central server. The authors in \cite{abdulrahman2020survey} compared and contrasted several ML-based deployment models before delving deeply into FL. Unlike previous assessments in the field, they introduced a fresh classification of FL challenges and research domains. This classification stems from a meticulous analysis of the primary technical barriers and ongoing endeavors within the discipline.
In \cite{khan2021federated}, the writers first outlined recent FL developments that have enabled FL-powered IoT applications. A comprehensive set of metrics, encompassing sparsification, resilience, quantization, scalability, security, and privacy, were established to thoroughly evaluate the latest advancements in the field. Next, a taxonomy for FL across IoT networks was introduced and executed.

The traditional path of centralized cloud-based learning and processing for IoT platforms encounters growing challenges stemming from the elevated costs associated with transmission and storage, along with escalating privacy concerns. The most promising alternative strategy to address this issue is FL. In FL, training data-driven ML models entails collaboration among multiple clients without the need to transfer the data to a centralized location. This approach effectively reduces transmission and storage costs while providing a high level of user privacy. Implementing FL systems on IoT networks still encounters obstacles across various sectors, including manufacturing, transportation, energy, healthcare, quality and reliability, business, and computing. Therefore, the advantages and disadvantages of FL in IoT systems are covered in \cite{zhang2022federated,divan2021recent,kontar2021internet,baccour2022pervasive}, as well as how it may support a variety of IoT applications. They specifically identified and analyzed several significant challenges in FL over IoFL and described new potential solutions to address the aforementioned challenges.

In \cite{arisdakessian2022survey}, a thorough analysis of the suggested intrusion detection systems for the IoT structures was studied, which consisted of IoT devices and communications between the layers of cloud, fog, and the IoT.
Considering all of these aspects, it is noteworthy that none of the prior surveys cited in Table \ref{CAESPSF} have provided exhaustive coverage of the FL-based structures' challenges requirements, and future works, thereby highlighting a significant gap in this area. Consequently, driven by recent indicators of advancement within the FL domain. To address extant deficiencies, we present this comprehensive survey as a means to systematically bridge this
gap by incorporating of the latest advancements and applications in the FL field. To this purpose, this work stands out in three distinct aspects: (1) Examining privacy concerns within the IoT structures, considering interactions across IoT, fog, and cloud computing layers, as well as IoT devices. (2) Offering an innovative two-tier categorization system that initially classifies the literature into groups based on the methods used to identify attacks. Subsequently, each method is further subdivided into several techniques for thorough analysis;(3) In order to enhance future IoT systems against cyberattacks, we introduce a comprehensive cybersecurity framework that integrates principles from Explainable AI (XAI), FL, game theory, and social psychology. The reliability of local models in FL for anomaly detection may vary. For example, multiple trained models are likely to exhibit characteristics of abnormal data as a result of noise corruption or failures in anomaly detection. Moreover, there is a possibility of contamination in the training data or model weights, as the communication protocol between edges could be exploited by attackers. Consistent with this view, the authors in \cite{qin2020selective} chose the local models participating in model aggregation while designing a federated training procedure. Their study uses an observed dataset to compute prediction errors to filter out the poor local models from federated training. 

In order to tackle the heterogeneity challenge in FL and enhance communication and computation efficiency, Huang \textit{et al.} in \cite{huang2022accelerating} introduced a novel technique called topology-optimized federated edge learning (TOFEL). Their objective is to minimize the weighted sum of energy consumption and delay. Therefore, the problem of joint optimizing the aggregation topology and processing speed is defined. They introduced a novel penalty-based sequential convex approximation approach to address the mixed-integer nonlinear problem. Deep neural networks (DNNs) are trained offline to imitate the penalty-based technique two to simplify real-time decision-making. The trained imitation DNNs are then deployed at the edge devices for online inference. Thus, the TOFEL architecture smoothly includes an effective imitation-based learning strategy. Due to their limited computational capabilities and poor-quality network connections, it is frequently impossible or extremely slow to train DNNs using the FL pattern on IoT devices. In \cite{xu2021accelerating}, a novel, effective FL framework called FL-PQSU was offered to deal with this issue. The pipeline consists of three stages: structured pruning, weight quantization, and selective updating. These stages are combined to reduce computation, storage, and communication costs, thereby accelerating FL training. The authors examined FL-PQSU by employing established DNN models such as AlexNet and VGG16, along with publicly available datasets like MNIST and CIFAR10. Their findings demonstrated that FL-PQSU effectively manages learning overhead while maintaining training performance.

To facilitate the identification of abnormal log patterns in large-scale IoT systems, the authors in \cite{li2022federated} presented a configurable and communication-efficient federated anomaly detection technique, and from now on, we refer to it as FedLog. 

They initially developed a model based on Temporal CN with an Attention Mechanism to extract detailed features from system logs. Additionally, to enable IoT devices to collaboratively produce a comprehensive anomaly detection model in a privacy-preserving manner, they devised a novel FL framework. Finally, they offered a masking approach inspired by the lottery ticket hypothesis to address non-independent and identically distributed (non-IID) log datasets. This approach is configurable and communication-efficient. They evaluated the performance of their proposed scheme using two widely used and publicly available real-world datasets HDFS and BGL. Their findings demonstrates that their adopted FedLog method is effective for identifying anomalies based on log data. Aouedi \textit{et al.} in \cite{aouedi2022intrusion} introduced a semi-supervised FL paradigm for IDS to address challenges like high bandwidth overhead, and the reluctance of devices to communicate private data. Correspondingly, substantial computing and storage resources are required on the server side to label and process such data.
They utilized network software for deployment and automation, wherein clients train unsupervised models to learn representative and low-dimensional features, and the server executed a supervised model. The authors in \cite{foukalas2021federated} provided an FL protocol for fog networking applications. The Internet Engineering Task Force's concept is compatible with the fog networking architecture. The FL protocol was devised and outlined to encompass IoT devices spanning from the edge to the cloud. To achieve this goal, experimental trials were conducted to assess the efficacy of their proposed distributed edge intelligence technology for specific application scenarios. Their findings demonstrated the effectiveness of the proposed FL protocol in terms of message latency and the accuracy of intelligence. These protocols are aimed at serving as the foundation of the next generation of the Internet, as they enable more efficient distribution of edge intelligence to the vast number of newly connected IoT devices. The authors in \cite{rasti2022graph} tackled the challenges of device selection and resource allocation in an indoor setting where multiple smart devices are involved in the FL process. To further minimize communication latency, the proposed system employs visible light communication for downlink transmission and utilizes a radio frequency access point for uplink transmission. Multi-Layer Hierarchical FL (MLH-FL), a unique method for FL, was presented in \cite{de2021novel} along with a multi-layer architecture. MLH-FL makes use of the traditional FL and MLH-FL techniques' accuracy. To achieve this objective, a hierarchical design at the edge was proposed to leverage model aggregations at different levels, in contrast to the conventional FL approach. As a result, model aggregations can be performed even when a subset of edge nodes is not consistently connected to the cloud. Moreover, this strategy reduces the frequency of communication with the cloud for model aggregations, conserving communication energy. Their study also addresses the concept of a low-level round, which permits aggregations to be iterated at the edge without transmitting updated models to the cloud after each iteration. 
The authors in \cite{geng2021adaptive} surveyed a method for diagnosing bearing faults based on FL. The model aggregation process selects high-quality local models to participate, utilizing the accuracy threshold adaptive algorithm to reduce communication overhead.

Centralized learning and FL differ significantly in that data in the former needs to be offloaded, whereas it is trained locally in the latter. Guo \textit{et al.} \cite{guo2020computation} investigated the issue of compute offloading for EC-based ML in an industrial setting, considering the machine learning models mentioned earlier. To minimize training latency, they formulated an offloading problem based on ML and addressed the problem using an energy-constrained delay-greedy method.
For businesses enabled by 5G, the authors in \cite{li2022data} offered FEDGS, a hierarchical cloud-edge-end FL architecture, to enhance industrial FL performance on non-IID data. FEDGS utilizes a gradient-based binary permutation approach to select a subset of devices within each factory and form homogeneous super nodes, which participates in FL training using naturally grouped factory devices. The training process is then coordinated both within and across these super nodes using a compound-step synchronization approach, which demonstrated outstanding resilience against data heterogeneity. The proposed guidelines saves time and can adapt to changing conditions without putting sensitive industrial data at risk through risky manipulation. They demonstrated that FEDGS outperforms FedAvg in terms of convergence performance, and then postulates a less stringent requirement under which FEDGS exhibits greater communication efficiency.

To address the computational and memory resource limitations that restrict the capabilities of hosted DL models, the authors in \cite{zhang2020efficient} offered a collaborative Big.Little branch design to allow effective FL for artificial IoT (AIoT) applications. Their method deploys deep neural network (DNN) models across both cloud and AIoT devices, drawing inspiration from BranchyNet's architecture, which incorporates multiple prediction branches. The Big-Little branch of the tiny branch model, tailored for AIoT devices, was placed on the cloud to enhance prediction accuracy. AIoT devices resort to the large branch for additional inference when they are not able to confidently make predictions using the local, tiny branches. Following that, the authors implemented a two-stage training and co-inference strategy, which takes into account the local features of AIoT scenarios. This approach enhances the prediction accuracy and minimize early departures of the Big-Little branch model. In \cite{yang2021tree}, Yang \textit{et al.} presented a novel decentralized model learning method called E-Tree, which uses an edge to systematically design tree structure. To enhance training convergence and model accuracy, consideration was given to planning the tree structure and determining the locations and order of aggregation on the tree. Specifically, by considering the data distribution on the devices and the network distance, they developed an effective device clustering technique called K-Means with average accuracy for E-Tree.\\
In \cite{tian2021faliotse}, the authors examined hostile attacks on time-series analysis in an IoT search engine (IoTSE) system. In particular, they utilized a simulated FL system to construct the basis model, LSTM Recurrent Neural Network (RNN). They proposed Federated Adversarial Learning for IoT Search Engine (FALIoTSE), which leverages shared parameters of the federated model to address adversarial example creation and enhance robustness. The impact of an attack on FALIoTSE is demonstrated under various levels of disruption using real-world data from a smart parking garage.\\ 
By establishing particular updating weights for each node based on the distinction between the global and local models, the authors in \cite{gao2021federated} created an elastic local update method that could train the customized models. Their approach takes into account both the individual characteristics of local models and their collective consistency. Additionally, they introduced an n-soft sync model aggregation technique, which significantly reduces training time by combining synchronous and asynchronous aggregations. To address the E2E reliability of FL, and  communications, Chiu \textit{et al.} in \cite{chiu2020semisupervised} focused on an intelligent, lightweight method based on the standard software-defined networking (SDN) architecture to manage the large FL communications between clients and aggregators. The handling method adjusts the model parameters and batch sizes for each individual client to reflect the observed network conditions identified by the kNN technique.\\ 
Math \textit{et al.} \cite{math2021reliable} investigated an edge learning system based on FL and semi-supervised learning to address data security and network bandwidth limitations. The system adapts FL technology to train AI models at edge devices using an enhanced semi-supervised learning method. It periodically uploads the training results to the cloud server to compile a unified model. Subsequently, they observed that the data on the end devices exhibits non-IID characteristics in real-world scenarios, which results in weight divergence during training and notably diminish model performance.
To mitigate the adverse effects of weight divergence, they proposed a novel technique called federated swapping. This method substitutes partial FL operations based on a subset of shared data points during federated training. 

In \cite{zhao2020cluster}, the devices are grouped into equally sized clusters, and clients are selected proportionally from each cluster. Following that, the authors devised a distinctive framework called cluster-based federated averaging to attain an equitable global model.
By implementing this approach, there is a possibility that the accuracy of the minority group may be significantly enhanced, potentially at the expense of the majority group. They adjusted the training weights as features to partition the users, ensuring that the training data of customers remains on their devices in accordance with FL's emphasis on privacy protection. 

In \cite{kim2021cooperative}, the author reached a mutual understanding by examining the mutual benefits to establish a reciprocal consensus between two distinct negotiating techniques: the weighted average solution and the constant elasticity substitution technique. Their strategy's primary innovation is the exploration of the dual-interactive bargaining process, which relied on the interdependence between IoT devices and the tactical edge server. Furthermore, to optimize tactical edge-assisted job offloading services, they collaboratively investigated various negotiating strategies.\\
While the literature has shown significant interest in optimizing client selection policies, less attention has been given to the design of the actual execution of such policies, particularly focusing on client discovery techniques. To address this gap, an edge-based framework was proposed to improve FL client discovery processes. This framework utilizes (i) the Message Queue Telemetry Transport (MQTT) protocol for FL client-server interactions to determine the capabilities of future clients, and (ii) the Lightweight Machine-to-Machine (LwM2M) standard from the Open Mobile Alliance (OMA) for the semantic definition of these capabilities \cite{genovese2022enabling}.\\
Orescanin \textit{et al.} in \cite{orescanin2021federated} presented two main contributions to address high communication costs for embedded applications: (1) In order to enhance system security and model performance, it is necessary to enhance the weight initialization phase of the FedAvg algorithm. Moreover, it is essential to limit the proportion of weights that can be averaged; and (2) deploying a realistic model using an edge device as the server node, which is a first for a federated system, to showcase the effectiveness of the proposed approach. They utilized a MobileNetV2 model that is pre-trained centrally on the CelebA dataset for evaluation purposes. They applied the modified FedAvg approach along with FFT to monitor the model parameters transmitted over the network and to measure CPU load, power consumption, device memory usage, and communication metrics. With MEC servers equipped with AI, the authors in \cite{li2021federated} examined the problem of cooperative data sharing in vehicular edge networks. They also introduced a particular way of exchanging data collaboratively. Subsequently, to ensure efficient and secure data sharing in the VEN, they introduced a novel collaborative data-sharing method leveraging a deep Q-network and FL. Han in \cite{han2021fedmes} considered FL with several local wireless edge servers. In a more realistic setting, their primary objective is to accelerate training. The core concept of their approach is to utilize clients in overlapped coverage areas between adjacent ESs. During the model-downloading stage, clients in overlapping areas receives multiple models from different ESs, averaged these models, and then updates the averaged model with local data. These clients utilize broadcasting to disseminate their updated models to multiple ESs, which act as bridges for transferring learned models between servers. 
Although certain ESs may have biased datasets within their coverage areas, clients in the overlapping regions of nearby servers could aid in the training processes of those ESs. As a result, compared to traditional cloud-based FL systems, their proposed technique significantly reduces the overall training time by eliminating the need for costly connections with the central cloud server, typically located at the upper tier of edge servers.\\
Traditional studies on order-and-driver matching solely relies on pure combinatorial optimization models, neglecting the long-term advantages of the dynamic MOD decision-making process. To address the aforementioned issue, the authors in \cite{sun2022long} introduced a systemic paradigm of online matching with federated neural temporal difference learning. This paradigm encompassed both learning and matching phases, aiming for long-term optimization. During the learning phase, the Markov decision process (MDP) is employed to simulate the long-term matching process. This process is typically tackled using data-driven RL in an offline central training scheme. Industrial MOD systems generate vast amounts of data on the network. However, due to network bandwidth limitations and security concerns, transferring all this large-scale industrial data to the cloud server for centralized model training is not feasible.
A novel and generalized form of federated neural temporal difference learning (FedTDLearning) was proposed to achieve long-term matching in a distributed manner. During the matching phase, a real-time bipartite matching optimization problem is formulated to maximize the obtained spatiotemporal value and minimize the pickup distance. This problem is expected to be reduced to the minimum-cost, maximum-weight bipartite graph matching problem. By jointly optimizing FedTDLearning and the combinatorial fractional programming technique, a distance-learned-value ratio algorithm is developed to achieve optimal matching in the bipartite network. Furthermore, to maximize computing efficiency, they addressed the real-time matching problem by constructing a bipartite kNN network where each order is connected to its kNN drivers. In \cite{fadlullah2021smart}, an asynchronously updating FL model for edge nodes was devised to develop regional AI models for smart remote sensing, with a specific application for forest fire warning, all without the need for explicit data exchange with the cloud. Their scheme decreases network overhead while simultaneously safeguarding data privacy. To enable efficient and effective FL across heterogeneous IoT devices, the authors in \cite{xia2022pervasivefl} introduced a novel framework called PervasiveFL. In PervasiveFL, a lightweight NN model known as modellet is deployed on each device without modifying the original local models.
Modellets and local models have the ability to selectively learn from each other through soft labels by leveraging locally collected data, utilizing deep mutual learning (DML) and their entropy-based decision-gating technique. Modellets can share the information they have acquired among devices in a traditional FL manner, as they shared the same architecture. This facilitates the broad application of PervasiveFL to any heterogeneous IoT system, ensuring high inference accuracy while minimizing communication overhead. The future directions, open areas, and accuracy of references in FL-based structures are presented in Table \ref{AFLS} along with detailed information.
\subsubsection{Mathematical Analysis of FL Averaging}
Analyzing a large amount of raw data in cloud centers is impractical due to security challenges, latency issues, and adverse impacts on data transmission. Hence, steering the IoT infrastructure and capabilities towards intelligent edge centers can significantly enhance data processing and wireless network functionalities. Intelligent edge centers aid in the aggregation, processing, optimization, and management of conventional wireless networks. Precisely, intelligent edge infrastructures have accelerated all communication applications while retaining data in local centers instead of transmitting it to the central hub. Nevertheless, smart edges are still in their infancy and have much room for improvement. Various algorithms have already been developed, employing different mathematical approaches to significantly enhance wireless network performance and security (see Table \ref{DFAA}).
\begin{table}[t]
\footnotesize
\caption{Different Federated Averaging Algorithms}\label{DFAA}
  \setlength{\aboverulesep}{0.05cm}
  \setlength{\belowrulesep}{0.05cm}
  \renewcommand{\arraystretch}{1.2}
\centering
\begin{tabular}{cc}
\specialrule{0.12em}{0pt}{0pt}
Papers& Algorithm \\
\specialrule{0.12em}{0pt}{0pt}
\cite{chen2021federated,chen2022federated,bai2022fedewa,yang2022efficient}& Weighted \\
\cite{reddi2020adaptive,wang2022communication,munoz2019byzantine,poli2022adaptive,jayaram2022adaptive,wu2022adaptive,wang2019adaptive} & Adaptive  \\
\cite{wang2023boosting,das2022faster,kim2022communication,zhao2023clustered,salazar2023fair} & Momentum  \\
\cite{pasquini2022eluding,bonawitz2017practical,stevens2022efficient,li2020secure,so2023securing,kim2023cluster} & Secure  \\
\cite{kang2022communication,lang2023joint,prakash2022iot,honig2022dadaquant,oh2022communication,liu2022hierarchical}& Quantization  \\\specialrule{0.12em}{0pt}{0pt}
\end{tabular}
\end{table}
\subsubsection{ Weighted Federated Averaging}
\begin{figure}[h]
	\centering
	\includegraphics[width=80mm]{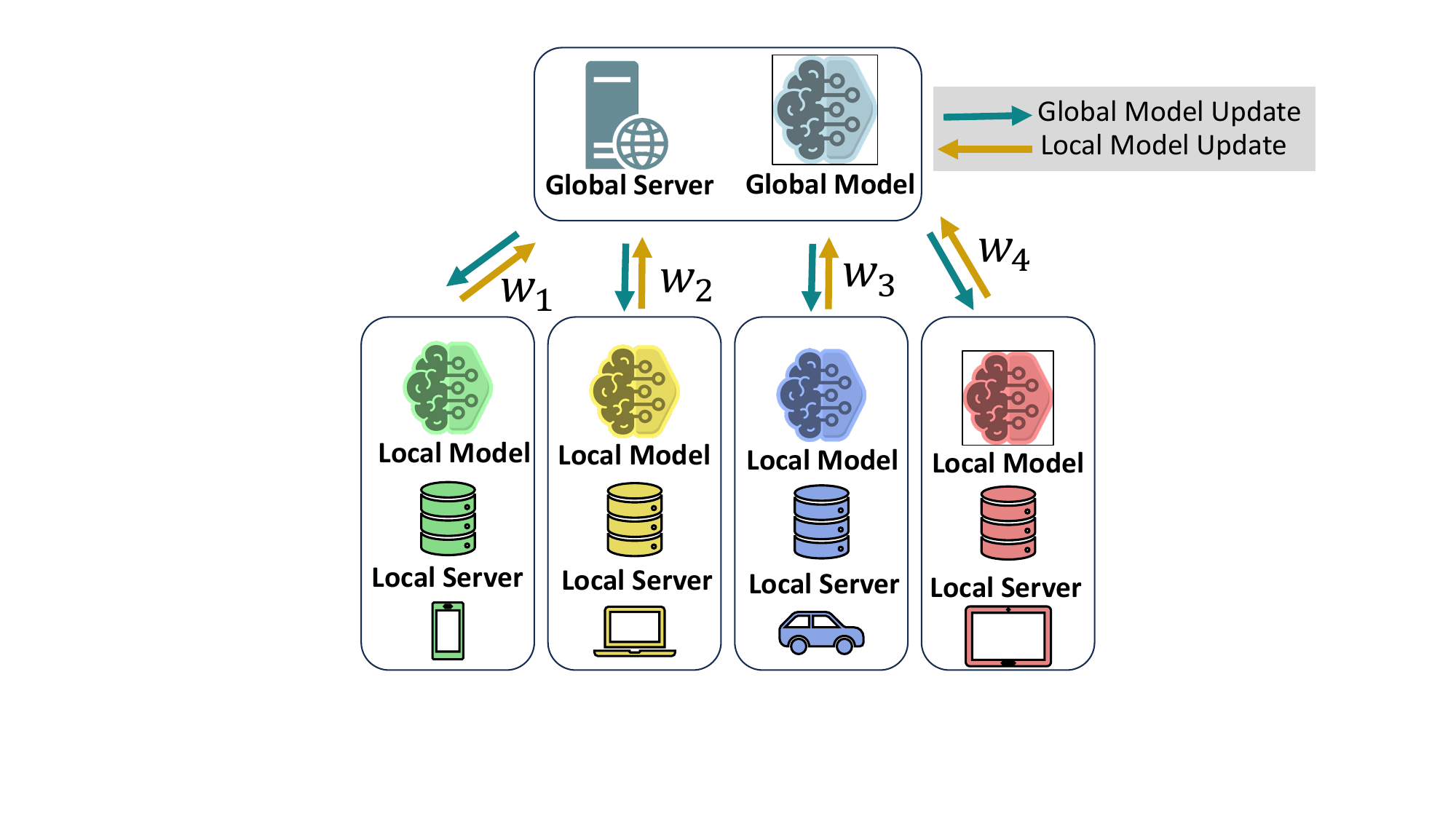}
	\caption{Weighted Federated Averaging \label{overflow}}
	\label{WFL}
\end{figure}

In recent times, concerns about communication costs and data security have sparked debate over designing a dependable central database system for local data aggregation in FL. To address this challenge, an FL model integrated with dynamic weighted averaging was proposed in \cite{chen2021federated}. In this approach, models are trained using local data, and subsequently, the model updates are transmitted to a central server for aggregation. Subsequently, the updated global model shares the latest updates with the participants while ensuring data privacy. Their dynamic weighted averaging model meticulously addresses the imbalance of distributed data and effectively mitigates the influence of numerous local updates (see Fig. \ref{WFL}). 

The conventional FedAvg methods utilized in FL overlook the significant domain shifts among different FL clients, diminishing their performance and applicability. To address this issue, a federated transfer learning algorithm incorporating discrepancy-based weighted federated averaging (D-WFA) was employed in \cite{chen2022federated}. This algorithm receives locally labeled source domain samples and unlabeled target domain samples to update local models with generalization capability. To achieve this objective, they devised an MMD-based dynamic weighted averaging algorithm to aggregate the updated local models, accounting for the domain change and adjusting the weights accordingly. However, in real-world industrial implementations, the domain transitions between training clients (between the target participant and source, or among several source participants themselves) are expected to be frequent. Thus, low-quality data from certain clients could deteriorate the performance of the global model if the clients were simply aggregated with a common weight. To optimize the matching of the diagnostic task, a weighted algorithm should be devised to aid the server in identifying good or bad clients. Therefore, the primary objective of D-WFA in \cite{chen2022federated} is to penalize (assign lower weight to) clients predicted to make a modest contribution, while rewarding (assigning higher weight to) clients expected to make a substantial contribution to the target global model. An MMD-based dynamic weighted technique was developed to measure such assistance, prompted by the MMD distance.

They assumed that $N$ clients are participating in the FL. Their proposed D-WFA consists of seven stages that must be followed when a new global training epoch begins  \cite{chen2022federated}. The steps of their algorithms are as follows:\\
a) First, the source clients receive the model $w_{G,t-1}$ (a global aggregated model of the $(t-1)^{th}$ round).\\
b) The distributed model is trained using various local data for numerous local epochs $E_{local}$, yielding various client models, where $k$ represents the $k^{th}$ training client.\\
c) The updated $w_{k,t}$ calculates the source feature vectors and yields the outcome of the final layer in the feature harvester with $f^{S}_{k,t}$. Once again uploaded to the server, $w_{k,t}$ and $f^{S}_{k,t}$ were then both present.\\
d) The target client downloaded the parameters $w_{k,t}$ and $f^{S}_{k,t}$. The unlabeled target client data is used to compute the target feature vectors $f^{T}_{k,t}$. Then, $f^{T}_{k,t}$ is also returned to the server.\\
e) The MMD distances, $MMD_i$ are computed with $f^{S}_{k,t}$ and $f^{T}_{k,t}$.

\begin{equation}
	L_{m}(x^{S}_{i},x^{T}_{i})=\norm{\frac{1}{n^{S}}\sum_{i=1}^{n^{s}}\varphi(x^{S}_{i})-\frac{1}{n^{T}}\sum_{j=1}^{n^{T}}\varphi(x^{T}_{j})}.
\end{equation}

f)The N MMD distances are turned into the weight vector as follows:
\begin{equation}
	\alpha_{k,t}=\frac{\frac{1}{MMM_{k,t}}}{\sum_{n=1}^{N} \frac{1}{MMD_{k,t}}}.
\end{equation}
where: $\sum_{n=1}^{N} \alpha_{k,t} = 1$\\
g) The final step is to aggregate the client's local models with the specified weight, as follows:
\begin{equation}
w_{G,t}=\sum_{n=1}^{N} \alpha_{k,t} w_{k,t}.
\end{equation}

Notably, the MMD distance computed in step f is considered as a quantitative measure of the degree of deep representational similarity in distribution between the target clients and the source. As the distance decreases, the degree of resemblance increases, aligning with federated training's objective of reducing domain inconsistency. Hence, to reward the successful optimization of model parameters, a particular source client in this round must receive greater weight if its data distribution is more similar to the target client after the local training phase (indicating a lower MMD). Conversely, if a client's MMD from the target client remains substantial after local training, that client must be assigned less weight. The Pseoudocode of their proposed method is represented in Algorithm \ref{WEFL}. The local training period and the global training round are denoted by the variables $E_{local}$ and $E_{global}$, respectively.
The batch size of the local model and learning rate are, respectively, B and $\eta$ (See Algorithm (\ref{WEFL}) in \cite{chen2022federated}).\\
\begin{algorithm}[t]
	1:\textbf{Input}: $\mathbf{X}^{k,S}, \mathbf{Y}^{S}, \mathbf{X}^{T}, E_{\text{global}}, E_{\text{local}}, \beta, B$ \\
	2:\textbf{Server executes:} \\
	3: \quad Initialize $\omega_{G,0}$ \\
	4: \quad \textbf{For} each global epoch $t$ from 1 to $E_{\text{global}}$ \textbf{do:} \\
	5: \quad \quad \textbf{For} each client $k$ \textbf{in parallel do:} \\
	6: \quad \quad \quad $\omega_{k,t}, f_{k,t}^{s} \leftarrow$ Client Local Train $(\omega_{G,t-1}, k)$ \\
	7:\quad \quad \quad $f_{k,t}^{T} \leftarrow$ Target Client Send Feature $(\omega_{k,t}, \mathbf{X}^{T})$ \\
	8:\quad \quad \quad $MMD_{k,t} \leftarrow \left\| \frac{1}{n^{s}}\sum_{i=1}^{n^{s}}\varphi (f_{i,k,t}^{s} - \frac{1}{n^{T}}\sum_{j=1}^{n^{T}}\varphi (f_{i,k,t}^{T}) \right\|_{H}$ \\
	9: \quad \quad \quad $\alpha_{k,t} \leftarrow \frac{\frac{1}{MMD_{k,t}}}{\sum_{k=1}^{N}\frac{1}{MMD_{k,t}}}$ \\
	10 \quad \quad \quad \textbf{$\omega_{G,t}$} $\leftarrow \sum_{k=1}^{N} \alpha_{k,t} \omega_{k,t}$ \\
	11:\textbf{Client Local Train} $(\omega_{G,t-1}, k)$: \\
	12:\quad $\omega_{k,t-1} \leftarrow \omega_{G,t-1}$ \\
	13:\quad \textbf{For} each local epoch $i$ from 1 to $E_{\text{local}}$ \textbf{do:} \\
	14:\quad \quad \textbf{For} batch \textbf{do:} \\
	15:\quad \quad \quad $\omega_{k,t} \leftarrow \omega_{k,t-1} - \eta \nabla_1 (\omega_{k,t-1}; b_{k,t-1})$ \\
	16:\quad  $f_{k,t}^{S} \leftarrow F_{\omega_{k,t}}(\mathbf{x}^{k,S})$ \\
	17:\quad \quad return $\omega_{k,t}, f_{k,t}^{S}$ to the server \\
	18:\textbf{Target Client Send Feature} $(\omega_{k,t}, \mathbf{X}^T):$ \\
	19:\quad $f_{k,t}^{T} \leftarrow F_{\omega_{k,t}}(\mathbf{X}^{T})$ \\
	20:\quad return $f_{k,t}^{T}$ to the server \\
	21:\textbf{Output:} Global Model $\omega_{\text{global}}$
	\caption{Weighted Federated Learning}
	\label{WEFL}
\end{algorithm}
Distinguishing between the internal parameters of different client models is crucial for the performance of FL-based systems. In line with this perspective, a novel parameter-wise elastic weighted averaging aggregation method was introduced to manage the integration of heterogeneous local models \cite{bai2022fedewa}. Every local model evaluates the importance of its model's internal parameters and estimates their essence coefficients accordingly. The central server utilizes these coefficients to perform parameter-wise weighted averaging for global model aggregation.
Current researches have primarily concentrated on validating the direct summation of updates received from local clients, neglecting the weighted average aggregation. 
To address this gap, the authors in \cite{yang2022efficient} proposed a secure and efficient FL method that employs verifiable weighted average aggregation of the global model. Their approach encrypts local updates and data sizes to ensure the protection of updates during model aggregation, with security evaluations confirming this safeguard. Their approach also introduces a verifiable aggregation tag and an efficient verification scheme to confirm the weighted average aggregation. 


\subsubsection{Adaptive Federated Averaging}
\begin{figure*}[t]
	\centering
	\includegraphics[width=0.95\linewidth]{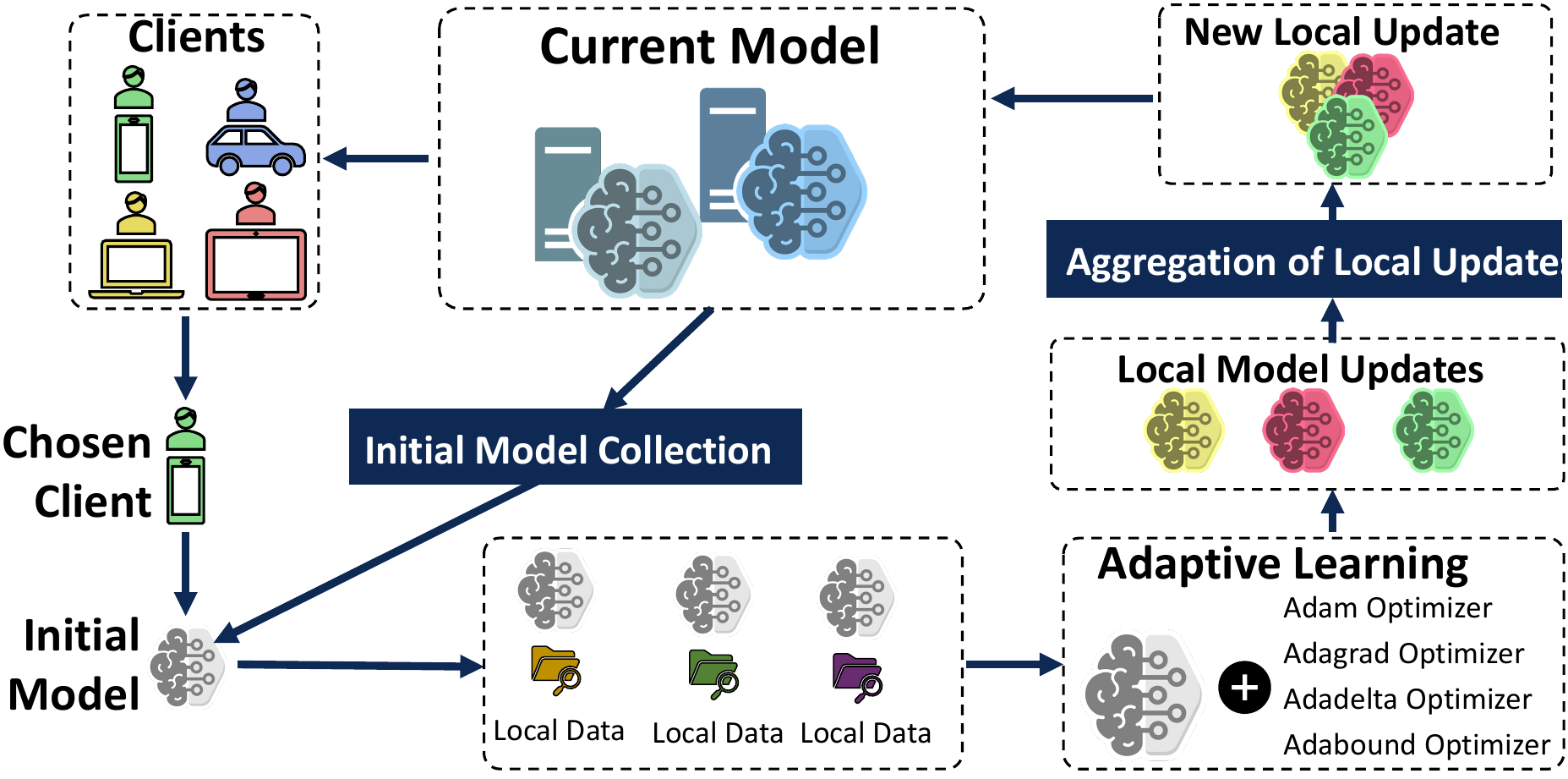}
	\caption{Adaptive Federated Averaging}
	\label{AFL}
\end{figure*}

Traditional federated optimization methods, like FedAvg, often encounter significant challenges related to convergence behavior and tuning issues. Hence, researchers have focused on implementing adaptive optimization schemes that demonstrate improved performance compared to traditional methods in non-federated environments (see Fig.\ref{AFL}).
For example, Reddi \textit{et al.} \cite{reddi2020adaptive} introduced federated versions of popular adaptive optimizers, namely ADAGRAD, ADAM, and YOGI. To address general non-convex problems, their analysis concentrates on examining the convergence behavior of adaptive optimizers in heterogeneous datasets, where the clients' training data may vary in distribution, size, or quality. They also delved into the intricate interplay between client heterogeneity and communication efficiency within federated learning. Diverse data across clients can pose challenges in achieving optimal convergence, because the model must still generalize effectively across various datasets adapt to new labels. Additionally, communication efficiency is crucial because model updates need to be transmitted between the central server (cloud) and clients, considering limitations in latency and bandwidth resources.
Two primary concerns in federated learning are the absence of adaptivity in stochastic gradient descent (SGD)-based model updates and the substantial transmission overhead resulting from frequent server-client synchronization. To address both of these concerns, various approaches have been developed, including gradient compression and quantization techniques, to mitigate communication costs by transmitting compressed gradients between the central server and clients. Additionally, federated versions of adaptive optimizers, such as FedAdam, have been employed to improve the adaptability of model updates.
In \cite{wang2022communication}, Wang \textit{et al.} demonstrated a new method called FedCAMS (Communication-Efficient Adaptive FL) to solve these issues. FedCAMS is proposed as a novel approach that combines communication efficiency and adaptivity within the context of FL. One of the key contributions of FedCAMS is its ability to reduce communication overhead, thereby enhancing the overall communication efficiency of the FL process. This is achieved through various techniques including gradient compression, quantization, and other communication optimization strategies.\\
In \cite{munoz2019byzantine}, the authors employed Adaptive Federated Averaging, which is designed to identify and mitigate failures, attacks, and problematic updates in collaborative model training process. To achieve this, the authors utilized the Hidden Markov Model to model and learn the quality of model updates from each client. This model ensures the reliability and trustworthiness of updates by identifying and filtering out harmful or malicious contributions at each training iteration. Their approach stands out from traditional robust FL models by proposing a robust aggregation rule that identifies and discards undesirable updates, thus ensuring the maturity of the collaborative model. Additionally, they introduced a new protocol aimed at filtering out unimportant clients, which enhances communication efficiency and reduces computational load, thereby making the approach more practical. In \cite{poli2022adaptive}, the Adaptive FL (AdaFed) method integrated two critical scenarios to improve FL. Initially, it dynamically allocates weights to the local models during the averaging process based on their local performance. Higher weights are assigned to more accurate models, indicating the significant contribution of these participants who share better updates and improves global model performance. The adaptive weighting algorithm effectively handle variations in the trustworthiness and capabilities of different participants.
Secondly, AdaFed assesses and adjustes the loss function at each communication period based on the training behavior monitored by the participants. With this method, AdaFed accounts for the concerns and properties associated with the specific training data distribution encountered in each communication period. Their adaptive loss function at each round helps their method mitigate unbalanced data distributions, contributing to their resilience against malicious users.\\
The authors in \cite{jayaram2022adaptive} scrutinized traditional adaptive averaging methods, including peer-to-peer, publish-subscribe, and stream processing research. They explained that each method mentioned above has its deficiency in terms of communication cost and resource usage for FL aggregation. To address these problems, they suggested AdaFed, which leverages serverless and cloud-based operations to achieve adaptive and cost-effective aggregation. Their system model enables dynamic deployment of aggregation only on demand. Their systen scalability allowed each client to join and leave the setting while satisfying fault tolerance from the aggregation operator's side. They highlighted that AdaFed not only substantially minimizes resource requirements and communication costs but also has a minimal impact on aggregation delay. In the following, a prototype implementation is carried out according to Ray, ensuring their system's scalability to thousands of clients and achieving a more than 90$\%$ reduction in communication costs and resource consumption.
In their implementation, they considered two FL deployment scenarios: cross-device and cross-silo. In the former scenario, a few clients with extensive processing capabilities contribute to tumor and COVID diagnosis and detection tasks. In the latter scenario, many clients with limited resources, such as mobile phones and IoT devices, collaborate with small amounts of data. These clients are less reliable, asynchronous, and susceptible to periodic joining and leaving. In \cite{wu2022adaptive}, Wu \textit{et al.} combined adaptive gradient descent and DP approaches tailored explicitly for multi-party collaborative modeling frameworks in the FL process. The adaptive learning rate approach is used to adjust the gradient descent process, preventing model from overfitting and fluctuations, thereby enhancing model performance in multi-party computational frameworks. Additionally, the DP mechanism is proposed to safeguard their system against various unauthorized users and malicious servers, ensuring a privacy-preserving environment. Their emphasis on the federated adaptive (FedAdp) learning rate gradient descent approach sets their work apart from existing studies, as it delves deeper into reducing the model's sensitivity to privacy concerns and hyperparameters. Hyperparameters like the learning rate and the number of iterations are crucial to be specified before training the ML systems. These parameters govern how the model learns and how many times it iterates over the dataset during training. Adjusting these hyperparameters can significantly impact the model's performance and convergence. The fixed learning rate in traditional gradient descent methods like SGD can often impede convergence and lead to suboptimal solutions. In response to the limitations of SGD, the authors proposed the Fadam technique in \cite{wu2022adaptive}, aiming to merging FL with an adaptive gradient descent algorithm. Fadam utilizes first- and second-order momentum based on previous gradients to determine the gradient of the objective function given the parameters such as $\rho^{1}_{t}=\varphi(\beta_{1},\beta_{2}, \cdots,\beta_{t})$, $\rho^{2}_{t}=\mu(\beta_{1},\beta_{2}, \cdots,\beta_{t})$. In each iteration, these parameters are utilized to keep the global model updated as follows \cite{wu2022adaptive}:
\begin{equation}
	\Omega_{t+1}=\Omega_{t}-\frac{1}{\sqrt{\rho^{2}_{t}+\epsilon}}\rho^{1}_{t}.
\end{equation}

The learning rate plays a crucial role in the convergence of the model through gradient descent. Each local model is trained at a specific learning rate, sends parameters to the central server, and reduces the loss function value to achieve gradient descent. All local models receive adjustments from the central server to increase their accuracy and generalizability. By changing the learning rate separately, the authors established adaptive gradient descent methods to prohibit model overfitting.

During the Fadam PROCESS, care must be taken when employing Stochastic Gradient Descent (SGD). SGD could reach a minimal value, but it operated more slowly than other algorithms, and this issue may trap it at saddle points for non-convex functions. To address this problem, the authors in \cite{wu2022adaptive} developed separate adaptive learning rates for various parameters using an optimization strategy that differs from the conventional gradient descent methodology. This approach technically computes the gradient's first-order momentum and second-order momentum estimations. Their proposed scheme not only addresses non-steady state issues of the function but also maintaines the adaptive gradient algorithm's (AdaGrad) performance advantage over the root mean square propagation algorithm and gradient sparse datasets.
Furthermore, the second-order momentum in Fadam is measured across a fixed time window. This poses a challenge in finding the best solution during the modeling training process because the data used for training may lose information if the time window changes.
To address the mentioned challenges, the Adabound method is technically integrated into the FL process. It utilizes a technique called Fadabound to achieve faster learning speed in the early stage and higher generalization ability in the latter stage. In the Algorithm (\ref{adabfl}), the Fadabound process is presented \cite{wu2022adaptive}.
\begin{algorithm}[t]
	\caption{Adaptive FL Algorithm}

1:\textbf{Input}:Dataset, privacy budget  $\epsilon$, learning rate $\eta$\\
2:\textbf{Output}:$\Psi.$\\
3:Begin set $\rho_{0}^{1} = 0, \rho_{0}^{2} = 0.$\\
4: \quad \textbf{For} each round of iterations \textbf{do:}\\
5:  \quad \quad   Gradient descent at time step t: $\beta_{t}\leftarrow  \Delta _{\Psi }f_{t}(\Psi_{t-1})$\\
6:   \quad \quad  Calculate the first-order mom: $\rho_{t}^{1}\leftarrow x_{1}^{t}\rho_{t-1}^{1}+(1 - \\ 
7: \quad \quad x_{1}^{t})\beta_{t}$\\  
8:   \quad \quad  Calculate the second-order mom: $\rho_{t}^{2}\leftarrow x_{2}\rho_{t-1}^{2}+(1 - \\
9: \quad \quad x_{2})\beta_{t}^{2}$\\ 
10:  \quad \quad   Clip learning rates by Clip: $(a/\sqrt(\rho_{t}^{2} ,\eta _{l},\eta _{u})$\\
11:   \quad \quad  Update the first-order mom estimation:\\ 
12: \quad \quad $\bar\rho_{t}^{1}=\frac{\rho_{t}^{1}}{1 - x_{1}^{t}}$\\
13:  \quad \quad   Update the second-order mom estimation:\\ 
14: \quad \quad$\bar\rho_{t}^{2}=\frac{\rho_{t}^{2}}{1 - x_{2}^{t}}$\\
15:  \quad \quad   Until the model converges for the local dataset\\
16:  \quad Send the model parameters  $\Psi_{t}$, of this iteration to the\\ 
17: \quad central server\\
18:  \quad Central server calculates the contribution $\Psi^{i}-\Psi^{i-1}$ of\\ 19: \quad the current iteration and delivers it\\
20:   \quad \textbf{End}.\\
21: \textbf{Return} resulting parameters $\Psi_{1}^{i},\Psi_{2}^{i},\cdots,\Psi_{k}^{i}$ from clients \\
22: to the central server.

 \label{adabfl}
\end{algorithm}
$ \eta_{l}$ and $ \eta_{u}$ are the learning rate’s lower and upper bound, respectively, and t denotes the number of iterations.
The authors in \cite{wang2019adaptive} primarily conducted a deep theoretical examination of the convergence bound for gradient-descent-based FL, assuming N i.i.d. data distributions along with a random number of local parameters. In their study, the authors explored an under-examined convergence bound and employed a control method to dynamically adjust the oscillation of global aggregation in real-time, aiming at reducing learning loss under a constant resource budget. 

\subsubsection{Momentum Federated Averaging}
\begin{figure}[t]
	\centering
	\includegraphics[width=\columnwidth]{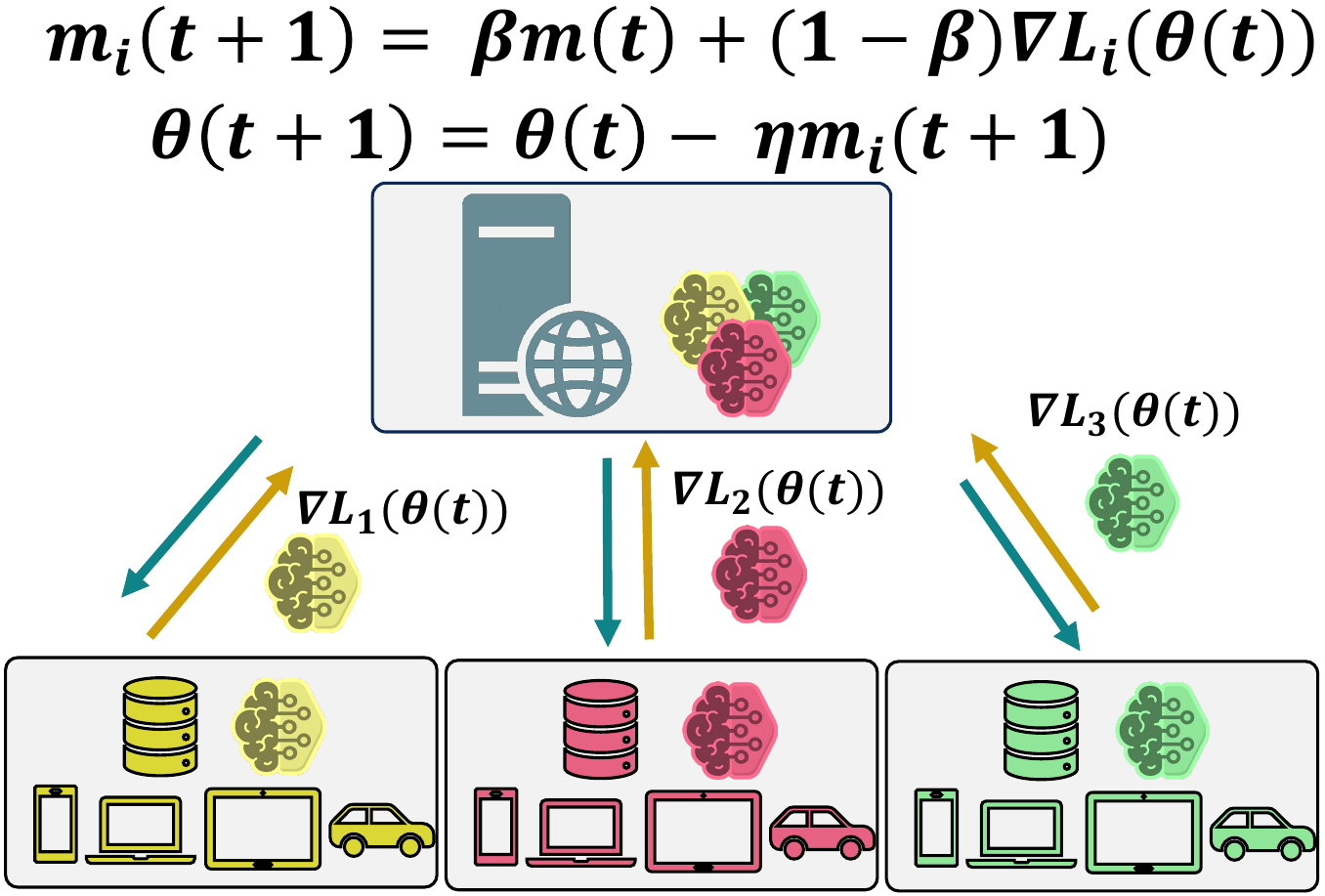}
	\caption{Momentum Federated Averaging}
	\label{MFL}
\end{figure}

One common optimization technique in SGD is momentum-federated learning. This method involves incorporating a fraction of the previous gradient into the current gradient update, which accelerates the convergence of the learning process. This enhances the optimization process by addressing small oscillations and speeding up convergence towards the optimal outcome. The primary objective of momentum federated learning is to integrate the concept of momentum into federated learning. Similar to SGD, momentum federated learning takes into account the previous local gradient direction along with the current local gradient when updating the local model on each node Fig.\ref{MFL}. In this figure, $m(t)$ and $\beta$ are momentum functions and momentum parameters, respectively.

The authors in \cite{wang2023boosting} adopted a new method by combining model personalization and client-variance-reduction to upgrade the semi-supervised FL (SSFL) structure. However, one main concern of the SSFL is the interaction between participants' heterogeneity and label deficiency, which deteriorates their negative impacts. Conventional methodologies designed for supervised FL are found to be inapplicable to SSFL, thereby diminishing its efficacy in addressing pertinent concerns. In response, the researchers formulated the problem by incorporating pseudo-labeling and model interpolation techniques. They introduced a novel approach called FedCPSL to mitigate participant and data heterogeneity, leveraging momentum-based client variance reduction alongside normalized aggregation averaging and other averaging strategies. The study assessed the robustness of FedCPSL against participant and data diversity by examining its convergence properties, revealing a sublinear convergence rate. Furthermore, FedCPSL is anticipated to exhibit sublinear convergence rates for nonconvex objectives, thereby establishing enhanced bounds and ensuring sustainability among participant and data heterogeneity. 

Another similar method is adopted in \cite{das2022faster} where the authors offered a new FL scheme called federated global and local momentum (FedGLOMO) to tackle the problems of data heterogeneity among participants, and compressed data communication. FedGLOMO reaches a better convergence rate with a complexity of $O(\epsilon^{-1.5})$ for smooth non-convex problems, compared to the complexity of previous schemes, which is $O(\epsilon^{-2})$. FedGLOMO's key contribution lies in its ability to alleviate noise in local participant-level stochastic gradients and reduce high variance during the central server aggregation phase. This is accomplished through the introduction of a variance-reducing momentum-based general update scheme at the central server, alongside a variance-reduced local update scheme at the local participants. By combining these schemes, FedGLOMO effectively tackles participant drift in environments characterized by heterogeneous data distribution, resulting in notable improvements in communication performance. 

Kim \textit{et al.} \cite{kim2022communication} introduced a novel algorithm named Federated Learning with Acceleration of Global Momentum (FedAGM), aimed at enhancing the convergence and resilience of FL methods to effectively address client heterogeneity and low participation rates. FedAGM expedites the propagation of a swiftly estimated model derived from the global gradient, ensuring the continuous updating of local gradients while reinforcing the resilience of the central server-side aggregation framework. To achieve this objective, the method incorporates the momentum of global gradient data into both client and server weights, thereby bridging the disparity between local and global losses and yielding comparable task-specific efficiency with fewer communication rounds. Furthermore, FedAGM is integrated as a regularization term within the objective function of participants to enhance the alignment between local and global gradients. Notably, FedAGM exhibits equivalent memory and communication overhead, thus maintaining consistency with low-participation and large-scale FL settings. 

\begin{algorithm}[t]
{1:Initialize global model: $\Psi_t$, number of rounds: $t$, Momentum\\
2: parameter: $\tau$, Momentum update: $\rho_t$ ($\rho_0 = 0$),Initialize 3:learning rate: $\eta _l$, batches: B, number of epochs:
	E.}\\
4:\textbf{For}{ each round of iterations t \textbf{do:}}\\
5: \quad	Select $\chi_t = $ random subset of clients $\chi$ ($|\chi_t| < |\chi|$)\;\\
4: \quad	\textbf{For} each client $x \in \chi_t$ \textbf{do:}\\
5:	\quad \quad	 \textbf{For} each local epoch i from 1 to E \textbf{do:}\\
6:	\quad \quad \quad \textbf{For} {batch $b \in B$ \textbf{do:}}{\\
7:	\quad  \quad \quad	\quad	$\Psi_{x}^{t+1} = \Psi_{t} -\eta_{l}\Delta l_x(\Psi_{t},b)
		$}\\
8:	\quad  \quad \quad	 \textbf{end For}\\
9:  \quad \quad	 \textbf{end For}\\
10: \quad  \textbf{end For}\\
11:	\quad	$\nu =  \sum_{x=1}^{|\chi_t|} \frac{n_x}{n} (\Psi_{x}^{t+1} - \Psi_t)$\;\\
12:	\quad	$\rho_{t+1} = \tau \rho_t + (1 - \tau)\nu$;\\
13:	\quad	$\Psi_{t+1} = \Psi_t + \rho_{t+1}$;\\
14:\textbf{end For}
	\caption{Momentum Federated Learning Algorithm}
	\label{Momfl}
\end{algorithm}
To further address the issue of data heterogeneity in FL, a novel paradigm known as Clustered Federated Learning based on Momentum Gradient Descent (CFL-MGD) was introduced to enhance the convergence rate of FL techniques by integrating cluster and momentum strategies. Comparative assessments were conducted against established methodologies such as k-means clustering, cosine distance-based approaches, and user-clustered algorithms. In CFL-MGD, participants with identical learning tasks are grouped into clusters according to their data characteristics. Subsequently, each participant within a cluster utilize its local data to update model weights employing momentum gradient descent for local parameter adjustments. The approach further incorporates gradient averaging and model averaging for global aggregation \cite{zhao2023clustered}. 

An additional concern in FL pertains to the presence of data heterogeneity among participants, leading to potential discrimination against unprivileged clusters presumed to store sensitive properties. To fill this gap, Salazar \textit{et al.} in \cite{salazar2023fair} introduced a novel fairness-aware FL approach termed FAIR-FATE, aimed at prioritizing equitable models during global aggregation while maintaining superior utility. FAIR-FATE entails a fairness-aware aggregation algorithm that accounts for individual participants' fairness in global model aggregation through the utilization of a fair Momentum term. The fair momentum term effectively mitigates fluctuations associated with non-fair gradients, establishing their method as pioneering within ML for integrating a fair Momentum estimate to achieve fairness objectives. FAIR-FATE addresses fairness concerns in FL by guiding participant collaboration to construct equitable models while safeguarding data privacy. By employing momentum gradient descent, FAIR-FATE overcomes oscillations arising from noisy gradients through the utilization of exponentially weighted averages. This renders it faster than DP methods as it offers superior gradient approximation.\\
The Federated Averaging with Standard Momentum (FedMom \cite{salazar2023fair}) method is represented by the algorithm \ref{Momfl}.
The local weight is computed by subtracting $\Omega_{t}$ from received $\Omega^{k}_{t+1}$. The local computed weights are averaged in the next step to build the global model weights, $\alpha$. In the following, a summation of the former weights, $\rho_{t}$, is used to obtain the Momentum update, $\rho_{t+1}$, and the global model weight, $\alpha$. The authors defined the momentum parameter $\tau$, to govern the impact of preceding weights. Consequently, the previous model is condensed to facilitate the update of the global model incorporating the momentum weight. Within the server, upon receiving each $\Psi^{k}_{t+1}$, the local update is computed by subtracting $\Psi{t}$. Subsequently, these local updates are aggregated to generate the global model update, denoted as $\nu$. Following this, the Momentum update $\rho_{t+1}$, is determined by adding a fraction of the previous update $\rho_{t}$ to the global model update $\nu$. The parameter $\tau$ controls the extent to which the previous update contributes. Ultimately, the global model is refreshed by adding the previous model to the momentum update. In the Algorithm (\ref{Momfl}), the Fadabound process is elaborated \cite{salazar2023fair}.
\begin{figure}[t]
	\centering
	\includegraphics[width=\columnwidth]{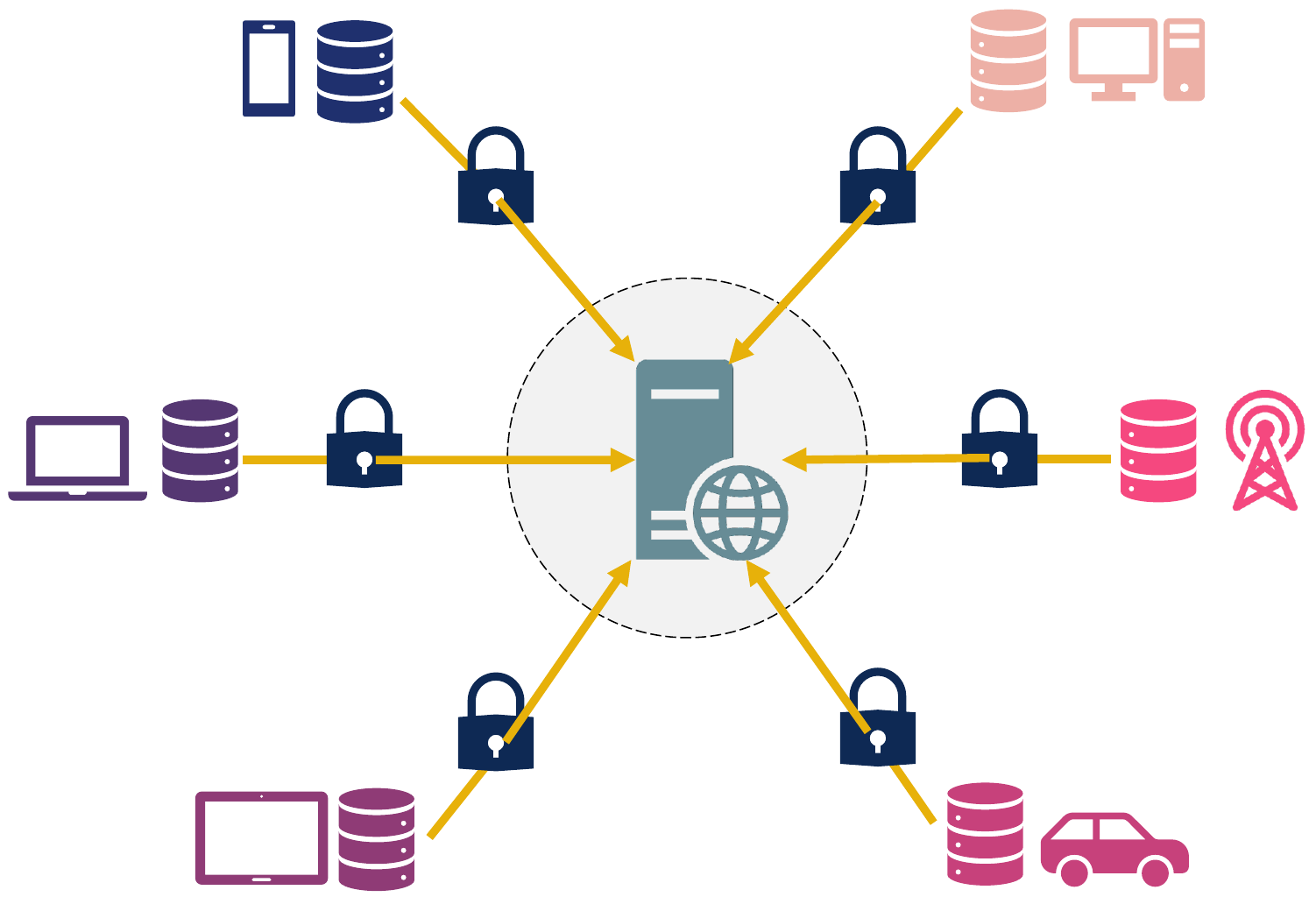}
	\caption{Security Federated Averaging }
	\label{SFL}
\end{figure}
\subsubsection{Secure Federated Averaging}
Among aggregation methods, security averaging is a cryptographic approach adopted within FL to combine client updates securely, thus safeguarding the confidentiality of the model. It provides the following benefits:

\begin{itemize}
    \item Preventing the server from learning the value and source of individual model updates.
    \item Protecting the FL system against inference and data attribution attacks. 
    \item Compensating for the lack of parameter validation in the FL.
    
\end{itemize}
Cryptographic primitives and fully homomorphic encryption (FHE) offer partial solutions for enhancing the security of FL systems by protecting sensitive information. Nonetheless, both methodologies encounter limitations and scalability challenges. In response, FL acts as an alternative approach, enabling clients to collaboratively train a shared neural network without the need to disclose their local data Fig. \ref{SFL}. The integration of secure aggregation (SA) mechanisms has emerged as a key strategy to enhance the security of FL. Recognized for its resilience against gradient inversion and inference attacks, SA facilitates clients in securely computing the sum of their private parameters. By concealing the source and location information of aggregated data, SA plays a pivotal role in preserving client privacy. 
However, the study detailed in \cite{pasquini2022eluding} reveals the susceptibility of SA methods to exploitation by malicious servers, leading to potential breaches in privacy within the context of FL. These adversaries possess the capability to manipulate model updates, devising a novel attack strategy termed model inconsistency, wherein various clients employ divergent interpretations of a shared model. Despite the implementation of SA, this inconsistency makes it possible for servers to extract sensitive information pertaining to users' privacy and datasets. To elucidate this vulnerability, the authors conducted a thorough investigation and deployed two distinct attack methodologies, highlighting the substantial threat posed by the model inconsistency attack vector \cite{pasquini2022eluding}. These attacks underscore how malicious servers can compromise the security provided by existing SA techniques, enabling the identification of individual model updates from aggregated data and their attribution to specific users regardless of the number of participants involved. Moreover, the authors devised several strategies to propose solutions that integrate with current SA techniques, ensuring performance and utility are preserved while addressing the vulnerability stemming from model inconsistency.

Another investigation aimed at enhancing privacy-preserving systems by leveraging secure multiparty computation (MPC) for confident aggregation of model updates from clients \cite{bonawitz2017practical}. Bonawitz \textit{et al.} introduced an SA method in \cite{bonawitz2017practical}, ensuring that no party exposes their model update or sensitive parameters to the global aggregator or any other third party. The SA primitive facilitates the private fusion of models' outputs locally to update a global model. This approach offers significant advantages, enabling users to share updates with the assurance that the service provider can only access the averaged updates post-aggregation. The authors primarily targeted mobile devices in resource-constrained environments where dropouts are prevalent. In contrast to transmitting the parameter vector in plain text, their adopted model aims to incur less communication overhead. Additionally, their mechanism proves sustainable for dropout users at any connection, a consideration overlooked by previous works \cite{bonawitz2017practical}.\\
Combining DP with SA forms the foundation of an end-to-end privacy-preserving system within FL. SA streamlines the consolidation of client weights without revealing or modifying any individual client weight. DP algorithms introduce noise into client weights to prevent trained local and global models from disclosing updates regarding the training data. However, traditional SA approaches have been characterized by their complexity and resource consumption. To shed more light on this topic, Stevens \textit{et al.} in \cite{stevens2022efficient} adopted the FLDP mechanism, which uses DP to provide precise, flexible, and effective FL without trusting on edge or cloud servers. The security of their scheme is based on the learning with errors (LWE) problem, where the noise mixed with DP also serves as the noise term in LWE. A main innovation of their study is offering a new algorithm that utilized DP to fulfill secure model aggregation, which, significantly reduces computational and communication overhead. In other words, their method significantly reduces the communications expansion factor and diminishes the server's computation complexity. FLDP integrates the discrete Gaussian distribution and gradient clipping to ensure robust computational DP with a secure model aggregation. Their designed model obtains a higher degree of accuracy compared to central-model training methods under differentially private DL and enhances efficiency and flexibility. In \cite{li2020secure}, another application of DP was demonstrated through the introduction of a secure FedAvg approach that incorporates Gaussian noise into shared updates. The authors' theoretical analysis showcases that their methodology achieves an $\mathcal{O}\left(\frac{1}{T}\right)$ convergence rate for local model parameters, where $T$ represents the total number of SGD updates. Their analysis comprehensively investigates the relationship between achievable privacy levels and various system parameters, including mini-batch size, local epoch duration, and the number of randomly selected clients, by using the amplification privacy theorem. To manage system complexity, they proposed a balanced approach that optimizes the convergence rate of their design while considering the selected parameters. Lastly, they delved into the ramifications of different approach parameters on the communication efficiency of their methodology. The secure FedAvg with standard security design is represented by the algorithm (\ref{secure}) \cite{li2020secure}. 

\begin{algorithm}[t]
	1: \textbf{Input}: Initial model $\bar{w}_0$ and step size $\eta_0$. The PS broadcasts $\bar{w}_0$ to all clients ($S_0 = [N]$).\\
	2: \textbf{for} $t = 0, 1, ..., T - 1$ \textbf{do:}\\
	3: \quad \textbf{Client side}:\\
	4: \quad \textbf{for} $k \in S_t$ in parallel \textbf{do:}\\
	5: \quad \quad \textbf{if} $\text{mod}(t,Q) = 0$ \textbf{then:}\\
	6: \quad \quad \quad Set $w_k^t = \bar{w}_t$.\\
	7: \quad \quad \textbf{end if}\\
	8: \quad \quad Sample a mini-batch $\xi_k^t$ from $D_k$ and calculate the\\ 		
	9: \quad \quad local gradient $g_k^t = \nabla F_k(w_k^t; \xi_k^t, b)$.\\
	10: \quad \quad \textbf{if} $\text{mod}(t + 1, Q) \neq 0$ \textbf{then:}\\
	11: \quad \quad \quad $w_k^{t+1} \leftarrow w_k^t - \eta_t g_k^t$.\\
	12: \quad \quad \textbf{else if} $\text{mod}(t + 1, Q) = 0$ \textbf{then:}\\
	13: \quad \quad \quad $w_k^{t+1} \leftarrow (w_k^t - \eta_t g_k^t) + z_{kt}, \quad z_{kt} \sim N(0, \sigma^2_{t,k}I_M)$.\\
	14: \quad \quad \quad Send $w_k^{t+1}$ to the PS.\\
	15: \quad \quad \textbf{end if}\\
	16: \quad \textbf{end for}\\
	17: \quad \textbf{for} $k \notin S_t$ in parallel \textbf{do:}\\
	18: \quad \quad $w_k^{t+1} = w_k^t$.\\
	19: \quad \textbf{end for}\\
	20: \quad \textbf{Server side:}\\
	21: \quad \textbf{if} $\text{mod}(t + 1, Q) = 0$ \textbf{then:}\\
	22: \quad \quad $\bar{w}_{t+1} = \frac{N}{K}\sum_{k \in S_t} p_k w_k^{t+1}$.\\
	23: \quad \quad Select a subset of clients $S_{t+1}$ by sampling without-\\
        24:\quad \quad replacement, and broadcast $\bar{w}_{t+1}$ to all clients.\\
	25: \quad \textbf{end if}\\
	26: \textbf{end for}\\
	\caption{Secure Federated Learning}
		\label{secure}
\end{algorithm}
Traditional approaches of SA have predominantly concentrated on securing privacy within singular training iterations. However, they inadvertently neglect significant privacy breaches that may occur across multiple rounds, often due to assumptions related to partial client selection. Addressing this challenge, the authors in \cite{so2023securing} devised a secure model aggregation protocol primarily focused on safeguarding client privacy across multiple consecutive training rounds. Introducing a novel metric, they aimed at upholding privacy assurances in FL secure aggregation throughout these rounds. Additionally, they implemented a systematic client selection mechanism termed Multi-RoundSecAgg, which substantially ensures the long-term privacy of each FL client while maintaining fairness and enabling an optimal average number of participants per round. In this context, Kim \textit{et al.} in \cite{kim2023cluster} introduced an innovative solution named Cluster-based Secure Aggregation (CSA), which addresses the challenge of managing dropout nodes while simultaneously optimizing computational and communication costs within FL scenarios. Their approach is tailored for FL environments characterized by the presence of heterogeneous devices with varying computational capacities and defferent sizes of training datasets distributed across diverse geographical locations. Within their framework, the CSA mechanism groupes clients based on their response times which are determined by factors such as communication delay and local computing duration. They utilized a grid-based clustering technique to cluster clients according to their similarity in processing capabilities and GPS coordinates, enabling the edge or cloud server to estimate maximum latency for each cluster and accurately identify dropout nodes. Within each cluster, intermediate summations are conducted for aggregation, and a novel additive sharing-based masking mechanism was devised to protect the authentic local client weights during SA. This mechanism allows for the removal of dropout nodes without relying on $(t, n)$ threshold factors, ensuring the confidentiality of weights and data even in the event of dropout node exposure. Furthermore, their mechanism incorporates mask verification, which makes FL clients to publicly validate the accuracy and integrity of provided masks through a discrete logarithm problem.


\subsubsection{Compressed and Quantized Federated Averaging}
\begin{figure*}[t]
	\centering
	\includegraphics[width=0.8\textwidth]{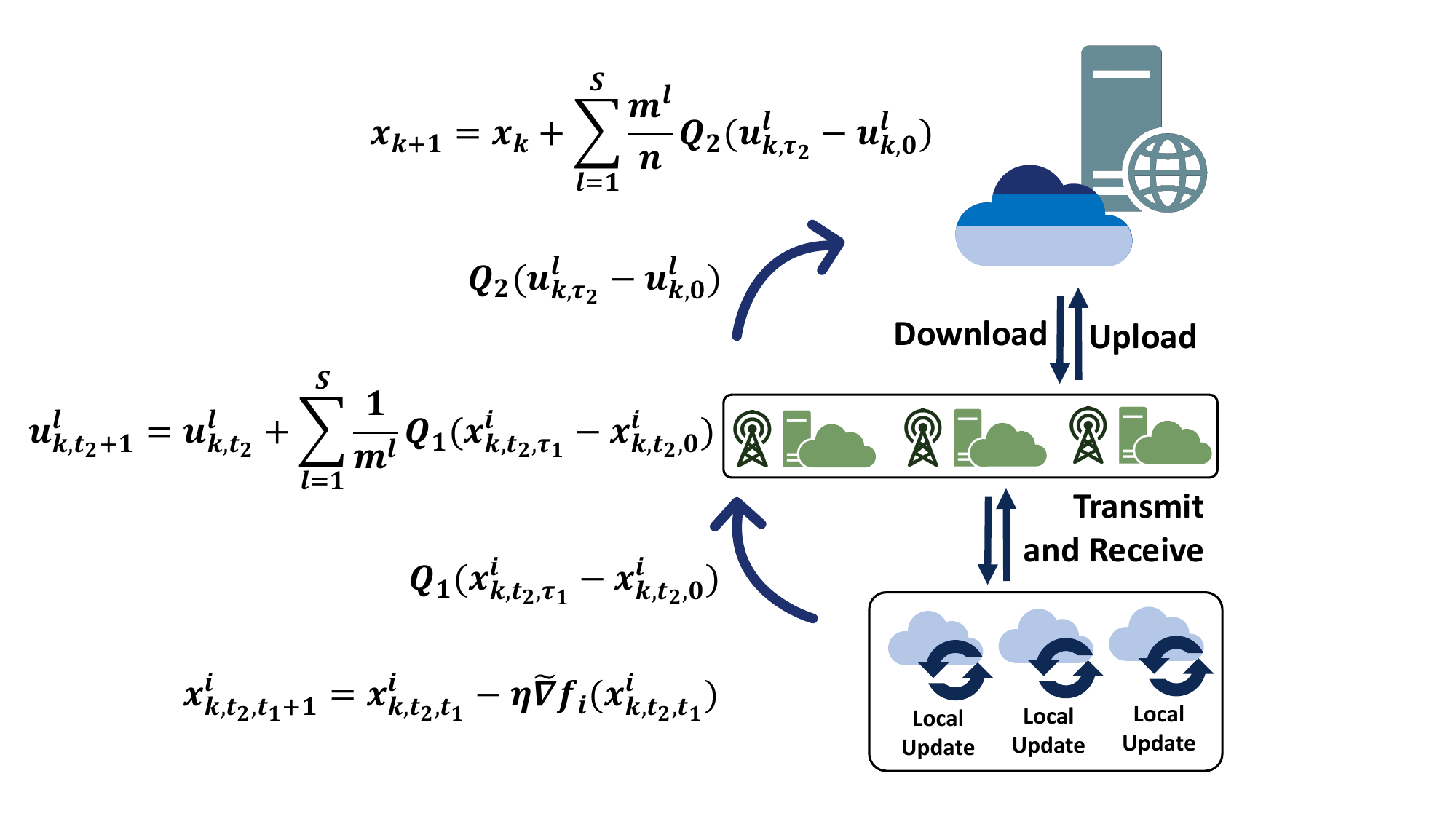}
	\caption{Compressed and Quantization Federated Averaging}
	\label{CQFL}
\end{figure*}

Blockchain technology has emerged as a distributed framework to uphold privacy in FL systems. Although, existing blockchain methodologies encounter several constraints, particularly pertaining to scalability and communication expenses within expansive networks Fig. \ref{CQFL}. To mitigate the drawbacks highlighted earlier, the authors in \cite{kang2022communication} pioneered a cross-chain framework that aims at creating flexible and scalable infrastructures in the realm of AIoT. Drawing upon pertinent blockchain methodologies, this framework was devised to address specific challenges within FL, ensuring both security and performance. Notably, the adoption of a cross-chain approach facilitates secure collaboration and interaction across diverse blockchain networks. Furthermore, the implementation of a compression technique for model updating enables the efficient management of communication costs without compromising the precision of the system. Additionally, the utilization of ML-based auctions for dynamic pricing in model training was proposed as an effective strategy. In \cite{lang2023joint}, an alternative method known as compressed averaging was explored, with Joint Privacy Enhancement and Quantization (JoPEQ) at its core, aimed at addressing communication efficiency and privacy concerns within FL. JoPEQ integrates privacy enhancement and lossy compression algorithms by leveraging vector quantization based on the random lattice, a suitable compression technique. This approach yields statistically equivalent additive noise as a result of distortion, that is utilized to enhance privacy by introducing a dedicated privacy-preserving noise model involving multiple variable quantities into the model updates. The study demonstrates that JoPEQ achieves concurrent data quantization at an optimal bit rate while ensuring an appropriate level of privacy without compromising the utility and accuracy of the global model. 

Consistent with this notion, communication in FL manifests in two distinct scenarios: downlink and uplink transmissions. During uplink transmission, clients dispatch their updated weights to the central server, often precipitating a more pronounced bottleneck compared to downlink transmission, which proceeds in the opposite direction. This discrepancy is typically attributed to the constrained upload bandwidth relative to download bandwidth and the imperative to aggregate weights from myriad clients. Researchers suggested compressing uplink transmission as a fundamental strategy to address this issue. To delve deeper, a prevalent adaptive quantization approach entails the utilization of lossy quantization coupled with optional lossless compression. Adaptive quantization dynamically adjusts the quantization rate and highlights asymmetries like differences in training duration and client contributions based on local dataset sizes to mitigate transmission expenses.
For instance, dynamic adaptations to the quantization level yield significant improvements in compression and quantization efficiency without compromising the fidelity of the trained model. In \cite{honig2022dadaquant}, the authors introduced a doubly adaptive quantization method (DAdaQuant) that dynamically adjustes the quantization volume among different clients over time, consistently enhancing client-server compression. This approach yields improvements of up to 2.8 times compared to non-adaptive baselines in the context of the FL process. The client-adaptive quantization technique involves assigning a minimal quantization rate to FL clients, with the expected variance of quantized parameters serving as a quantization error metric. This approach effectively reduces the volume of data transmitted from FL clients to the central server (cloud) while controlling the quantization error.\\
\begin{algorithm}[t]
	\caption{ Quantization Federated Learning Algorithm}
	1: initialize the model on the cloud server $X_0:$\\
	2: \textbf{for} k=0,1,$\cdots$ ,K-1 \textbf{do}:\\
	3: \ \textbf{for} $\imath =1 , \cdots,s$ edge servers in parallel \textbf{do}:\\
	4: \  \  \  \ Set the edge model same as the cloud server model:\\
	5: \  \   \  \ $u_{k,0}^{\imath}=X_k:$\\
	6: \  \  \  \  \textbf{for} $t_2=0,1, \cdots ,\tau_2-1$ \textbf{do}:\\
	7:  \  \  \  \  \ \textbf{for} i \ in $C^{\imath}$ clients in parallel \textbf{do}:\\
	8:\  \  \  \  \  \  \  \  \ Set the client model same as the associated edge\\ 
	9:\  \  \  \  \  \  \  \  \ server model:$x_{k,t_2,0}^{i}=u_{k,t_2}^{\imath} :$  \\
	10:	\  \  \  \  \  \  \   \  \ \textbf{for} $t_1=0,1,\cdots,\tau_1-1$ \textbf{do}:\\
	11:\  \  \  \ \   \  \   \  \ 	\ \ \ \ $x_{k,t_2,t_1+1}^{i}=x_{k,t_2,t_1}^{i}-\eta \nabla f_i(x_{k,t_2,t_1}^{i})$\\
	12:\  \  \  \  \  \  \   \  \	\textbf{end}\\
	13:	\  \  \  \  \  \  \ Send $Q_1(x_{k,t_2,\tau_1}^{i}-x_{k,t_2,0}^{i})$ to its associated edge server:\\
	14: \  \  \   \   \textbf{end}\\
	15:	\  \  \  \  \  Edge server aggregates the quantized updates from the clients:\\
	16:\ \ \ \  \ $u_{k,t_2+1}^{\imath}=u_{k,t_2}^{\imath}+\frac{1}{m^{\imath}}\sum_{i \in D^{\imath}}^{}Q_1(x_{k,t_2,\tau_1}^{i}-x_{k,t_2,0}^{i})$\\
	17 :\  \ \ \textbf{end}\\
	18:\ \ \  \ Send $Q_2(u_{k,\tau_2}^{\imath}-u_{k,0}^{\imath})$\\
	19:\ \ 	\textbf{end}\\
	20:\ \ Cloud server aggregates the quantized updates from the edge servers:\\
	21:\ \ $x_{k+1}=x_{k}+\sum_{\imath=1}^{s}\frac{m^{\imath}}{n}Q_2(u_{k,T_2}^{\imath}-u_{k,0}^{\imath})$\\
	22: \textbf{end}
	\label{QFLA}
\end{algorithm}
Yongjeong \textit{et al.} in \cite{oh2022communication} proposed a communication-efficient FL framework named FedQCS which leverages quantized compressed sensing. This framework addresses concerns regarding communication costs and transmission overhead while maintaining the accuracy of gradient communication. FedQCS integrates dimensional reduction, sequential block sparsification, and quantization blocks for gradient compression. Through the utilization of quantization and dimension reduction techniques, this framework achieves higher compression ratios compared to conventional one-bit gradient compression methods. To facilitate aggregation of local updates from compressed signals, the authors introduced an approximate Minimum Mean Square Error (MMSE) algorithm for gradient reconstruction employing the Expectation-Maximization Generalized Approximate-Message-Passing (EM-GAMP) scheme. Furthermore, the framework incorporates a low-complexity gradient reconstruction method based on the Bussgang theorem. 
To mathematically analyze the quantization method, the authors in \cite{liu2022hierarchical} discussed a hierarchical FL called Hier-Local-QSGD in which one cloud server, $P$ edge servers, and $N$ clients are considered. ${D}^{l}_{i}$ $i=1,\cdots, M$ denoted the distributed training datasets. Their algorithm has two phases. The first one is frequent Edge Aggregation and Infrequent Cloud Aggregation, where periodic aggregation is considered as an effective tool for communication costs. A big aggregation time slot $T$ leads to a small communication round while reducing the system's performance. This is because if local models undergo too many steps of local SGD updates, they begin to come close to the local loss function $h_{i}(x)$'s optima rather than the global loss function $h(x)$. To this end, before the cloud aggregation, each edge server effectively aggregates the models in its immediate proximity many times. Each edge server averages the models of its clients after every $T_{1}$ local SGD update on each client. Then, the cloud server averages all edge servers' models after each $T_{2}$ edge aggregation. Thus, communication with the cloud occurrs after every $T_{1} T_{2}$ local updates. In contrast to FedAvg with an aggregation interval of $T=T_{1} T_{2}$, the local model is thus less likely to be skewed towards its local minima. The second stage of Hier-Local-QSGD, Quantized Model Updates phase, plays a crucial role in determining the communication volume and overall communication cost in FL. During this phase, the size of the DL model is routinely reduced using quantization techniques to minimize communication overhead. However, employing a low-precision quantizer significantly decreases communication costs while introducing additional noise into the FL process, thereby diminishing the performance of the updated model.
The Hier-Local-QSGD process is presented in the Algorithm (\ref{QFLA}) \cite{liu2022hierarchical}. 
\begin{figure*}[t]
	\centering
	\includegraphics[width=0.85\textwidth]{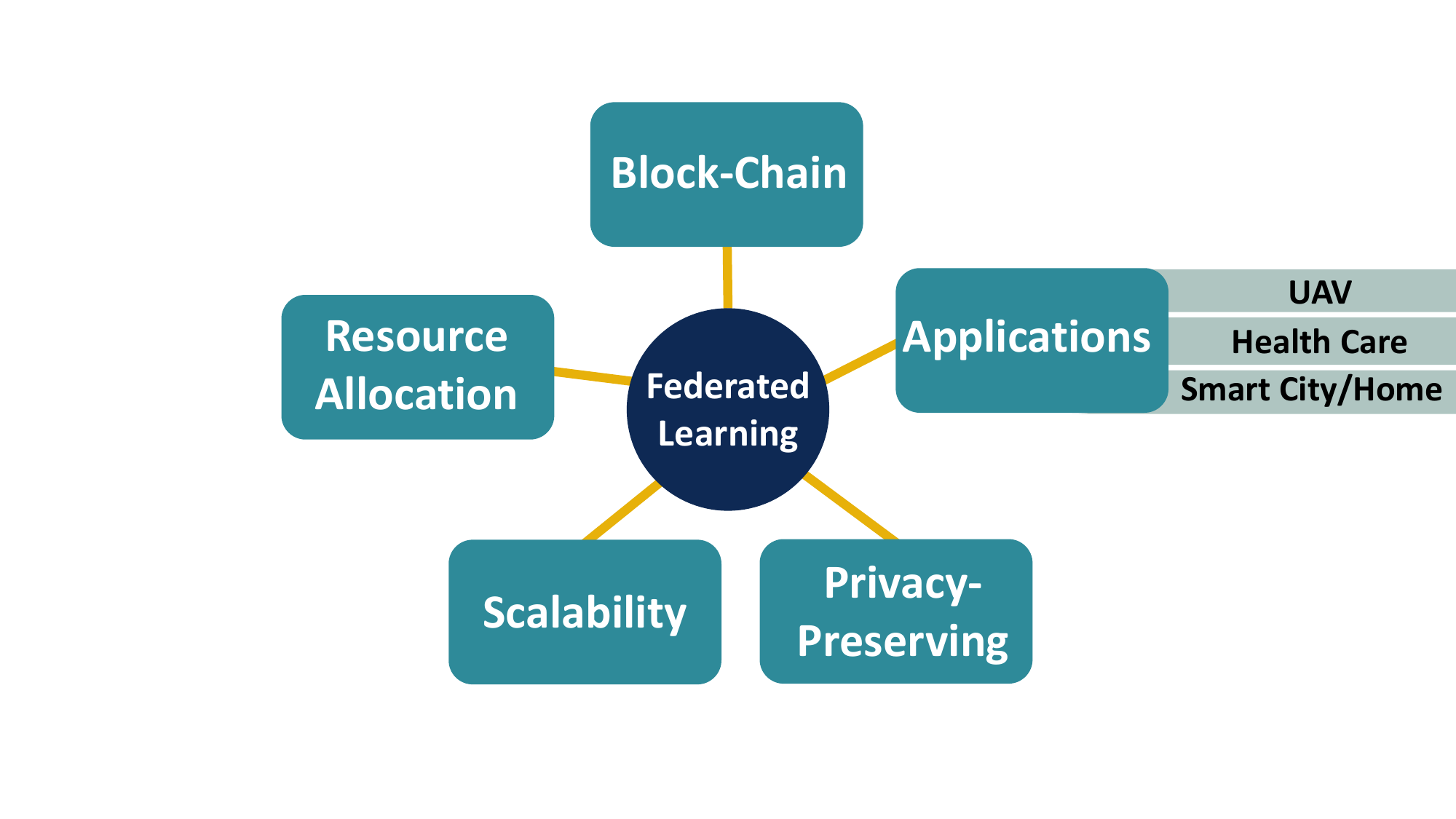}
	\caption{Federated Learning; Challenges and Applications}
\end{figure*}
\section{Security in FL} \label{sec:sec}
Security in FL refers to the protection of sensitive data, models, and communications throughout the FL process. It involves safeguarding against various threats and vulnerabilities that may compromise the privacy, integrity, and confidentiality of FL systems. In the subsequent two sections, our attention is directed towards the methods and strategies utilized to safeguard the confidentiality of data and models within FL frameworks.


\subsection{Block-Chain}
The IIoT facilitates the acquisition of confidential data through a multitude of smart devices, with the analysis of this data holding promise for guiding decision-making across various tiers of operations. Leveraging FL, model updates are communicated across IIoT networks rather than the actual private data, providing a mechanism for evaluating the accumulated data while upholding user privacy. Although, vulnerabilities persist within the FL framework, as malicious actors possess the capability to disrupt model update transmissions.


The authors in \cite{wang2019survey} conducted an examination of Blockchain's application in addressing various data security issues within the IoT landscape. They explored how the extensive presence of IoT devices, coupled with challenges such as constrained computing capabilities, limits communication bandwidth, and makes unreliable radio links, therefore, it influences the effectiveness of Blockchain. The latest advancements in Blockchain technologies were thoroughly assessed, and a comparative analysis was conducted to evaluate their suitability for IoT contexts. Additionally, the study identified areas for further research aimed at enhancing the capacity, security, and scalability of Blockchain systems to facilitate their seamless integration with IoT technologies in the future.

The article \cite{sameera2024privacy} reviewed research efforts to develop privacy solutions in scenarios using Blockchain-Enabled FL. It covers background information on FL and blockchain, evaluates integration architectures, identifies key attacks, and suggests countermeasures for privacy assurance. Additionally, it explores successful application scenarios for blockchain-enabled FL. The survey aims at assisting both academic and industry practitioners in understanding available theories and techniques to improve FL performance via Blockchain for privacy protection, while also highlighting main challenges and future directions in this emerging field.

The authors in \cite{heo2024blockchain} introduced a secure data processing framework leveraging blockchain technology and DP. The system operates by generating requested information for service providers from urban computing data via ML algorithms. To safeguard privacy, the data undergoes DP measures. However, recurrent queries may compromise privacy protection under standard DP protocols. To address this, the authors implemented a strategy to mitigate total privacy costs by reusing noise for identical data and privacy parameters through blockchain mechanisms. Although, the introduction of noisy data during training could potentially reduce machine learning accuracy. Therefore, they devised a method to enhance accuracy by storing and effectively utilizing model parameters derived from the same data within the blockchain infrastructure.
Zhang \textit{et al.} \cite{zhang2024decentralized} outlined a standard decentralized FL framework leveraging blockchain technology, and delineated its operational process and its utilization across domains like the IoT, and IoV. They systematically outlined the challenges encountered within this framework and assessed the proposed solutions aimed at mitigating these challenges. Lastly, they offered perspectives on potential avenues for future research endeavours in this domain.
The authors in \cite{wang2020bsv} introduced a blockchain-driven system called Blockchain-based Special Vehicles Priority Access Guarantee Scheme (BSV-PAGS) that integrates blockchain technology into IoT domain. This system establishes a decentralized network of edge nodes based on blockchain, enabling seamless information sharing among them and efficient data processing. Special vehicle information is transmitted to blockchain-based edge nodes by intelligent terminal devices, then shared across all nodes via smart contracts. This consequently facilitates collaborative information exchange. Additionally, they used Support Vector Machines (SVM) to improve special vehicle detection accuracy. A dedicated smart contract ensures real-time scheduling to prioritize special vehicles.


In \cite{wei2022redactable}, the authors presented a distinctive chameleon hash algorithm featuring a configurable trapdoor (CHCT), specifically tailored to ensure secure FL within IIoT contexts. A comprehensive security analysis of CHCT system was conducted, aiming at the development of a redactable medical blockchain (RMB) that integrates the CHCT concept. Zhang \textit{et al.} in \cite{zhang2021bc} introduced a secure data transfer mechanism leveraging the advantages of EC, FL, and the unique characteristics of blockchain technology. To enhance learning efficiency, they initially discriminated the local model updating procedure from the mobile device-independent process. Subsequently, an edge server is incorporated to delegate the majority of computations to the server, thereby optimizing computational resources. Finally, they adopted a distributed blockchain architecture to further improve security and decentralization. 

Assuring FL security and optimizing performance mitigates the impact of malicious nodes on model training, while simultaneously encouraging reliable nodes to engage in the learning process. In \cite{xu2021besifl}, the authors introduced the Blockchain Empowered Secure and Incentive FL (BESIFL) paradigm as a significant contribution to the domain. BESIFL leverages blockchain technology to establish a fully decentralized FL framework, wherein streamlined approaches for incentive administration and identification of malicious nodes are integrated. Zhang \textit{et al.} in \cite{zhang2022blockchain} initially addressed the need for data security and intelligent computation offloading within Power IoT (PIoT) frameworks by introducing a blockchain and AI-driven architecture termed as safe cloud-edge-end collaboration PIoT (BASE-PIoT). In \cite{liao2021blockchain}, the authors introduced a novel architecture named space-assisted PIoT (SPIoT) to tackle security concerns and latency challenges within IoT-based networks. SPIoT leverages a satellite in low earth orbit (LEO) to broadcast consensus messages, thereby reducing block formation delays. To address the long-term security constraint and minimize overall queuing time, they developed the blockchain and the semi-distributed learning-based secure and low-latency computation offloading method (BRACE). Using Lyapunov optimization, they initially separated server-side computational resource allocation and task offloading. Subsequently, their proposed Federated Deep Actor-Critic (FDAC) technique is employed to address the latency issue. Finally, Lagrange optimization and smooth approximation methods are applied to optimize resource allocation.

Zhang \textit{et al.} in \cite{zhang2020blockchain} examined and devised an architecture for FL systems that integrates blockchain technology to enhance IIoT failure detection that ensures the authenticated integrity of client data. Under this architecture, each client generates a Merkle tree at regular intervals, storing the tree root on a blockchain, and representing client data records as leaf nodes. Additionally, they introduced a new technique called centroid distance weighted federated averaging (CDW-FedAvg), that tackles data heterogeneity in IIoT failure detection by considering the distance between positive and negative classes in each client dataset. Furthermore, they devised an innovative contract-based incentive mechanism to incentivize user participation in FL. This mechanism relies on the volume and centroid distance of user data used for local model training. Furthermore, they introduced an inventive contract-based incentive mechanism aimed at encouraging user participation in FL. This mechanism is based on the volume and centroid distance of user data utilized for local model training.
The authors in \cite{zhou2020cefl} delved into strategies for synchronizing the edge and the cloud, with the aim of enhancing the overall cost-effectiveness of FL. Leveraging Lyapunov optimization theory, they developed and evaluated a cost-effective optimization framework called CEFL. This framework enables near-optimal control decisions for dynamically incoming training data samples which are generated from tasks like admission control, load balancing, data scheduling, and online accuracy tuning. CEFL control framework is adaptable and allows for the integration of diverse design decisions and practical FL requirements. For instance, it enables the utilization of cost-effective cloud resources for model training which improves cost efficiency while facilitates on-demand privacy preservation.
To enable the integration of security-focused offloading algorithms, Qu \textit{et al.} \cite{qu2021chainfl} introduced a novel simulation platform called ChainFL. ChainFL is designed to establish an EC environment among IoT devices while remaining compatible with FL and blockchain technologies. It boosts interoperability and lightweight characteristics and facilitates swift connectivity among devices with diverse architectural styles to configure complex network setups.
Additionally, due to its distributed architecture and integrated blockchain functionality, ChainFL functions as an FL platform across diverse devices, ensuring robust security. By integrating an offloading-decision model into the platform and deploying it within an IIoT environment, they showcased the adaptability and efficacy of ChainFL.

To anticipate cached files, the authors in \cite{cui2020creat} devised an innovative compressed algorithm called CREAT that integrates blockchain assistance with FL compression techniques. In the CREAT approach, each edge node utilizes its local dataset to train a model that extracts user and file attributes to predict commonly accessed files and consequently improves cache efficiency. Through FL, multiple edge nodes engage in model training while preserving data confidentiality without necessitating data sharing. Furthermore, their method facilitates the burden on FL's communication system by enabling gradient compression at the edge nodes to expedite transmission. Additionally, blockchain technology is integrated into the CREAT algorithm to ensure the security of transmitted data. In order to address the needs of decentralized entities, the researchers devised four smart contracts tasked with supervising and authenticating transactions, consequently upholding data security.

 Employing an innovative FL-Block (Federated Learning integrated with blockchain) methodology, Qu \textit{et al.} \cite{qu2020decentralized} devised a scheme to construct a comprehensive learning model on a blockchain infrastructure. This approach empowers local learning updates of end-device transactions while leveraging blockchain's Proof-of-Work consensus protocol to facilitate autonomous ML, ensuring the integrity of the global model without central oversight. Furthermore, the evaluation of FL-Block’s latency performance determines the requisite block production rate to effectively manage communication, consensus delays, and computational resources.
Blockchain for data exchange was studied in \cite{chen2021ds2pm}. After examining blockchain-based data sharing models, they encounter several challenges, including the complex task of safeguarding users' data ownership rights, privacy, and integrity. Another significant concern is the absence of a structured mechanism within blockchain to handle diverse types and inconsistent formats of data, leading to substantial storage requirements. Additionally, the consensus process often proves to be inefficient or lacks fairness. In light of these issues, the authors proposed an FL approach to address these challenges and devised a blockchain-based model called Data-Sharing Privacy Protection Model (DS2PM).
The authors in \cite{aloqaily2022towards} introduced a hierarchical Federated Learning (HFL) framework coupled with blockchain technology to facilitate swift, secure, and precise decision-making for industrial machinery. Their approach encompasses a two-stage FL process. Initially, industrial devices are clustered to undergo localized ML training. Subsequently, the resulting local models are distributed to network edge devices, where FL averaging consolidates them into multiple global models. Employing a FL aggregation technique, a primary global model is synthesized from the decentralized first-stage global models in the subsequent phase. Finally, the acquired models are validated and verified on the edges using blockchain.

To complete Cyber-Physical Systems (CPS) tasks, Al \textit{et al.} \cite{al2022intelligent} investigated a collaborative strategy that integrates blockchain technology for the sharing of resources and capabilities. Tailored for Next-Generation Networks, their approach leverages intelligent IoT devices conducive to FL. To address challenges in CPS, they employed a multi-stage clustering blockchain and FL algorithm to generate both local and global models. Initially, local models are formulated within individual clusters. Subsequently, fog devices construct fog models through federated averaging. Finally, Federated Aggregation is utilized to construct a comprehensive global deep model in the cloud. The incorporation of blockchain technology ensures the documentation and authentication of additional models, providing resilience against cyberattacks aimed at tampering with records. 

Given real-time data and EC, the authors in \cite{meese2022bfrt} focused on BFRT, an architecture that merges blockchain and FL to predict internet traffic patterns. With a focus on data security, their framework facilitates distributed model training in real-time at IoT edge nodes. Through the integration of gated recurrent unit (GRU) neural networks and LSTM models, they conducted extensive trials using dynamically captured segments of arterial traffic data. They developed a functional prototype of their permissioned blockchain network on Hyperledger Fabric, subjecting it to rigorous testing utilizing virtual machines simulating edge nodes. Decentralized FL mechanisms typically operate under two specific assumptions. First, they presume that participating clients are trustworthy entities, and incapable of injecting low-quality model updates into the aggregation process. Second, these mechanisms often rely on the client's ability to share local models with other clients or a third party for verification purposes.

Ranathunga \textit{et al.} \cite{ranathunga2022blockchain} presented a blockchain-based decentralized framework tailored for situations where participating organizations inject low-quality model updates and resist exposing their local models for verification. Their innovative framework utilizes a hierarchical network of aggregators, equipped with mechanisms to encourage or penalize organizations based on the quality of their local model updates. Providing flexibility, it ensures that no single entity possesses the aggregated model in any FL training round. The effectiveness of their framework is assessed through two Industry 4.0 use cases: predictive maintenance and product visual inspection, considering both off-chain and on-chain performance metrics. 

Users of mobile or IoT devices are involved in various operations, like encryption, decryption, and mining to complete transactions. These operations require energy and computational resources, which impose limitations on the operational efficiency of users, who typically rely on battery power and have constrained computing capabilities. One possible solution is to offload these tasks to common servers provided by Mobile Edge Computing (MEC) or cloud computing. In this approach, the required steps for transaction processing within the blockchain are viewed as virtual blockchain functions that can be performed on standard servers.
The authors in \cite{taskou2022blockchain} enhanced the Blockchain Function Virtualization (BFV) architecture that allows MEC or cloud computing to perform all blockchain operations virtually. Moreover, they explored the potential applications of the BFV framework and identified resource allocation challenges it presents in mobile networks. Furthermore, they introduced an optimization problem with the objective of simultaneously reducing energy consumption costs and increasing incentives for miners. 

Factory-as-a-Service (FaaS) is a novel approach to quickly adapting the production process to meet the demands of the supply chain and customers. Effective support for FaaS requires flexibility in networking and cloud services. To facilitate the alignment of networking resources between service providers and clients, a 5G Network Slice Broker (NSB) serves as an intermediary, streamlining resource allocation processes. To support FaaS, Hewa \textit{et al.} \cite{hewa2022blockchain} presented a secure NSB solution deployed on a blockchain platform. Collaborating with slice and Security Service Level Agreements (SSLA) managers, their Secure NSB (SNSB) model provides secure, cognitive, and distributed network services for resource allocation and SSLA establishment. They proposed a federated slice selection technique that combines a Stackelberg game model with an RL algorithm to compute real-time and optimal unit pricing within the SNSB framework. 

Through a shared ML model driven by blockchain technology and an FL framework called iFLBC edge, the authors in \cite{doku2020iflbc} investigated a strategy for deploying edge-AI to end nodes that introduces the Proof of Common Interest (PoCI) technique to mitigate the issue of inadequate relevant data. PoCI distinguishes between relevant and irrelevant data, with the former utilized for model training. Upon aggregation with other models to create a shared model, the relevant data is stored on the blockchain. Network participants have access to this aggregated model to deliver edge intelligence to end-users. Intelligence using FL blockchain (iFLBC) network members is enabled to provide end-user intelligence services, thereby expanding the reach of AI.

The decentralized learning approach exposes the entire solution to various types of attacks that could severely compromise the accuracy of the resultant global model. Despite refraining from outsourcing sensitive data to cloud-hosted services, the system distributes the workload for data processing. However, this approach leaves it susceptible to malicious users that can undermine the integrity of the global model. To address this, Esposito \textit{et al.} in \cite{esposito2022attacks}, assessed the impacts of two widely recognized attacks on FL in an objective manner while also proposing a preliminary defense mechanism leveraging blockchain technology. 
Within \cite{xu2023blockchain}, the authors introduced a management scheme designed to tackle diverse forms of malicious conduct within the smart grid using blockchain technology. This approach combines a consortium blockchain with the best-worst multi-criteria decision method (BWM) to precisely quantify and address such behavior. Moreover, the utilization of smart contracts enables the implementation of a penalty mechanism, facilitating the imposition of appropriate penalties on various malicious users. Through a thorough exposition of the proposed algorithm, logic model, and data structure, the authors elucidated the principles and operational flow of this scheme for effectively mitigating malicious behavior.
Another application is Triabase, a distinctive permissioned blockchain database scheme that was examined in \cite{drakatos2021towards} and \cite{drakatos2021triastore}. It abstracts ML models into simplified data blocks stored and retrieved through the blockchain. Positioned as an ML application on the edge server, Triabase avoids extensive data storage on inherently sluggish mediums like blockchains due to the costly verification process. For its consensus mechanism, Triabase leverages the Proof-of-Federated-Learning (PoFL) concept. Integration of the Triabase prototype system into the Hyperledger Fabric blockchain architecture ensures promising initial outcomes.

Decentralized Finance (DeFi) utilizes blockchain, and enables customized execution of predefined transactions among multiple parties that facilitates activities like lending and trading. However, the DeFi has multiple security challenges as evidenced by a decline in total locked value from $\$200$ billion in April 2022 to $\$80$ billion in July 2022. To address this gap, the authors in \cite{li2022survey} presented the first comprehensive analysis of DeFi security. They summarized vulnerabilities across different layers of blockchain technology relevant to DeFi. Then, they delved into vulnerabilities at the application level. Subsequently, they categorized and examined real-world DeFi attacks, aligning them with the identified vulnerabilities and their underlying principles. 

Recently, there has been a notable acceleration in the development of Space–Ground Integrated Networks that is responsible for linking diverse networks across remote areas. This integration allows for the connection of various devices, enabling the utilization of previously inaccessible data for collaborative model training and the emergence of novel service models. However, privacy concerns persist as a significant barrier to data sharing among multiple parties.
The authors in \cite{yin2021blockchain} introduced a collaborative training approach based on blockchain technology that leverages its decentralized ledger to address trust issues among various participants. By utilizing the blockchain's inherent non-repudiation features, the accurate execution of model aggregation is guaranteed. Additionally, they devised a privacy-preserving method based on encryption function, wherein the aggregator only retrieves the aggregated model results without accessing other participants' models uploaded to the blockchain. Subsequently, they developed a prototype system utilizing blockchain to assess the time efficiency of their collaborative training method and function encryption module.
The authors in \cite{cao2021blockchain} developed a token-based access control system for smart contracts that incorporates dynamic adjustments of access control rules (ACRs) to ensure that only authorized users can initiate and execute particular smart contracts. They also introduced an intrusion detection mechanism for smart contracts, enabling the real-time and effective detection of attacks against them. Leveraging these mechanisms, they suggested an integrated framework named ACID (Access Control and Intrusion Detection), that counteracts diverse attacks while preserving all features and capabilities of the underlying blockchain-based SATCOM system.
Guan \textit{et al.} \cite{guan2021bsla} addressed the challenge of private issues in Identity-Based Cryptosystems (IBC) by proposing a Blockchain-assisted Secure and Lightweight Authentication (BSLA) scheme. BSLA integrates blockchain technology to enhance authentication robustness and reliability. In BSLA, a user's private key is derived autonomously by combining partial private keys from multiple Private Key Generators (PKG) mitigating the risk of single PKG-related key exposure. Moreover, the blockchain facilitates the synchronization of user revocation lists across spatial nodes, enabling rapid user revocation detection. Finally, they conducted a security analysis of BSLA and demonstrated its ability to meet diverse security requirements.
The future directions, open areas, and accuracy of references in Block-chain FL-assisted systems are listed in Table \ref{FLBCS} with details.

\begin{table*}[t]
\footnotesize
  \caption{References on FL-based Block-Chain Structures}
  \label{FLBCS}
  \setlength{\aboverulesep}{0.05cm}
  \setlength{\belowrulesep}{0.05cm}
  \renewcommand{\arraystretch}{1.2}
\begin{tabularx}{\textwidth}{cccc>{\raggedright\arraybackslash}m{11.7cm}}
    \specialrule{0.12em}{0pt}{0pt}
    Reference& Year&Acc$>90\%$&AI/ML approach & \multicolumn{1}{c}{Open Areas and Future Challenges and Directions}\\
    \specialrule{0.12em}{0pt}{0pt}
\cite{zhang2021bc}& 2022 &\cmark &\textbf{---}&Assessing the complexity of data transfer and developing a transmission security mechanism to cope up with significant network attacks. 

\\\specialrule{0.0005em}{0pt}{0pt}
\cite{xu2021besifl}& 2021   &\cmark&BESIFL& Improving node-selection method to reduce iterations and considering model-variable gradient protection to increase learning reliability. \\\specialrule{0.0005em}{0pt}{0pt}
\cite{zhang2022blockchain}&2022 &\textbf{---}&FDRL& Addressing Poor Scalability using lightweight hierarchical storage and Security threats with the integration of AI-based secure networks,zero-trust-based identity, and block-chain.
\\\specialrule{0.0005em}{0pt}{0pt}
\cite{liao2021blockchain}&2021 &\textbf{---} &BRACE & DRL with adversary will be examined for their  a Blockchain and semi-distRibuted leArning-based seCure and low-latEncy computation offloading algorithm (BRACE). \\\specialrule{0.0005em}{0pt}{0pt}
\cite{zhang2020blockchain} & 2021 &\cmark& CDW-FedAvg & Improving the trustworthiness of node devices for the aggregation and modifying the model by relocating modules from node servers to node devices with storage facilities and better processing.
\\\specialrule{0.0005em}{0pt}{0pt}
\cite{qu2020decentralized} & 2020 &\cmark&FL-Block& Expanding the model by maximizing the trade-off between efficiency and security and identifying the ideal conditions for computation and communication costs using the MDP and game theory. 
\\\specialrule{0.0005em}{0pt}{0pt}
\cite{aloqaily2022towards}&2022&\cmark&HFL&\textbf{---}
\\\specialrule{0.0005em}{0pt}{0pt}
\cite{al2022intelligent} & 2022 &\xmark&\textbf{---}& Considering the non-cooperative game model in the industrial setting, training variables must be traded in a game theory way to enable distributed knowledge sharing.
\\\specialrule{0.0005em}{0pt}{0pt}
\cite{meese2022bfrt} & 2022 &\cmark&BFRT/LSTM& Studying novel methods for online multi-output prediction, as well as experimenting using multiple blockchain designs to improve the FL process.
\\\specialrule{0.0005em}{0pt}{0pt}	
\cite{ranathunga2022blockchain} & 2022 &\textbf{---}&\textbf{---}& Assessing the structure's resistance against poisonous attacks.
\\	\specialrule{0.0005em}{0pt}{0pt}
\cite{drakatos2021towards} & 2021 &\textbf{---} &DL/Triabase&The data integrity of transactions is protected by using binary hash trees, also known as Merkle trees, and only byte hashes are employed to construct a Merkle path from the root to a given transaction.
\\\specialrule{0.0005em}{0pt}{0pt}
\cite{drakatos2021triastore}& 2021 &\cmark&DL/Triastore& Exploring the performance of Triastore on edge DL datacenter with strong GPU cards and developing the Triastore with pervasive querying layers to conduct experiments with more datasets.
\\\specialrule{0.0005em}{0pt}{0pt}
\cite{li2022survey}& 2022 &\textbf{---}&DL& Exploring attack analysis and monitoring are critical to empower DeFi security using the blockchain.\\
\specialrule{0.12em}{0pt}{0pt}
\end{tabularx}
\end{table*}


\subsection{Privacy Preserving}
Preserving privacy in FL is crucial due to its decentralized nature where multiple parties share sensitive data. Various strategies have been proposed to ensure privacy in FL settings. These strategies hide individual data contributions during the model aggregation process that results in preventing the exposure of sensitive information to unauthorized parties. By implementing these privacy-preserving measures, FL systems elevate trust among participants while mitigating the risk of data breaches or privacy violations. To protect privacy in FL settings multiple techniques including DP, SA, Homomorphic Encryption, and ML-based techniques have been proposed.

Existing literatures have identified various attack methods, like membership inference that exploits vulnerabilities in ML models and coordinating servers to extract private data. Therefore, FL requires additional safeguards to ensure data privacy. Moreover, processing big data often exceeds the capabilities of standard computing resources. The authors in \cite{chamikara2021privacy} tackled these challenges by introducing a distributed perturbation algorithm called DISTPAB that is designed for preserving privacy in horizontally partitioned data. DISTPAB addresses computational limitations by distributing the privacy preservation task across a distributed environment, and utilizing various resource capacities ranging from resource-constrained devices to high-performance computers.

Zhang \textit{et al.} \cite{zhang2021adaptive} presented the AdaFed strategy that brings together several shipping agents to construct a model by exchanging model parameters without the risk of data leakage for defect diagnosis in IoS. They first utilized two typical activities as examples to show that a tiny portion of the model parameters can disclose the shipping agents' unprocessed information. Based on this, the Paillier-based communication system is created to protect the shipping agents' original data. Additionally, a control technique is presented to adaptively alter the model aggregation interval throughout the training process to minimize the costs associated with cryptographic computation and communication.\\
To guarantee the context-aware privacy of task offloading, Xu \textit{et al.} in \cite{xu2022c} presented C-fDRL, a model to enable context-aware federated deep reinforcement learning (FDRL). Their proposed method consists of three stages; CloudAI, EdgeAI, and DeviceAI. C-fDRL examines whether the privacy of low context-aware data is distributed at the EdgeAI and high context-aware data is preserved locally at the DeviceAI with the work being offloaded. To this end, C-fDRL employs a context-aware data management technique (i.e. CDMA) to decouple the context-aware (privacy) data from the offloading tasks if a user requests to offload the data. This facilitates the implementation of a novel scheduling approach termed as a context-aware multi-level scheduler. This scheduler segregates the context-aware data from the job for local processing. To perform the computational process before the actual job execution, CDMA sets high context-aware data at local devices and low context-aware data at the edge device.

The authors in \cite{tayyab2023comprehensive} performed an extensive investigation on the security and privacy implications inherent in Deep Learning (DL) algorithms. Their examination elaborates on the applications and challenges intrinsic to these algorithms, and introduces attacks and relevant taxonomy. They delved into countermeasures against prominent attacks like poisoning, evasion, model extraction, and model inversion. They also outlined various privacy-preserving methods to ensure the confidentiality of datasets. 
Hussain \textit{et al.} in \cite{hussain2024measuring,hussain2023survey} investigated potential backdoor signals in code models by analyzing model parameters such as attention weights, biases, activation values, and context embeddings in both clean and poisoned CodeBERT models. While activation values and context embeddings shows noticeable patterns in poisoned samples, attention weights and biases do not exhibit significant differences. Additionally, the authors explored literature on Explainable AI and Safe AI to understand the poisoning of neural models of code, establishing a taxonomy for Trojan AI for code and presenting an aspect-based classification of triggers. They also highlighted recent works and state-of-the-art poisoning strategies that manipulate such models, offering insights for future research in the area of Trojan AI for code.

By introducing a secure cryptographic framework based on hash functions and Homomorphic Encryption, they proposed a resilient approach to dataset security. The authors in \cite{neto2022collaborative} studied a collaborative DDoS detection and classification strategy that leverages FL for dispersed and multi-tenant IoT settings. Through collaborative endeavors, numerous tenants enhance their proficiency in detecting and categorizing DDoS attacks across all edge nodes while upholding data privacy. After training DL instances on locally scaled traffic data, tenants exchange model parameters with each other. This proposed method enhances the safety of IoT operations and holds promise for diverse applications.
In \cite{ferrag2022edge}, Ferrag \textit{et al.} explored the development of the Edge-IIoT-set, an innovative and comprehensive cybersecurity dataset tailored for IoT and IIoT applications. . Edge-IIoT-set facilitates ML-based intrusion detection systems in both centralized and FL modes. This dataset encompasses a diverse collection of IoT/IIoT testbed devices, sensors, protocols, and configurations for EC environments. It consists of data generated by over ten IoT devices, featuring various sensors like digital sensors for temperature and humidity, ultrasonic sensors, water level detection sensors, pH sensor meters, soil moisture sensors, heart rate sensors and flame sensors. The authors identified and evaluated fourteen cyberattacks associated with IoT and IIoT communication protocols which can be categorized into five threat categories: DoS/DDoS attacks, information-gathering attacks, man-in-the-middle attacks, infiltration attacks, and malware attacks. Additionally, they extracted features from multiple data sources including alerts, system resources, logs, and network traffic, resulting in a dataset containing 1176 distinct features out of which 61 features exhibiting strong correlations. 

To cope with unreliable users, the authors in \cite{li2021efficient} proposed an efficient privacy-preserving federated learning (EPPFL) method. Their approach ensures that the target model receives updates from high-quality data sources which result in mitigating the adverse effects of unreliable users. Utilizing Excluding Irrelevant Components and Weighted Aggregation mechanisms, FL model achieves rapid convergence with minimal communication and processing overheads. Consequently, this approach maximizes both model accuracy and training efficiency. Furthermore, they established a secure framework leveraging the threshold Paillier cryptosystem to robustly protect all user-related sensitive data throughout the training process.
Song \textit{et al.} \cite{song2020fda} introduced  a strategy for federated defense called FDA3 that integrates defensive competencies against adversarial instances sourced from multiple outlets. Their proposed cloud-based architecture facilitates the dissemination of protective capabilities against assorted attacks across IIoT devices by utilizing FL.

In \cite{liu2021federated}, Liu \textit{et al.} suggested an FL system tailored for privacy preservation that integrates asynchronous updating to mitigate the challenges arising from vehicle heterogeneity. This system maintains the confidentiality and privacy of vehicular training data. Furthermore, their methodology incorporates dynamic temporal weights based on the computational and communicative capabilities of individual vehicles to fully exploit pre-trained local models. Typically, the conventional weighted average approaches on the server side factors in the number of samples. Departure from these methods could lead to reduced network traffic and enhanced learning efficiency. Consequently, it augments security and privacy measures. 

In healthcare domain, the conventional training model's susceptibility to fraud within the decentralized Internet of Medical Things (IoMT) poses a significant challenge for research endeavors. In response, an FL-based blockchain-enabled task scheduling framework (FL-BETS) was devised in \cite{lakhan2022federated} that integrates various dynamic heuristics to address this concern. In their study, the authors examined various healthcare applications deployed across distributed fog and cloud nodes, each subject to stringent hard constraints (e.g., deadlines) and more flexible soft constraints (e.g., energy resource utilization). To effectively address the time-sensitive nature of healthcare workloads, FL-BETS endeavors to ensure data privacy and combat fraud across multiple . This encompasses local fog nodes and distant cloud infrastructure while minimizing energy consumption and latency.

The authors in \cite{wu2020fedhome} developed a Generative Convolutional Autoencoder (GCAE) to facilitate accurate and individualized health monitoring. By refining the model with a class-balanced dataset generated from user-specific data, the GCAE effectively handles the imbalanced and non-IID characteristic of users' monitoring data. Xu \textit{et al.} in \cite{xu2020fedmax} introduced a distributed ML framework designed for privacy preservation within FL settings and called it FedMax. They tackled several practical obstacles like worker dropout, computational disparities among workers, and communication limitations. Their approach involves the utilization of a relaxed synchronization communication scheme and a similarity-based worker selection method. 

Wang \textit{et al.} \cite{wang2020industrial} proposed a Heterogeneous Brainstorming (HBS) approach as a solution for real-world IoT item detection issues. HBS is characterized by a unique brainstorming methodology along with programmable temperature settings. HSB makes adaptable bidirectional FL of heterogeneous models to be trained on remote datasets. 

The authors in \cite{liu2022large} devised an F-DFRCNN-NE model that enhances IoT privacy and security. This federated neural evolution framework explores neural architecture through evolutionary computation techniques. It encompasses encoding network connections and modules, configuring optimization parameters, defining a search space, employing evolutionary search methods, and generating requisite neural architectures. The federated environment ensures information security throughout the process of federated neural development. DP techniques are deployable to safeguard training data and mitigate misuse risks. Optimization variables encode convolution, pooling, and fully connected modules, enabling flexible neural architecture construction via evolution. The F-DFRCNN-NE model protects participants from cyberattacks while maintains their privacy. 

In \cite {zhao2020local}, Zhao \textit{et al.} focused on integrating FL with Local Differential Privacy (LDP) to facilitate the development of machine learning models in crowdsourcing applications that result in mitigating privacy threats, and reducing communication costs. Specifically, they concentrated on four LDP techniques perturb gradients generated by vehicles. When facing constraints on the privacy budget, the three-Outputs technique offers three alternative output choices to maintain high accuracy. To minimize communication overhead, the output options of Three-Outputs can be encoded using two bits. Furthermore, an optimal piecewise technique (PM-OPT) demonstrates improved performance when the privacy budget is not restricted. Additionally, they introduced a suboptimal technique that is characterized by a simple formula yet comparable to PM-OPT. Subsequently, they devised a novel hybrid mechanism by amalgamating Three-Outputs with PM-SUB. Eventually, they employed an LDP-FedSGD method to jointly coordinate vehicles and a cloud server for model training. 

The authors in \cite{wu2020personalized} offered a customized FL framework deployed within a cloud-edge architecture to serve the needs of intelligent IoT applications. In response to the heterogeneity challenges inherent in IoT environments, they delved into innovative customized FL approaches designed to alleviate the negative impacts of heterogeneity across multiple facets. They highlighted the ability of EC to support the development of advanced IoT applications necessitating swift processing and minimal latency. Alotaibi \textit{et al.} \cite{alotaibi2022ppiov} developed the PPIoV framework that is founded on Blockchain and FL technologies and maintains automobile privacy within the IoV. Given that traditional ML techniques are often inappropriate for distributed and highly dynamic systems like IoV, they employed FL to train a global model while preserving privacy. Moreover, their approach evaluates the reliability of automobiles participating in FL training. Additionally, PPIoV leverages blockchain technology to enhance trust among numerous communication nodes. All transactions take place on the blockchain due to the fact that the local model that are updated by cars and fog nodes communicate with the cloud to update the global model. The authors in \cite{chi2018privacy} introduced an approach for privacy preservation in EC-based DL classification tasks that leverages bipartite topology threat modeling and interactive adversarial deep network construction. Termed Privacy Partition, their strategy addresses deployment scenarios such as IoT smart spaces where users look for both protection and service using DL techniques. In such contexts, a bipartite topology comprising a trusted local partition and an untrusted remote partition emerges as a suitable alternative to centralized and federated collaborative DL frameworks. 

In an EC system based on FL and blockchain technology, the authors in \cite{qin2021privacy} devised a secure method for exchanging MIoT data, safeguarding node privacy by integrating its distinct distributed architecture with the MIoT Edge Computing architecture. Employing blockchain as a decentralized storage mechanism for FL workers ensures security and mitigates tamper-proof concerns. They proposed utilizing reputation and quality as selection criteria for FL participants. They introduced a quality proof technique termed as proof of quality that is implemented on the blockchain to enhance the credibility of edge nodes. Furthermore, their study developed a maritime environment model, extending its applicability to marine contexts through subsequent analysis. Zhou \textit{et al.} \cite{zhou2018real} proposed the Real-Time Robots (RT-robots) architecture, a real-time data processing framework for multi-robot systems that utilize Differential FL. This architecture makes use of substantial data insights through knowledge sharing while ensuring both real-time data processing and data privacy. In this setup, robotic tasks are executed locally in real-time following iterative learning of a globally shared model with DP protection on the cloud. Subsequently, this model is disseminated to multiple edge robots in each round for execution. In \cite{ghimire2022recent}, Ghimire \textit{et al.} focused on the security aspect, where they examined various approaches to tackle performance challenges like accuracy, latency, and resource constraints associated with FL, which subsequently impact the security and overall performance of the IoT. They delved into current prominent research endeavors, identified challenges, explored research trends, and provided forecasts regarding the evolution of this new paradigm in the future. The authors in \cite{su2021secure} suggested an FL-enabled approach for AIoT in facilitating private energy data sharing within collaborative edge-cloud smart grids. Initially, they introduced a communication-efficient and privacy-preserving FL framework tailored for edge clouds within smart grid environments. Subsequently, considering the impact of non-IID data, they formulated two optimization challenges for energy data owners (EDOs) and energy service providers, along with a local data evaluation method within the FL framework. Moreover, to encourage participation from EDOs and ensure high-quality model contributions, they devised a two-layer deep RL-based incentive mechanism that accommodates the lack of multidimensional private user information in real-world scenarios. 

For IIoT-enabled systems, Bugshan \textit{et al.} \cite{bugshan2022toward} designed a robust privacy-preserving framework tailored for FL-based Deep Learning (FDL) services. By combining numerous locally trained models without necessitating dataset distribution among participants, FL effectively addresses privacy concerns inherent in traditional collaborative learning paradigms. However, the reliability of the FDL model encounters challenges due to vulnerabilities to intermediate findings and potential data structure leaks during the model aggregation phase. Their framework introduced a Residual Network-based FDL model integrated with a DP service model to construct reliable locally trained models. Additionally, they proposed an edge and cloud-powered service-oriented architecture. To ensure dependable execution while upholding privacy, the service model breaks the functionality of the entire FDL process into distinct services. Finally, they devised a local model aggregation approach for FDL prioritizing privacy protection. 

To accomplish model aggregation for FL while protecting privacy, the authors in \cite{kanagavelu2020two} investigated the implementation of MPC for model aggregation among peers. This technique is often restrictive due to its significant communication costs and scalability limitations. In response, the authors proposed a two-phase solution. Initially, they recommended the formation of a small committee, followed by the provision of MPC-enabled aggregated model services to a wider array of committee participants. This approach integrates into a smart manufacturing IoT platform, facilitating the adoption of an MPC-driven FL framework. This platform empowers a group to collaboratively train high-quality models utilizing their datasets on their respective premises without compromising execution efficiency in terms of communication costs, execution time, and model accuracy relative to traditional ML methods, while preserving privacy. 

Sharing security records among the involved devices leads to privacy concerns. Therefore, by outlining a novel detection System (IDS) for EoT, the authors in \cite{lalouani2022robust} introduced a solution to address these challenges. Their proposed IDS uses FL to enable edge nodes to share models instead of raw data and to aggregate these models hierarchically. Moreover, their approach includes mechanisms to detect any attempts, whether individual or collaborative, to undermine the IDS by disseminating false (poisonous) data. They employed a Louvain technique to identify fraudulent groups and an iterative voting procedure to assign trust levels to participating devices.

The rapid proliferation of AI-enabled security applications has underscored the necessity for gathering diverse and scalable data sources to effectively evaluate their performance. The authors in \cite{moustafa2020data} collected a new dataset that is denoted as ToN IoT datasets. This dataset consists of distributed data from Telemetry datasets of IoT services, Operating systems datasets and datasets of Network traffic. Moreover, they delineated a testbed architecture crafted for collecting Linux datasets derived from audit traces of hard disks, memory, and processes. This architecture is structured into three distributed layers: edge, fog, and cloud. The edge layer encompasses IoT and network systems. The fog layer integrates virtual machines and gateways. The cloud layer integrates data analytics and visualization tools that are interconnected with each other. The combination of these layers is programmatically managed using SDN and Network-Function Virtualization (NFV) technologies that is facilitated by platforms such as VMware NSX and vCloud NFV. The Linux ToN IoT datasets serve as a foundational resource for training and validating various innovative federated and distributed AI-enabled security solutions, including intrusion detection, threat intelligence, privacy preservation, and digital forensics.

\begin{table*}[htbp]
\footnotesize
\centering
\caption{References on FL-based Privacy-Preserving Structures}
\label{FLPPS}
\setlength{\aboverulesep}{0.05cm}
\setlength{\belowrulesep}{0.05cm}
\renewcommand{\arraystretch}{1.2}
\begin{tabularx}{\textwidth}{cccc>{\raggedright\arraybackslash}m{11cm}}
\specialrule{0.12em}{0pt}{0pt}
Reference& Year&Acc$>90\%$&AI/ML approach & \multicolumn{1}{c}{Open Areas and Future Challenges and Directions}\\\specialrule{0.12em}{0pt}{0pt}
\cite{zhang2021adaptive}& 2022 &\cmark &AdaPFL& Enhancing the FL's convergence pace, and creating and deploying an FL key distribution method in real-ship settings. \\ \specialrule{0.0005em}{0pt}{0pt}
\cite{xu2022c}& 2022&\cmark&fDRL& Examine the accuracy of the created FL scheme across DQN for numerous data points at different edges (local models) and add the associated processing cost of C-fDRL. \\ \specialrule{0.0005em}{0pt}{0pt}
\cite{tayyab2023comprehensive}&2023 &\textbf{---} &DL, RL, CNN& Outlining current research gaps and suggesting future avenues for exploration in the realm of security and privacy, including concerns with encrypted data, Whitebox attack concerns, and more investigation on adversarial vulnerability. \\ \specialrule{0.0005em}{0pt}{0pt}
\cite{neto2022collaborative}&2022 &\xmark &DL& Examining model aggregation, hyperparameter customization, various attacks, computation time, and connectivity issues. \\ \specialrule{0.0005em}{0pt}{0pt}
\cite{ferrag2022edge}&2022&\cmark& \textbf{---}& \textbf{---}\\ \specialrule{0.0005em}{0pt}{0pt}
\cite{li2021efficient}&2022& \textbf{---}&EPPFL& \textbf{---}\\ \specialrule{0.0005em}{0pt}{0pt}
\cite{song2020fda}&2021&\xmark&DDN/FDA& \textbf{---}\\ \specialrule{0.0005em}{0pt}{0pt}
\cite{abdel2021federated} &2022 &\cmark&DL-Fed-TH& Investigating the load balancing in multi-cloud IIoT setups, installing a single threat intelligence microservice, and creating a trade-off between productivity and anonymity. \\ \specialrule{0.0005em}{0pt}{0pt} 
\cite{lakhan2022federated}&2021 & &FL-BETS&  Focusing on mobility fraud awareness and intrusion detection for civil maritime usages and determining the cost functions for the system's scalability and security limitations. \\ \specialrule{0.0005em}{0pt}{0pt}
\cite{wu2020fedhome}&2022&\cmark&GCAE& \textbf{---}\\ \specialrule{0.0005em}{0pt}{0pt}
\cite{xu2020fedmax}&2020&\cmark&FedMax& \textbf{---}\\ \specialrule{0.0005em}{0pt}{0pt}
\cite{wang2020industrial}&2021 &\cmark&DL& Applying the asynchronous training for the fog/edge-based networks. \\ \specialrule{0.0005em}{0pt}{0pt}
\cite{liu2022large}&2022 & \textbf{---}&F-DFRCNN-NE& To automatically optimize network to decrease cost, a multiobjective neural search method may be used and testing standards for FL can be further standardized due to its fast growth.\\ \specialrule{0.0005em}{0pt}{0pt}
\cite{zhao2020local}&2021 &\cmark&LDP/SVM& Creating LDP protocols for cutting-edge FL systems. \\ \specialrule{0.0005em}{0pt}{0pt}
\cite{chi2018privacy}&2018 & \textbf{---}&DL& Investigating the viability of deploying the model to large scale ML setups with integration of IoT hardware and software, and studying security issues and best invertibility situations.\\ \specialrule{0.0005em}{0pt}{0pt}
\cite{qin2021privacy}&2022 &\xmark& \textbf{---}& Creating methods to optimize the number of workers to save resources, as well as how to dynamically adjust reputation thresholds to reduce the adverse effects of malicious workers. \\ \specialrule{0.0005em}{0pt}{0pt}
\cite{zhou2018real}&2018 &\xmark&DP&In addition to the real-world robotic identification work, they incorporate their architecture into future real-time IoT applications. \\ \specialrule{0.0005em}{0pt}{0pt}
\cite{ghimire2022recent}&2022 & \textbf{---}& \textbf{---}&  IID, nonmodified, and equal data distribution will be used to address large-scale FL. The prevention of attacks on real FL settings without compromising accuracy will be studied. \\ \specialrule{0.0005em}{0pt}{0pt}
 \cite{su2021secure} &2022 &\cmark&DRL& In AIoT, the blockchain-based robust FL and the local model assessment framework will be examined according to DP-based gradient disturbance. \\ \specialrule{0.0005em}{0pt}{0pt}
\cite{bugshan2022toward}&2022 & \textbf{---}&FDL& Using real datasets for various DL models with smaller parameter sizes. Providing better protection for sharing parameters using various privacy methods like GAN-based policies. \\\specialrule{0.0005em}{0pt}{0pt}
\cite{kanagavelu2020two}&2020 &\cmark&MPC-FL& TL and vertical FL will be examined. Improving efficiency and scalability by assuming a high number of parties and datasets on AWS-CrossRegion and AWS-SameRegion. \\\specialrule{0.0005em}{0pt}{0pt}
\cite{lalouani2022robust}&2022&\cmark&FLACI& \textbf{---}\\ \specialrule{0.0005em}{0pt}{0pt}
 \cite{liu2022distributed}&2022 & \textbf{---}& \textbf{---}& Creating a safe aggregation protocol in a hostile environment, as well as a novel aggregation approach to identify corrupted, robust aggregation recipes and local models. \\\specialrule{0.0005em}{0pt}{0pt}
\cite{fagbohungbe2021efficient}&2022 & \textbf{---}&DL/SplitNN& For image classifications, a comparison between FL and SplitNN in terms of classification accuracy vs. model complexity and transmission time will be performed. \\\specialrule{0.0005em}{0pt}{0pt}
\cite{huong2021lockedge}&2021 & \textbf{---}& DL& Increasing the detection rate of theft-data-Typed attacks by collecting more data sets and learning more about them. \\ \specialrule{0.0005em}{0pt}{0pt}
\cite{ren2022privacy}&2022 &\cmark& \textbf{---}& Taking into account the fact that certain malevolent individuals may also conduct poison attacks throughout the learning procedure,. \\\specialrule{0.0005em}{0pt}{0pt}
\cite{yu2021privacy}&2022 & \textbf{---}&FLCH& Studying blockchain-enabled FL techniques for material storage to improve privacy, as well as studying a theoretical structure to examine the suggested method's convergence. \\\specialrule{0.0005em}{0pt}{0pt}
\cite{jiang2021privacy}&2021 &\cmark&DP& The topic of how to fully utilize DP in FL will be studied, and varied FL frameworks will also be investigated as it plays a role in industrial EC. \\\specialrule{0.0005em}{0pt}{0pt}
\cite{zhang2022privacy}&2022 &\cmark&TP-AMI/ATT-BLSTM& A hardware-based service platform is scheduled to be constructed, and the system will be able to better assess their proposed framework, particularly when transitory faults occur. \\\specialrule{0.0005em}{0pt}{0pt}
\cite{uddin2023sdn} &2022 & \textbf{---}& \textbf{---}& To evaluate system efficacy, actual data is acquired by incorporating device variations, including air, land, and underwater. NS will divide terrestrial IoT networks to discrete slices. \\\specialrule{0.0005em}{0pt}{0pt}
\cite{garg2021security} &2021 & \textbf{---}& \textbf{---}& Detecting the attack in the MEC cluster with the two LSTM stages without improving the DL model by performing the computationally expensive procedure for learning the model. \\\specialrule{0.12em}{0pt}{0pt}
\end{tabularx}
\end{table*}
In the context of fog computing (FC), Liu \textit{et al.} \cite{liu2022distributed} adopted a SA mechanism relying on efficient additive secret sharing that is essential for FL training. Considering the extensive adoption of SA in FL, their protocol prioritizes minimizing communication and processing overhead. Initially, they deployed local services aids the cloud server in aggregating total data during training, utilizing a fog node (FN) as an intermediary processing unit. Subsequently, they formulated a simple Request-then-Broadcast mechanism to enhance protocol resilience against client dropouts. Furthermore, their protocol suggested two straightforward yet effective techniques for client selection. 
In \cite{fagbohungbe2021efficient}, the authors developed an edge intelligent computing architecture for image categorization in the IoT. In this architecture, every edge device independently train an autoencoder before transmitting the acquired latent vectors to the edge server for training. This strategy guarantees the safeguarding of end-users' data and diminishes transmission overhead. Unlike FL, their proposed system allows for individual autoencoder training at each edge device without dependence on a server during classifier training. This methodology alleviates the substantial communication costs typically associated with collaborative intelligence algorithms like Split Neural Networks (SplitNN). Additionally, the transmission of latent vectors maintains end-users' data privacy without incurring supplementary encryption expenses. 
In \cite{huong2021lockedge}, the authors proposed an edge-cloud architecture to perform detection tasks directly at the edge layer and close to the origin of attacks. This strategy facilitates rapid response, adaptability, and diminishes the workload on the cloud. Furthermore, they introduced LocKedge, a multi-attack detection system recognized for its high accuracy and ease of implementation at the edge.
While most previous techniques only considered the privacy of the local model, the authors in \cite{ren2022privacy} introduced additive homomorphic encryption and double-masking to secure the user's local model and the aggregated global model simultaneously. Furthermore, linear homomorphic hashes and digital signatures were employed for traceable verification that allows users to validate the accuracy of aggregate results and detect incorrect epochs. Additionally, their proposed protocol ensures the privacy of both local and global models and guarantees verification traceability, even in instances of collusion between the cloud server and fraudulent users.
Yu \textit{et al.} \cite{yu2021privacy} discussed FL-based cooperative hierarchical caching (FLCH), a mechanism leveraging IoT devices to construct a collective learning model for forecasting content popularity while maintaining data integrity locally. FLCH enhances vertical collaboration between the baseband unit (BBU) pool and F-APs, alongside horizontal cooperation among proximate F-APs, for caching items with varying degrees of popularity. Moreover, to ensure robust privacy protection, FLCH integrates a DP approach. 

For industrial data processing, the authors in \cite{jiang2021privacy} proposed a federated edge learning architecture integrating hybrid DP and adaptive compression strategies. In order to protect the transmission risks associated with gradient parameters in an industrial environment, the architecture initiates adaptive gradient compression preparation, constructs the industrial FL model, and subsequently utilizes the adaptive DP model for optimization. This approach significantly improves the safeguarding of terminal data privacy against inference attacks by maximizing the collaborative potential between hybrid DP and adaptive compression techniques.

In \cite{zhang2022privacy}, a TP AMI service paradigm based on differentially private FL was modeled to balance the QoSs and protect users' privacy. Their service model involves training neural network models locally and exclusively sharing model parameters with the central server which results in avoiding the transmission of private energy data to the cloud server. Furthermore, individual identities are hidden by incorporating random Gaussian noise during SA. The challenge of long-range dependency in traditional neural networks is addressed using an attention-based bidirectional LSTM (ATT-BLSTM) neural network model. In a case study, a residential short-term load forecasting task is utilized to evaluate the performance of the proposed model.
Uddin \textit{et al.} in \cite{uddin2023sdn} developed a strategy to prevent data breaches that could compromise the integrity of the satellite-IoT architecture deployed for space communication. Leveraging SDN, the framework uses FL techniques within distributed systems to securely transmit sensitive data across IoT devices. Additionally, DP mechanisms are incorporated during data exchange among devices to bolster privacy protection. To device a secure MEC environment, the authors in \cite{garg2021security} represented a SecEdge-Learn architecture utilizing DL and transfer learning techniques to tackle diverse attack scenarios under blockchain control. Additionally, their model securely stores the acquired knowledge from the MEC clusters. 

By utilizing FL approaches and creating reputation mechanisms, Zhang \textit{et al.} \cite{zhang2020towards} developed a reliable and privacy-preserving QoS prediction strategy to create a privacy-preserving system. The future directions, open areas, and accuracy of references in FL-based privacy-preserving systems are listed in Table \ref{FLPPS} with details.
\section{Resource Allocation in FL} \label{sec: resalofl}
IoT devices generate substantial amounts of data necessitating transmission, storage, and analysis. To shed more light on the issue, IoT devices are equipped with tiny sensors to collect data from an area of interest (AoI), process it, and transmit it to the servers. Thanks to wireless technologies and the availability of sensors for end users, IoT structures are fully supported. However, one of the main existing concerns is completing the assigned tasks properly during a specific time to address the delay issues on latency-sensitive platforms. 
Yin \textit{et al.} \cite{yin2021privacy} introduced a hybrid privacy-protection method for FL to address both privacy and efficiency concerns. This method primarily employs an enhanced function encryption approach that safeguards the features of the data uploaded by each IoT device and the weights of each client during global model aggregation. Additionally, they designed a local Bayesian DP mechanism in which the noise policy significantly improves adaptability across various datasets. Additionally, the Sparse Differential Gradient technique is employed to enhance the storage and data transmission efficiency of the FL process.

 In \cite{feng2019joint}, the relay network is considered in the construction of a platform for cooperative communication that facilitates the sharing and trading of model updates. Mobile devices, based on their training data, update the system's models, and the cooperative relay network conveys these modifications to the model owner. The model owner values the instructional service provided by the mobile devices, that charge specific fees in return. Given the coupled wireless transmission interference among mobile devices utilizing the same relay node, rational mobile devices must choose their relay nodes and transmission strengths carefully. To address this, a Stackelberg game model was formulated to examine the interactions between the mobile devices and the model owner, with the outer point approach used to analyze the Stackelberg equilibrium. 
 
To deliver improved model accuracy with strict privacy assurances and excellent communication efficiency, the authors in \cite{han2022pcfed} suggested PCFed, a revolutionary privacy-enhanced FL framework. To adaptively reduce communication frequency, they introduced a sampling-based intermittent communication technique utilizing a PID (proportional, integral, and derivative) controller on the cloud server. Additionally, they developed a budget allocation system to balance model accuracy with privacy concerns. Addressing the continuous data streams on edge servers, they designed PCFed+, an enhanced version of PCFed. Extensive experiments demonstrated that both PCFed and PCFed+ outperforms previous systems in terms of communication efficiency, privacy protection, and model accuracy.
Luo \textit{et al.} in \cite{luo2022communication} developed a semi-asynchronous federated learning (SAFL) framework aims at addressing data security concerns and enhancing model performance by enabling multiple devices to train a shared model. This framework seeks to reduce computational and transmission latency while increasing the learning rate. To ensure communication efficiency, they formulated a combined problem of terminal device selection and resource allocation within the SAFL framework. They employed a deep deterministic policy gradient (DDPG) method to determine the optimal solution to their proposed MDP problem.
The authors of \cite{majumder2022review} examined IoT-based EC applications and the associated ML tools. One persistent challenge in FL is reliably completing tasks within the allocated time frame to meet the latency constraints of latency-sensitive applications. Given the limited processing resources of IoT devices, assistance from backend computing equipment is often necessary to perform complex operations. These backend devices, located near the network's edge, are typically not offloaded to the cloud to avoid significant network delays. Edge computing emerges as the optimal solution for IoT applications requiring minimal latency.

The authors in \cite{qiao2022adaptive} revealed that pooling model parameters from all participating devices may not optimize FL-based content caching performance when training data is non-IID. Instead, adjusting the local iteration frequency adaptively when resources are limited was found to be more crucial. FL faces challenges in aggregating updates from all participating devices simultaneously and maintaining a fixed iteration frequency in local training due to the non-IID nature of data and restricted edge resources. To mitigate these issues, a distributed resource-efficient FL-based proactive content caching (FPC) policy was proposed to improve resource efficiency and content caching performance. The FPC problem is formulated as a stacked autoencoder model loss minimization issue that takes resource limitations into account through theoretical analysis. An adaptable FPC (AFPC) algorithm was then presented, incorporating DRL with client selection and the determination of local iteration numbers.
Tam \textit{et al.} \cite{tam2021adaptive}, suggested an adaptive model communication system for edge FL that incorporates virtual resource optimization. Utilizing a deep QL algorithm, a self-learning agent interacts with a network function virtualization orchestrator and an SDN-based architecture. Within a virtualized infrastructure manager, the agent optimizes the resource control policies of virtual multi-access EC entities. Their proposed method includes a learning model trained to identify the optimal actions for specific network states. The approach takes into account various spatial-resolution sensing conditions during the exploitation phase and allocates compute offloading resources for aggregating global multi-CNN models based on congestion states.
The authors in \cite{al2021energy} presented an Energy-aware Multi-Criteria Federated Learning (EaMC-FL) model for EC. This model combines locally trained models on selected
representative edge nodes (workers) to enable collaborative training of a
shared global model. Initially, the edge nodes are grouped into clusters based on the similarity of their local model parameters. Subsequently, a small subset of representative workers is chosen using a multi-criteria evaluation approach for each training round. The study evaluates the representativeness or significance of these workers, considering factors such as the node's local model performance, energy consumption, and battery life.
Zhao \textit{et al.} \cite{zhao2021communication} developed a semi-hierarchical federated analytics framework, incorporating the benefits of various earlier described architectures. This framework requires no central server or cloud architecture since it uses many ESs to combine learned model weights and aggregate updates from IoT devices. Additionally, they introduced a new local client update rule to improve communication effectiveness by lowering the number of communication rounds between ESs and IoT devices. The provided approach's characteristics are investigated, and convergence properties are analyzed after considering variables like changing parameters, erratic links, and packet loss.
The authors in \cite{park2022completion} explored FL within a fog radio access network scenario, where multiple IoT devices collaborate to jointly train a shared ML model. Enabled by dispersed access points (APs), these devices establish communication with a cloud server. To address the constrained capacity of the fronthaul links connecting the APs to the cloud server, they proposed the adoption of a rate-splitting transmission technique for IoT devices. This approach enables the decoding of split uplink signals utilizing both edge and cloud resources. Their primary objective is to diminish the FL completion time that is achieved through enhancements in fronthaul quantization techniques, rate-splitting transmission, and adjustments to training hyperparameters such as precision and iteration rates.
Xu \textit{et al.} \cite{xu2021cybertwin} developed a method to achieve efficient EC within a heterogeneous wireless environment, termed as Cyber-Twin assisted asynchronous federated learning (AFL). The primary aim is to optimize the utilization of local computing resources, with the Cyber-Twin acting as a communication facilitator. Initially, within the AFL training process, they introduced the Cyber-Twin as a communication coordinator to facilitate individual model aggregation between users and the cloud server. Subsequently, the Cyber-Twin serves as an intelligent agent for EC resource optimization, taking into account both local computing capabilities and uplink transmission. They formulated a resource optimization problem considering various factors such as computer capacities, data sizes, and available connection bandwidth by utilizing the block coordinate descent method to derive optimal resource management strategies.

A non-orthogonal multiple access (NOMA)-based joint resource allocation and IIoT device orchestration strategy for MEC-assisted HFL was presented in \cite{zhao2022drl}. Utilizing DRL, this approach mitigates overhead, optimizes resource allocation, and enhances model accuracy. To address challenges such as latency, energy consumption, and model accuracy within the constraints of IIoT devices' computational and transmission capabilities, they formulated a multi-objective optimization problem. In response, they devised a DRL technique grounded in the DDPG to tackle these challenges. 

Communication quality among clients and servers is crucial during training rounds to manage available resources. For example, the authors in \cite{zhai2021dynamic} demonstrated that the time-varying reliability of links in the wireless network of the smart grid posed challenges to communication between clients and the server during FL training rounds. This variability in link dependability results in a deceleration of the model convergence rate and inefficient utilization of resources during local training processes. Recognizing the highly dynamic nature of link reliability, this study investigated a dynamic FL problem within the domain of power grid mobile edge computing (GMEC). They devised a delay deadline-constrained FL system to mitigate prolonged training delays and developed a dynamic client selection mechanism to enhance computational efficiency within this learning framework.
The authors in \cite{zhu2021dynamic} presented a prediction-assisted task offloading system for the IoT power grid that optimizes offloading decisions while preserving privacy. Through an accurate selection of local servers, they devised an FL strategy to expedite the training of a task prediction model. Subsequently, they augmented a dynamic prediction-assisted task offloading system based on the traffic loads of computation workloads at EC nodes and the behavioral characteristics of electrical users.
The authors in \cite{al2022energy} looked at a resource allocation strategy to lower FL overall energy usage in relay-assisted IoT networks. Their goal was to reduce IoT device energy usage while meeting the FL time restriction, which consists of wireless transmission latency and model training calculation time. To achieve this, they modeled a joint optimization problem that considered IoT device scheduling with relays, transmit power distribution, and computation frequency distribution. However, due to the problem's NP-hardness, developing a globally optimal solution is impossible. They consequently suggested using graph theory to jointly find near-optimal and low-complexity suboptimal solutions. 
The authors in \cite{wu2021fast} proposed a FedAdp weighting approach to expedite model convergence in the presence of nodes with non-IID datasets. This simple yet effective method substantially decreases the number of communication rounds by dynamically motivating positive node contributions while suppressing negative ones. 

Yan \textit{et al.} \cite{yan2020federated} focused on distributed power allocation for edge users in decentralized wireless networks to enhance energy and spectrum efficiency while maintaining privacy within an FL framework. Considering the dynamic and intricate nature of wireless networks, they selected an online Actor-Critic (AC) architecture as the local training model. FL facilitates collaboration among edge users through the distribution of gradients and weights generated in the Actor-network. To address overfitting resulting from data leakages in a non-IID data environment, they introduced a federated augmentation mechanism employing the Wasserstein Generative Adversarial Networks (WGAN) method for data augmentation. With this federated augmentation, each device refreshes its data buffer using a generative WGAN model until it achieves an i.i.d. training dataset. Through comparison with direct data sample exchange approaches, they achieved a substantial reduction in communication overhead in distributed learning.
The authors in \cite{hou2023joint} addressed the challenge of optimizing resource allocation and compute offloading concurrently in Cellular Vehicle-to-Everything (C-V2X) networks. Their proposed solution involves a hierarchical MEC/C-V2X network that takes into account diverse computation offloading patterns and the dynamic nature of vehicular networks. Additionally, they devised an offloading model for cooperative processing that accommodates various offloading patterns. Leveraging MDP principles, they formulated the resource allocation and dynamic computation offloading problem as a sequential decision-making task. They introduced ORAD, a DL system based on the DDPG algorithm, to optimize offloading success rates in real-time that facilitates intelligent and automated decision-making.
To improve the accuracy of federated models in non-IID circumstances, the authors in \cite{yan2021federated} introduced the FedCC algorithm to address a fundamental issue encountered in FL, which pertains to the excessive communication required for parameter synchronization. This communication overhead not only wastes bandwidth and prolongs training time but also has the potential to impact model accuracy. To mitigate this challenge, the FedCC technique partitions clients into groups based on the similarity of their data and selects a representative model from each group for the task of uploading to the cloud server for model aggregation. However, it is important to note that the utilization of the FedCC algorithm leads to a modest reduction in the test accuracy of the federated model and an increase in the communication cost during the training phase of the federated model.
To enhance the decentralized SGD (DSGD) algorithms, the authors in \cite{xing2021federated} devised versatile digital and analog device-to-device (D2D) wireless implementations of communication-efficient DSGD algorithms, leveraging over-the-air computation for simultaneous analog transmissions and random linear coding for compression. To effectively allocate resources, they established convergence bounds for both digital and analog implementations under convexity and connectivity assumptions. The authors in \cite{foukalas2022federated} outlined a set of application requirements for implementing FL within an IoT device architecture, distinguishing between confined and unconstrained scenarios according to Internet Engineering Task Force standards. The FL protocols consist of three phases: initial configuration, distributed training, and cloud updates. Utilizing an IoT platform, they conducted experiments to evaluate the performance of the FL protocols in terms of accuracy, timing, cost-effectiveness, and latency.
The authors in \cite{jiang2022fedsyl} presented the Federated Synergy Learning (FedSyL) paradigm that addresses the trade-off between the risk of data leakage and training effectiveness. They investigated the intricate relationship between local training latency and multi-dimensional training configurations, offering a method for predicting training latency uniformly using polynomial-quadratic regression analysis. Additionally, they offered an ideal model offloading approach, considering the resource constraints and computational heterogeneity of end devices. This method correctly assigns device-side sub-models to end devices whose capabilities vary. 
Saha \textit{et al.} \cite{saha2020fogfl} developed the FogFL framework, which facilitates FL in resource-limited IoT environments, particularly for latency-sensitive applications. Despite FL's popularity, it suffers from drawbacks such as high computational demands and communication overheads. Vulnerability to malicious attacks stems from reliance on a single server for global aggregation in FL, leading to inefficiencies in model training. To address these issues, geospatially positioned fog nodes are integrated into the FL architecture as local aggregators. These nodes serve specific demographics that facilitates the exchange of location-based data among applications with related contexts. Additionally, they proposed a greedy heuristic method for selecting the optimal fog node to serve as a global aggregator in each interaction between the edge and the cloud. By reducing reliance on centralized server operations, FogFL leverages fog nodes to minimize communication latency and energy consumption of resource-constrained edge devices while maintaining the rate of global model convergence, thus enhancing system reliability.
The authors in \cite{balasubramanian2021intelligent} demonstrated two fundamental dimensions of an IoT network: security and streamlined data connectivity. Leveraging the FL architecture, which efficiently allocates data and computational resources on end-user devices, they capitalized on FL principles, particularly caching, within the domain of EC for IoT applications. The authors in \cite{guo2021inter} investigated FL for ultra-dense edge computing (UDEC). They offered a resource-effective inter-server FL method that makes the servers and clients interact with each other collaboratively.\\ 
To investigate the difficulties of independent DRL training, deployment, and inference in the microgrid cluster scenarios, \cite{duan2022lightweight} suggested a distributed microgrid cluster-specific federated DRL-based request scheduling method where maximizing the system's long-term utility is the goal. The DRL model improves its applicability for edge nodes with limited resources. The authors in \cite{lee2020market} introduced a market model for managing distributed learning resources across various MEC operators to meet budget and latency requirements for IoT data analytics. Within the FL framework, a cloud-based coordinator efficiently distributes sensing data from IoT devices to numerous MECs, focusing on a hybrid cloud-MEC architecture for distributed learning. Each MEC receives a shared model from the coordinator and conducts local training using the acquired partial sensing data. Subsequently, the cloud coordinator integrates local training results from MECs to form a global model. A Stackelberg game model was developed and solved to represent the hierarchical decision-making structure resembling market behavior, with MEC operators acting as leaders. A pricing strategy was developed to optimize their utility by considering the trade-off between revenue and energy usage. The coordinator balances the cost and satisfaction in distributed learning and coordinates IoT sensors to transfer sensing data via MECs in alignment with decisions made by MEC operators. A distinct Stackelberg equilibrium point is obtained to enhance the utility of each market member.
In \cite{hou2021multiagent}, Hou \textit{et al.} integrated EC and AI within a Cybertwin-based network, introducing a hierarchical task offloading approach suitable for both delay-tolerant and delay-sensitive missions. With a focus on addressing the significant challenges posed by diverse IoT application requirements, heterogeneous multidimensional resources, and time-varying network conditions, their objective was to ensure user QoE, minimize latency, and provide reliable services. Their approach utilizes a multi-agent Deep Deterministic Policy Gradient (MADDPG) algorithm to expedite task processing, facilitate dynamic real-time power allocation, and reduce overhead. Moreover, they adopted FL to train the MADDPG model, thereby enhancing system processing efficiency and task completion rates. 

The authors proposed a Dynamic Cooperative Cluster Algorithm (DCCA) in \cite{xin2021optimization} to reduce delays in a problem that has been shown to be NP-hard. D2D and opportunistic communication are also used for node-to-node communication. Clusters are set up using the DCCA approach based on the ability of several nodes to train models together and cut down on delays. An original dynamic cooperative cluster algorithm that was based on similarity is established first. Then, the dynamic cooperative cluster is adjusted using a different algorithm focusing on the core nodes' computing and transmission capacities. 
The authors of \cite{zhang2021optimizing} optimized the collaborative decision-making process concerning computing, device selection, and spectrum resource allocation, with the goal of maximizing the efficiency of distributed IIoT networks through FL. Introducing a three-layer collaborative FL architecture, they devised a framework to enable efficient FL across geographically dispersed data and facilitate DNN training. Locally, devices leverage their distributed data to train DNN models via industrial gateways, while ESs aggregate local models to generate the global model at each FL epoch or periodically at a cloud server. To address the challenge of optimizing device selection and resource allocation while minimizing FL evaluation loss, they formulated a stochastic optimization problem. Given the inherent complexity of FL's objective function, which assesses loss alongside energy usage, conventional optimization techniques are inadequate. To tackle this, they proposed a decentralized approach based on deep Multi-Agent Reinforcement Learning (MARL) called Reinforcement on Federated (RoF). Implemented at ESs, the RoF scheme facilitates group decision-making for device selection and resource allocation, with an additional device refinement subroutine to expedite convergence while conserving on-device energy. 
Using Earth Mover's Distance to determine the weights of various node characteristics, the authors of \cite{lu2021parameters} proposed an approach to address the balance issue in FL that mitigates the bias towards specific distributed nodes within the dataset and alleviate the impact of non-IID data problems on the model. Additionally, the study investigated a method to reduce redundant communication between nodes and servers during training, thus optimizing the efficiency of the FL process.
To address the issue of standby energy reduction in residential structures, the authors of \cite{gao2022residential} proposed a residential energy management system with a personalized FDRL framework. The solution was designed to protect privacy, enhance communication effectiveness, and eliminate the reliance on cloud services. 
By eliminating the need for a centralized cloud server, Khan \textit{et al.} \cite{khan2020self} suggested a revolutionary FL framework that makes FL possible. They first devised a social awareness-based clustering method, and then they selected the cluster heads (CHs). The global FL time is reduced by formulating an optimization problem. They introduced a heuristic technique to optimize the total FL time after recognizing the NP-hardness of the optimization problem. To improve FL communication efficiency, the authors of \cite{asad2021thf} presented a unique 3-way hierarchical framework (THF). To reduce client communication costs, their structure includes a CH corresponding with the cloud server through edge aggregation. According to this method, clients submit their local models to their associated CHs, who then send them to the appropriate edge server. Once edge precision is attained, the edge server applies model averaging and iterations. The edge models are then uploaded to the cloud server by each edge server for global aggregation. Through closer distances between source and destination, this 3-way hierarchical network structure optimizes model downloading and uploading. The authors also established a shared communication and computation resource management framework by choosing customers, lowering the FL's overall cost. The authors of \cite{dong2021towards} developed an efficient FL-based NIDS (Network Intrusion Detection System). They used the fact that network traffic data is tabular in nature, therefore, small value changes do not affect the data's fundamental properties. They applied data binning to extract feature data from clients. The classifier on the server is then trained using the extracted feature data. Techniques for data masking are also used to further improve energy efficiency and data privacy.

Yu \textit{et al.} \cite{yu2020deep} enhanced an intelligent UDEC (I-UDEC) framework to develop resource allocation strategies and real-time and low-overhead computation offloading judgments. To this end, a two-timescale deep RL method was proposed that combines a fast-timescale and slow-timescale learning procedure to achieve the goal. By optimizing compute offloading, resource allocation, and service caching placement, their approach reduces total offloading time and network resource utilization while protecting edge device data privacy. 
The authors in \cite{shamseddine2020novel} offered an adaptable and intelligent approach to federation construction using Genetic Algorithm and ML models and a novel architecture for the federated fog concept. The fog federations' main goal is to make it possible for fog providers to provide the necessary QoS. The idea allows for effective load dispersion by sharing resources among various fog suppliers. As a result, the problem of QoS degradation caused by local overloads is successfully resolved. It enables end users to experience real-time applications without delays. The continuous real-time management of electrical equipment presents significant challenges concerning reliability and communication efficiency. Several issues that happen, include the detrimental impact of electromagnetic interference on the reliability of the digital twin (DT) technology, the high communication costs associated with the training of DT models, and the inadequacies in coordinated resource allocation across the cloud, edge, and device layers. The authors in \cite{liao2022cloud} introduced a solution termed as C3-FLOW (Cloud-edge-device Collaborative reliable and Communication-efficient Digital Twin). By optimizing device scheduling, channel allocation, and computing resource allocation in a coordinated manner, C3-FLOW minimizes the long-term global loss function and reduces the time-averaged communication cost.
To achieve ubiquitous intelligence in 6G, the authors of \cite{huang2021collaborative} developed a decentralized and collaborative ML architecture for intelligent edge networks. Their architecture incorporates a compute offloading and resource allocation strategy, facilitated by multi-agent DRL. Recognizing the critical importance of energy efficiency in the establishment of sustainable edge networks, the primary objective is to minimize overall energy consumption while satisfying latency requirements. To address the challenges posed by computing complexity and signaling overhead during the training phase, a federated DRL system was devised.
The authors in \cite{zhou2022communication} proposed a technique known as Distilled One-Shot Federated Learning (DOSFL) to dramatically reduce communication costs while achieving performance parity. In DOSFL, each client compresses its dataset into a single round, transmitting synthetic data to the server for collective model training. Post-model updates, the distilled data becomes irrelevant, resembling noise. With up to $99\%$ of centralized training performance preserved, DOSFL's weightless and gradient-free design yields communication costs up to three orders of magnitude lower than FedAvg. DOSFL's effectiveness is demonstrated across various tasks, utilizing models such as CNNs, LSTMs, and Transformers. Notably, even with stolen distilled data, an eavesdropping attacker cannot successfully train a model without prior knowledge of baseline model weights. Offering communication cost reductions of less than $0.1\%$ compared to conventional methods, DOSFL presents an economical strategy for swift convergence on successful pre-trained models.
The authors in \cite{su2021data} focused on a dynamic strategy for adaptive sensor scheduling and power regulation in federated edge learning, employing a residual feedback mechanism. In this approach, each sensor retains a local residual to preserve gradients that are not transmitted to the central server, rather than discarding them. Additionally, they leveraged the Lyapunov drift optimization technique to investigate the relationship between training progress and resource allocation, linking model update iterations to a dynamic evolutionary process. The decentralized optimal solution derived from this study is customized to incorporate information about channel conditions and data significance, enabling efficient selection of transmission opportunities and important gradients.
Tao \textit{et al.} \cite{tao2022data} presented a data-driven matching methodology for vehicle-to-vehicle (V2V) energy management. In the offline phase, they applied DRL to ascertain the long-term benefits of matching actions within a formulated MDP. Furthermore, they devised an FL architecture that facilitates collaboration among multiple Electric Vehicle (EV) aggregators while safeguarding the privacy of EV owners' data. To enhance computational efficiency, a matching optimization model was developed and converted into a bipartite graph problem during the online matching phase. This approach assists EV owners in reducing expenses and increasing revenues.

\begin{table*}[htbp]
\footnotesize
\centering
\caption{References on FL-based Resource Allocation}
\label{FLRAS}
\resizebox{\linewidth}{!}{%
\setlength{\aboverulesep}{0.05cm}
\setlength{\belowrulesep}{0.05cm}
\renewcommand{\arraystretch}{1.2}
\begin{tabularx}{\textwidth}{cccc>{\raggedright\arraybackslash}m{11.7cm}}
\specialrule{0.12em}{0pt}{0pt}
Reference& Year&Acc$>90\%$&AI/ML approach & \multicolumn{1}{c}{Open Areas and Future Challenges and Directions}\\\specialrule{0.12em}{0pt}{0pt}
\cite{han2022pcfed}   &2022 &\xmark &PCFed/PCFed+& To extend their model, they are considering specific DP for FL devices with varied security resource allocations and examining its effects on communication and privacy issues. \\ \specialrule{0.0005em}{0pt}{0pt}
\cite{majumder2022review}&2022 &\xmark &\textbf{---}& Studying FL methods in deeper depth, focusing on advancements while analyzing IID and Non-IID Data. \\ \specialrule{0.0005em}{0pt}{0pt}
\cite{qiao2022adaptive}  &2022 &\cmark&AFPC& Using budgets effectively for collaborative learning in heterogeneous edge systems, and extending the AFPC to more non-convex optimization challenges with limited resources. \\ \specialrule{0.0005em}{0pt}{0pt}
\cite{tam2021adaptive}   &2021 &\cmark&CNN/DQL& In the edge FL models, NS for diverse spatial image classifications will be evaluated regarding power allocation, service cache placement, and computation offloading decisions. \\ \specialrule{0.0005em}{0pt}{0pt}
\cite{al2021energy}  &2021 &\cmark& EaMC-FL & Further evaluating their proposed EaMC-FL system by investigating its viability on bigger scale actual datasets and various FL situations. \\ \specialrule{0.0005em}{0pt}{0pt}
\cite{zhao2021communication}   &2022 &\xmark&\textbf{---}& Enhancing the actual efficiency of FL structures through tuning communication networks. \\ \specialrule{0.0005em}{0pt}{0pt}
\cite{park2022completion}  &2022 &\xmark&\textbf{---}& The convergence rate determines the number of iterations necessary for convergence, although its analysis for broad sequential convex approximation methods will be performed. \\ \specialrule{0.0005em}{0pt}{0pt}
\cite{xu2021cybertwin} &2021 &\cmark&AFL& More research into the scalability, mobility, and connectivity of Cybertwin-assisted wireless asynchronous federated learning (AFL). \\ \specialrule{0.0005em}{0pt}{0pt}
\cite{zhao2022drl} &2022 &\textbf{---}&HFL& Testing the effectiveness of our proposed solution on actual data and modifying it accordingly while employing DRL in the context of FL. \\ \specialrule{0.0005em}{0pt}{0pt}
\cite{zhai2021dynamic}&2021 &\textbf{---}&\textbf{---}& Upgrading the specific diffusion/uploading latency paradigm connected with special access technology. \\ \specialrule{0.0005em}{0pt}{0pt}
\cite{al2022energy}&2022&\cmark&JEADS-G&\textbf{---}\\ \specialrule{0.0005em}{0pt}{0pt}
\cite{wu2021fast}&2021&\cmark&FedAdp&\textbf{---}\\ \specialrule{0.0005em}{0pt}{0pt}
\cite{hou2023joint}  &2023 &\textbf{---}&DRL/ORAD& Concentrating on decentralized ML-based resource consumption in C-V2X structures. \\ \specialrule{0.0005em}{0pt}{0pt}
\cite{xing2021federated}  &2021 &\textbf{---}&DSGD& Evaluating convergence bounds for non-convex wireless designs based on diminishing consensus rate and decreasing training step size. \\ \specialrule{0.0005em}{0pt}{0pt}
\cite{jiang2022fedsyl}&2022&\cmark&DNN/FedSyL&\textbf{---}\\ \specialrule{0.0005em}{0pt}{0pt}
\cite{balasubramanian2021intelligent} &2021 &\textbf{---}&\textbf{---}& Extending the mobile device cloud Network in terms of scalability, self-organization, and automation, as well as using different datasets and FL methodologies. \\ \specialrule{0.0005em}{0pt}{0pt}
\cite{duan2022lightweight}&2022&\xmark&DRL&\textbf{---}\\ \specialrule{0.0005em}{0pt}{0pt}
\cite{hou2021multiagent}   &2021 &\textbf{---}&MADDPG& Using FL and blockchain-based techniques to train their suggested framework in a decentralized manner for low training overhead and data privacy. \\ \specialrule{0.0005em}{0pt}{0pt}
\cite{xin2021optimization}  &2021 &\textbf{---}&HFL& Furtherly investigating clients' spontaneous cooperation and resource allocation difficulties. \\ \specialrule{0.0005em}{0pt}{0pt}
\cite{zhang2021optimizing}  &2021 &\xmark&DNN& A data importance-aware device selection strategy is investigated in order to optimize FL in large-scale IIoT systems. \\ \specialrule{0.0005em}{0pt}{0pt}
\cite{lu2021parameters} &2021 &\xmark&\textbf{---}& Determining whether or not part edge clients have the authorization to communicate with the parameter edge/cloud server. \\ \specialrule{0.0005em}{0pt}{0pt}
\cite{khan2020self}   &2020 &\textbf{---}&\textbf{---}& Considering the center client's distance from other clients, evaluating various scenarios, budget management, and learning methods for self-governing FL in multiple clusters situations. \\ \specialrule{0.0005em}{0pt}{0pt}
\cite{asad2021thf} &2022 &\cmark&THF& The effect of implementing common approaches for strengthening privacy and an inquiry into the inclusion of adversaries in their suggested framework will be investigated. \\ \specialrule{0.0005em}{0pt}{0pt}
\cite{dong2021towards}  &2021 &\xmark&\textbf{---}& Incorporating a system of initial data feature extractors into their model, i.e., incorporating an FL-based feature extractor for initial packet data into the present data binning method. \\ \specialrule{0.0005em}{0pt}{0pt}
\cite{yu2020deep}  &2021 &\textbf{---}& I-UDEC & Investigating the mechanism in which small cell cloud-enhanced e-Node Bs have private service caching policy to train their model with  private data and the trained global model.\\ \specialrule{0.0005em}{0pt}{0pt}
\cite{liao2022cloud} &2022 &\textbf{---}&C$^3$-FLOW&To improve the efficiency of electrical equipment management, the computing resources and heterogeneities of device-side communication will be investigated.\\ \specialrule{0.0005em}{0pt}{0pt}
\cite{zhou2022communication}   &2022 &\cmark&DOSFL& The accuracy of Distilled One-Shot Federated Learning (DOSFL) decreases from 10 to 100 users across both IID and non-IID. It is a DOSFL constraint that will be studied. \\ \specialrule{0.0005em}{0pt}{0pt}
\cite{akubathini2021evaluation}  &2021&\cmark&FastGRNN & The resource limitations of edge nodes made that on-device retraining would be researched. \\ \specialrule{0.0005em}{0pt}{0pt}
\cite{zhang2021federated}  &2021 &\textbf{---}&FDAC& Concentrating on optimizing UAV distribution to increase AGE-IoT effectiveness. \\ \specialrule{0.0005em}{0pt}{0pt}
\cite{zarandi2021federated}&2021&& DDQN/DRL &\textbf{---}\\ \specialrule{0.0005em}{0pt}{0pt}
\cite{wang2021federated}&2021&\cmark&SVM&\textbf{---}\\ \specialrule{0.0005em}{0pt}{0pt}
\cite{sun2022fedtar}&2022&\cmark&WCPN/FedTAR&\textbf{---}\\ \specialrule{0.0005em}{0pt}{0pt}
\cite{malandrino2022flexible}    &2022 &\xmark& FPL & Extending the performance evaluation of the flexible parallel learning approach to consider the junction layer's capacity to weight the input variously with more complicated DNNs. \\ \specialrule{0.0005em}{0pt}{0pt}
\cite{shah2021joint} &2021 &\textbf{---}&DL/DRL& Caching and service function chains are examples of specific applications of their proposed space-terrestrial integrated network, which may be developed for vehicular communications. \\ \specialrule{0.0005em}{0pt}{0pt}
\cite{zhang2021spectrum}   &2022 &\xmark&DNN&A distributed learning resource-management system in large-scale IIoT will be studied.\\\specialrule{0.12em}{0pt}{0pt}
\end{tabularx}
}
\end{table*}

Akubathini \textit{et al.} \cite{akubathini2021evaluation} examined a unified optimization challenge involving power control and multi-timescale job offloading. Targeting all Electricity IoT devices, the objective was to minimize queuing latency while adhering to a long-term energy consumption constraint. Initially, they decomposed the problem into two sub-problems: power control optimizations at a smaller scale and task offloading optimizations at a larger scale. Subsequently, for multi-timescale optimization, they introduced the Federated Deep Actor-Critic-based Task Offloading Algorithm (FDAC). This technique employed two AC networks.

The authors in \cite{zarandi2021federated} suggested a federated DRL framework to address a multi-objective optimization problem. The objective is to minimize the energy consumption of IoT devices while also reducing the anticipated long-term job completion latency. To achieve this, optimization of offloading decisions, computing resource allocation, and transmitting power allocation are undertaken. The authors employed the DDQN technique, with offloading decisions serving as actions to resolve the MINLP issue as a multi-agent distributed DRL problem. Following the selection of offloading options, transmit power optimization or local computation resource optimization are resolved to determine the immediate cost for each agent. FL is integrated at the end of each episode to accelerate the learning process of IoT devices (agents). FL contributes to increased scalability, promoted agent collaboration, and mitigated privacy concerns.
The authors in \cite{wang2021federated} addressed the challenge of minimizing energy and time consumption for job computation and transmission in MEC-enabled balloon networks. These networks utilizes high-altitude balloons (HABs) as flying wireless base stations, leveraging their processing capabilities to handle computational tasks delegated by connected users. Given that each user's workload generated varying amounts of data over time, the HABs need to dynamically adjust their resource allocation strategies to meet evolving user demands. To minimize energy and time requirements for task computation and transmission, the problem is reformulated as an optimization problem. This involves adapting the algorithms for user association, service sequence, and job allocation. A Support Vector Machine (SVM)-based FL technique was devised to determine user associations and tackle this issue. Leveraging the SVM-based FL approach, HABs collectively constructs an SVM model without revealing user historical associations or computational tasks to other HABs. Subsequent to projecting the optimal user associations, enhancements are made to each user's service sequence and job allocation, aiming at reducing the weighted total of energy and time consumption.
In order to support certain computing workloads, the authors in \cite{sun2022fedtar} devised a Wireless Computing Power Network (WCPN) that arranges the computing and networking resources of diverse nodes. Within this framework, they introduced FedTAR, a task-aware and resource-aware FL paradigm, enabling intelligent services within the WCPN. FedTAR was designed to reduce the overall energy consumption of computing nodes by collaboratively optimizing the operational procedures of each node and employing a cooperative learning methodology. By considering specific job requirements and resource constraints, the solution to the optimization problem allows for adaptable NN depth and collaboration frequency across nodes. Additionally, they proposed an energy-efficient asynchronous aggregation approach for FedTAR to accommodate heterogeneous computing nodes, thereby accelerating the convergence rate of FL within the WCPN.
The authors in \cite{malandrino2022flexible} underscored the recurring necessity of integrating both techniques within fog-based and IoT-based environments, and suggested a framework for Flexible Parallel Learning (FPL) that accomplishes both data and model parallelism. Furthermore, they investigated how diverse methods of allocating and parallelizing learning tasks across the involved nodes yield varying computing, communication, and energy costs.
Shah \textit{et al.} \cite{shah2021joint} introduced the Space-Terrestrial Integrated Network (STIN), a network control and resource allocation challenge that happens in extensive IoT deployments. To address this issue, the authors employed state-of-the-art Hierarchical Deep Actor-Critic (H-DAC) networks. The expansive IoT networks cover urban areas, facilitating potential collaboration among them. This collaboration is harnessed to devise a collective strategy whereby IoT networks share the cost of spectrum per unit.
The integrated network control and resource allocation challenge is formulated as a utility maximization problem, which was addressed using deep actor-critic-based RL. These RL-based algorithms manage the allocation of data rates for each IoT network and IoT device, as well as the determination of the cost per unit spectrum for the federated cloud of IoT networks.
Zhang \textit{et al.} \cite{zhang2021spectrum} investigated resource management challenges in dispersed IIoT networks for FL. To facilitate FL, they constructed a three-layer collaborative architecture wherein DNNs are trained locally on selected IIoT devices. To update the global DNN model, ESs or cloud servers periodically collect DNN model parameters. Effective FL in resource-constrained IIoT networks necessitates the distribution of CPU and spectrum resources for training and broadcasting DNN model parameters. They introduced a joint device selection and resource allocation problem to minimize FL evaluation loss while strictly adhering to FL epoch delay and device energy consumption constraints. Recognizing the inherent interdependency between device selection and resource allocation decisions, they transformed the joint optimization problem into an MDP. Subsequently, they considered a dynamic resource management strategy based on DRL techniques. The future directions, open areas, and accuracy of references in proper resource-allocation FL-based systems are listed in Table \ref{FLRAS} with details.

\section{Applications of FL} \label{sec:appfl}


\begin{table*}[t]
\footnotesize
\caption{References on Aerial and Non-Terrestrial Networks}
\label{FLUAV}
\setlength{\aboverulesep}{0.05cm}
\setlength{\belowrulesep}{0.05cm}
\renewcommand{\arraystretch}{1.2}
\begin{tabularx}{\textwidth}{cccc>{\raggedright\arraybackslash}m{10.8cm}}
\specialrule{0.12em}{0pt}{0pt}
Reference& Year&Acc$>90\%$&AI/ML approach & \multicolumn{1}{c}{Open Areas and Future Challenges and Directions}\\\specialrule{0.12em}{0pt}{0pt}
\cite{lim2021towards} & 2021 & \textbf{---}&Gale-Shapley &Using wireless power harvesting to help the UAVs conduct continuous sensing and model training without returning to their bases. \\\specialrule{0.0005em}{0pt}{0pt}

\cite{han2023lmca} & 2023 & \cmark & DL& Taking into account the scalability of detection models within the progressively complex network landscape, an adaptable open-set identification model holds greater promise for real-world network scenarios. Simultaneously, integrating intrusion detection technology with other advancements can enhance IoT security measures. \\\specialrule{0.0005em}{0pt}{0pt}

\cite{lim2021uav} & 2021 & \textbf{---} & \textbf{---}&Integrating worker mobility into the FL supported by UAVs, where employees can move between subregions. \\\specialrule{0.0005em}{0pt}{0pt}
\cite{islam2022triggerless}&2022&\cmark & FDRL&Expanding their FDRL for many attackers with suitable adaptation for battery-limited UAVs. \\\specialrule{0.0005em}{0pt}{0pt}
\cite{aloqaily2021design}   &2021 & \textbf{---}&\textbf{---} & Using Off-Chain Blockchain Storage to manage large-size data and satisfy the consensus on the divisibility and location of services between the service providers. \\\specialrule{0.0005em}{0pt}{0pt}
\cite{cheng2021intelligent}& 2021 & \textbf{---}& FDRL & Utilizing their FDRL architecture to manage resource allocation and scheduling intelligently in accordance with various service needs. \\\specialrule{0.0005em}{0pt}{0pt}
\cite{song2021non}&2022&\cmark&FDRL&\textbf{---}\\\specialrule{0.0005em}{0pt}{0pt}
\cite{yang2022federated}& 2023 & \textbf{---}& SIL &Extending generative adversarial imitation learning model by tackling obstacle-avoidance flight paths, swarm collaboration efficiency, and connection management using self-imitation learning.\\\specialrule{0.0005em}{0pt}{0pt}
\cite{cheng2022auction}&2022&\textbf{---}&FLSDs&\textbf{---}\\\specialrule{0.0005em}{0pt}{0pt}
\cite{do2021deep}&2022&\textbf{---}&DRL&\textbf{---}\\\specialrule{0.0005em}{0pt}{0pt}
\cite{zhang2022robust}&2023&\cmark&SSFL&Developing their method to better use unlabeled data and upgrading the theory of SSFL to make it more applicable in real-life situations.\\\specialrule{0.0005em}{0pt}{0pt}
\cite{zou2023hierarchical}&2023&\textbf{---}&HFL-LSTM/ MADDQN&\textbf{---}\\\specialrule{0.0005em}{0pt}{0pt}
\cite{abderrahim2023data}&2023&\textbf{---}&---&Enhancing the efficacy of the data center-enabled HAP system using various ML methodologies.\\\specialrule{0.12em}{0pt}{0pt}\\\

\end{tabularx}
\end{table*}

FL has extensive applications in various aspects of our lives. FL is crucial for non-terrestrial and aerial networks, supporting the development of robust communication systems to improve connectivity, resource management, and real-time data processing in remote and hard-to-reach areas. In smart cities and homes, FL enables advanced data-driven services while preserving privacy, facilitating efficient energy management, traffic optimization, and personalized services. In the healthcare system, FL allows collaborative training of medical models across different institutions without sharing sensitive patient data, enhancing diagnostic accuracy and treatment outcomes. These diverse applications underscore FL's potential to revolutionize how we leverage data across various domains, ensuring both efficiency and privacy. Each application is elaborated with details in the following subsections \cite{lim2021towards,lim2021uav,islam2022triggerless,aloqaily2021design,cheng2021intelligent,song2021non}.

\subsection{FL in Aerial and Non-Terrestrial Networks}
FL offers significant potential for enhancing the performance and efficiency of aerial networks, including Unmanned Aerial Vehicles (UAVs), High Altitude Platforms (HAPs), Low Altitude Platforms (LAPs), and Low Earth Orbit (LEO) satellites. By enabling decentralized data processing and model training, FL allows these systems to collaboratively learn from vast amounts of data without compromising privacy and security. This approach is particularly beneficial in scenarios where data is distributed across multiple aerial nodes, such as UAVs performing surveillance, HAPs providing broadband connectivity, and LEO satellites facilitating global communications. By leveraging FL, these aerial networks can optimize resource allocation, improve communication reliability, and enhance overall system performance. For example, integrated with FL, UAVs enhance privacy-preserving data collection, enable real-time analytics, optimize resource management, and augment situational awareness. FL and UAVs form a potent synergy that advance fields like IoT applications, intelligent transportation systems, and environmental monitoring, while ensuring data security and efficiency in dynamic environments. In \cite{lim2021towards}, the authors proposed an FL-based methodology to enable collaborative ML while safeguarding privacy across a federation of distinct DaaS providers for the development of IoV applications, such as traffic prediction and parking occupancy management. They leveraged the self-revealing properties of a multi-dimensional contract to ensure accurate reporting of UAV types by considering various sources of heterogeneity, including sensing, computation, and transmission costs. This approach address information asymmetry and incentive mismatches between UAVs and model owners. Subsequently, the Gale-Shapley method is employed to match the most cost-effective UAV to each subregion. Motivated by the integration of UAV-assisted communications within 5G heterogeneous networks, Lim \textit{et al.} \cite{lim2021uav} introduced UAV-assisted FL. Their approach facilitates the utilization of UAVs for intermediate model aggregation in aerial space and as mobile relays for transmitting updated model parameters from data owners to the model owner in FL scenarios. This extension of FL to data owners operating in environments with uncertain network conditions significantly enhances communication efficiency and enables collaborative learning initiatives. To incentivize participation from UAV service providers, they introduced a multi-dimensional contract incentive design, ensuring that UAVs select incentive packages tailored to their unique characteristic, which includes factors like travel expenses. This incentivizes active involvement from UAVs, fostering a collaborative system conducive to the success of FL applications in diverse and challenging environments. For resource-constrained UAVs, the authors in \cite{islam2022triggerless} proposed an FDRL-based intelligent and decentralized job offloading method to enhance the operational efficiency of MUAV systems. This approach improves the quality of offloading policies while maintaining the privacy of MUAV data. However, the inherent intelligence of such systems renders them vulnerable to backdoor attack that disrupts normal functioning and degrades performance. To evaluate the robustness of the offloading strategy against potential adversaries, they devised a triggerless backdoor attack technique tailored for intelligent task offloading UAVs and evaluated its impact on system performance. This comprehensive assessment helps uncover vulnerabilities and strengthens the resilience of intelligent offloading systems against potential threats. Aloqaily \textit{et al.} \cite{aloqaily2021design} built a 5G network empowered by blockchain-enabled UAVs to dynamically balance the supply of network access with users' evolving demands. This technology promises reliable and secure routing to and from end users, along with decentralized service delivery through the concept of DaaS. To achieve a diverse range of sophisticated authenticated services and ensure data availability, both public and private blockchains are integrated into the UAVs. These blockchains are supported by fog and cloud computing infrastructure, ensuring robustness and scalability. The proposed solution is evaluated by contrasting its message exchange and data transfer success rates with those of conventional UAV-supported cellular networks, emphasizing its potential to revolutionize connectivity and service delivery in next-generation telecommunications. 

The offloaded tasks from ground IoT devices are cooperatively carried out by UAVs acting as an edge server and cloud server connected to a ground base station (GBS) in an energy-constrained mobile edge-cloud framework. A UAV is specifically fueled by the laser beam that a GBS transmits, and also wirelessly charges IoT gadgets. In \cite{cheng2021intelligent}, a task offloading and energy allocation problem is explored that maximizes long-term rewards while considering factors such as executed task size and execution delay within constraints like energy causality, task causality, and cache causality. To tackle this challenge, the authors proposed an FDRL framework. This framework is designed to learn the joint decision-making process for task offloading and energy allocation, with the goal of reducing training costs and preventing privacy leakage during the DRL training process. By leveraging FDRL, the study optimizes resource utilization and enhances overall system performance in dynamic and resource-constrained environments.
The authors in \cite{song2021non} devised a Non-Orthogonal Multiple Access (NOMA) FL architecture tailored for a UAV swarm comprising a leader UAV and multiple follower UAVs. Each follower UAV autonomously updates its local model using its acquired data. Subsequently, all follower UAVs form a NOMA group to simultaneously broadcast their individually trained FL parameters or local FL models to the leader UAV. To expedite FL iterations until reaching a certain accuracy threshold, they proposed a combined optimization strategy involving the uplink NOMA transmission durations, downlink broadcasting durations, and calculation rates of both the leader UAV and all follower UAVs. This holistic optimization approach minimizes execution time while ensuring accurate FL model convergence within the UAV swarm environment. 

To shed more light on FL-UAV applications, the popularization of UAVs has boosted various civil applications. Traffic monitoring, where the effective coordination of the UAV swarm plays a significant role in expanding the monitoring range and enhancing execution efficiency. However, due to the isolated local environments and the heterogeneous execution capabilities, it is challenging to achieve highly consistent actions. To coordinate the UAVs' maneuvers by interactively mimicking the leader UAV's actions, the authors in \cite{yang2022federated} leveraged the Generative Adversarial Imitation Learning model to enhance the accuracy of following the leader UAV's actions during the interagent global model download phase. This approach mitigates biased estimations of imitation parameters, thereby improving the fidelity of emulation. Additionally, they integrated the Self-Imitation Learning model into the intransigent local model training phase. This allows the follower UAVs to rectify subtle imitation errors using their own historically significant experiences that results in enhancing learning robustness. Furthermore, they implemented regular updates to the federated gradient, fostering coordinated swarm policies and more effective distributed parameter interactions. These methodologies collectively optimize the performance and coordination of the UAV swarm in FL scenarios.

Cheng \textit{et al.} \cite{cheng2022auction} examined a multiple FL service trading scenarios within networks facilitated by UAVs. FL service demanders (FLSDs) seek diverse datasets from practical clients, such as smartphones and smart vehicles, along with model aggregation services from UAVs, to fulfill their requirements. To facilitate trade among FLSDs, scattered client groups, and UAVs, they devised a trading market based on auctions. This auction system maximizes income for all participants by optimizing winner selection and payment rules. They introduced a concept that merges seller pairs and joint bids, transforms various vendors into virtual seller pairs and considers two distinct seller categories: data sellers and UAV sellers. They proposed one-sided matching-based and Vickrey-Clarke-Groves (VCG)-based techniques, with the former yielding optimal outcomes albeit computationally intensive, while the latter, particularly suitable for scenarios involving numerous players, provided near-optimal solutions with lower computing costs. These methodologies contribute to the efficient and equitable trading of FL services in UAV-supported networks, enhancing collaborative innovation and resource utilization. \\
To allow sustainable FL with energy-harvesting user devices, the authors in \cite{do2021deep} adopted a DRL-based system for cooperative UAV placement and resource allocation, with the objective of maximizing long-term FL performance while accounting for constraints such as limited network bandwidth, collected energy, and energy budgets for UAVs. To address the challenge of long-term energy restrictions, they employed the Lyapunov optimization method, that transforms the energy constraint into a deterministic problem, thereby simplifying the original complexity. By reframing the optimization problem as a MDP, they applied an DRL-based approach to find solutions. This methodology ensures the sustainable operation of UAV-aided wireless networks by enhancing long-term energy savings while optimizing FL performance.

The authors in \cite{zhang2022robust} initially focused on an SSFL framework designed for privacy-preserving UAV image identification that addresses key challenges in this domain. Specifically, they proposed the utilization of a model parameter mixing technique called Federated Mixing (FedMix) to enhance the fusion of FL and semi-supervised learning methods under two scenarios: labels-at-client and labels-at-server. Moreover, they recognized significant statistical heterogeneity in the local data acquired by UAVs, including variations in the amount, characteristics, and distribution of data captured across different situations and camera modules. To mitigate the effects of statistical heterogeneity, they introduced an aggregation method based on the client's involvement in the training process. This method is referred as the FedFreq aggregation rule and dynamically adjusts the weight of the associated local model based on its frequency during training that results in enhancing the robustness and performance of the SSFL framework in heterogeneous environments.

Zou \textit{et al.} \cite{zou2023hierarchical} addressed a day-ahead energy scheduling problem targeting urban prosumers with access to UAV charging facilities. The primary objective is to optimize the overall energy satisfaction of prosumers while maintaining QoS standards for charged UAVs. Their approach involves two distinct stages: the prediction of day-ahead energy requirements, and the energy scheduling process for each prosumer. To address this challenge, they introduced a hybrid technique utilizing stochastic game-based Multi-Agent Double Deep Q-Learning (MADDQN) with a community agent-independent approach, rooted in HFL on LSTM architecture. To uphold data privacy, they employed the HFL-LSTM technique to locally anticipate each prosumer's energy requirements, circumventing centralized data processing. Subsequently, they subjected the problem to stochastic game analysis to identify the Nash equilibrium (NE) approach. Finally, the optimal energy scheduling strategy for each prosumer was determined using MADDQN and the community agent-independent methodology, ensuring efficient energy management while safeguarding privacy and meeting UAV charging requirements. 
In \cite{10.1109/MIC.2023.3307431}, the authors introduced a multiagent FL and dueling double-deep Q-network (D3QN)-based resource allocation approach, termed as Fed-D3QN, to jointly optimize channel selection and power control to fulfill the low latency and reliability demands of IoV services. Their results indicate that the Fed-D3QN algorithm demonstrates robust stability in highly dynamic air–ground integrated networks. Moreover, it effectively reduces the overall delay in vehicle-to-infrastructure (V2I) links and enhances the transmission success rate in vehicle-to-unmanned aerial vehicle (V2U) links. 
For example, urban sensing involves real-time data collection from city environments that is crucial for urban planning, resource management, and enhancing quality of life. However, data collection and network congestion introduces significant latency. FL mitigates these latency issues by processing data locally and only sending aggregated results to a central server. The authors in \cite{wang2023covert} proposed a wireless sensor air-ground integrated FL (AGIFL) network with covert communication. This system employs a HAP as a parameter server to enhance the mobility and sustainability of urban sensors, and a friendly jammer UAV to ensure information security. Their proposed model minimizes FL latency in the urban sensor network and ensures secure data transmission by optimizing local training accuracy, interference power, and transmit power of the FL sensors. The authors in \cite{abderrahim2023data, elmahallawy2022fedhap, elmahallawy2022asyncfleo} presented asynchronous FL AsyncFLEO and FedHAP, two approaches that integrate HAPs as distributed parameter servers (PSs) into FL for satellite communication, specifically LEO constellations, to facilitate rapid and efficient model training. FedHAP is composed of three key components: 1) a model aggregation algorithm, 2) a hierarchical communication architecture, and 3) a model dissemination algorithm. The future directions, open areas, and accuracy of references in FL-based UAV-assisted systems are listed in Table \ref{FLUAV} with details.

\subsection{FL in Secured Healthcare System}
The rapid expansion of the smart healthcare system has made early detection of dementia more accessible and cost-effective. However, the strain on medical facilities caused by the COVID-19 pandemic has necessitated remote diagnosis and treatment by doctors. This shift has led to a surge in the use of IoT-enabled medical equipment, as people become more mindful of their health. Unfortunately, this growth has also attracted cyber attackers to the IoMT market. To address these challenges, establishing a Secured Healthcare System (SHCS) is crucial, ensuring the privacy and security of medical data against various types of attackers. In this context, many industrialized nations collect medical data from sensors and wearable technology. FL emerges as a promising solution, enabling the collaborative development of health-related predictive algorithms while preserving the security and privacy of individual health data. FL allows healthcare participants to leverage distributed data without compromising privacy, which results in enhancing innovation in disease detection and management in the era of remote healthcare delivery. For example, Shed \textit{et al.} \cite{shen2023efficient} proposed an effective and secure online diagnostic method for e-healthcare systems based on Federated Learning Method (FLM). Initially, FLM is utilized to transform the data-sharing challenge faced by data owners into a ML problem. This approach preserves the security of training datasets by providing computed local model updates instead of raw data. Subsequently, patient medical information is effectively categorized without compromising security, employing a homomorphic cryptosystem in conjunction with the SVM method. Additionally, they introduced a novel technique to retrieve the decision function of the SVM model that effectively prevents leakage of model variables. This comprehensive approach ensures both the effectiveness of diagnostic procedures and the confidentiality of sensitive medical data in e-healthcare systems.

The authors in \cite{hossein2021bchealth} introduced BCHealth, an architecture that establishes the transparency-access control trade-off by granting data owners the authority to define access policies for their sensitive healthcare data. BCHealth comprises two separate chains dedicated to storing access policies and data transactions. To address real-world challenges such as scalability, latency, and overhead inherent in blockchain development, they employed a clustering approach. This technical strategy helps optimizing system performance and efficiency, ensuring that BCHealth can effectively manage and safeguard healthcare data while balancing transparency and access control requirements.
Sakiba \textit{et al.} \cite{sakib2021asynchronous} considered arrhythmia detection using ECG analysis a crucial application for heart activity monitoring. Utilizing the private ECG data collected within each smart logic-in-sensor deployed at the ultra edge nodes, they studied two FL architectures for categorizing arrhythmias. Their proposed paradigm enables privacy protection while also enabling online knowledge exchange through lightweight, localized, and distributed learning. To further tailor their FL-based architecture for ECG analysis, the parameters of a CNN AI model are asynchronously updated to save critical communication bandwidth.
The authors in \cite{lim2020dynamic} employed FL to enable scattered IoT users to train collaborative models at the network's edge while protecting user privacy. However, the FL network's members may differ in their willingness to participate (WTP), a secret the model owner is unaware of. Additionally, creating healthcare apps necessitates frequent long-term user involvement, for example, for the ongoing data-gathering process during which a user's WTP may alter over time. In order to explore a two-period incentive mechanism that fulfills intertemporal incentive compatibility and maintains the contract's self-revealing mechanism throughout both periods, they used the dynamic contract design.

To develop a precise, efficient, and simplified DL system for pandemic detection and prediction, the authors \cite{ajagbe2024deep} conducted a systematic literature review focused on DL techniques in this domain. Anchored by four objectives, the study examined forty-five papers out of a pool of seven hundred and ninety from diverse scholarly databases. Through analysis and evaluation, the review assessed the prevalence of DL technique applications for pandemic detection and prediction.

In order to identify COVID-19 infections using medical diagnostic image analysis, the authors in \cite{zhang2021dynamic} investigated a dynamic fusion-based FL technique tailored for medical diagnostic images. Initially, they designed an architecture specifically catered to dynamic fusion-based FL systems for assessing medical diagnostic pictures. Furthermore, they introduced a dynamic fusion approach that dynamically selects participating clients based on their local model performance and plans model fusion according to the training duration of these clients. Additionally, they offered insights into a collection of medical diagnostic image datasets capable of detecting COVID-19, that is utilized by the ML community for image analysis. This comprehensive approach not only addresses the unique challenges of medical image analysis but also contributes to the ongoing efforts in COVID-19 detection leveraging ML technologies. In order to enable privacy-enhanced COVID-19 detection with generative adversarial networks (GANs) on edge cloud computing, the authors in \cite{nguyen2021federated} introduced a FL technique called FedGAN, which combines the concepts of Generative Adversarial Networks (GANs) with FL for COVID-19 data analytics. Specifically, they devised a GAN architecture where both a discriminator and a generator, based on CNNs, are trained alternately at each edge-based medical institution to match the genuine COVID-19 data distribution. To enhance the global GAN model's ability to generate realistic COVID-19 images without sharing actual data, they proposed an FL approach. This approach enables local GANs to collaborate and exchange learned parameters with a cloud server. Moreover, they implemented DP solutions at each hospital institution to enhance privacy in federated COVID-19 data analytics. Additionally, they suggested a new FedGAN architecture integrated with blockchain technology for secure COVID-19 data analytics. This architecture decentralizes the FL process and utilizes a mining technique to minimize running latency, ensuring safe and efficient data analytics while preserving privacy and security. 

The authors in \cite{fourati2021federated} highlighted the pivotal role of FL systems in the IoMT battle against the COVID-19 pandemic. Initially, they delineated the architecture of the smart healthcare system, with a particular emphasis on the fog layer. They then delved into preprocessing tasks applicable at the fog layer, highlighting ML and DL tasks. Subsequently, they provided an overview of FL's application across various COVID-19 scenarios. Gupta et al. adopted an anomaly detection model based on FL that utilizes edge cloudlets to locally run AD models without sharing patient data. Their study centered on HFL, enabling aggregation at multiple levels to facilitate multiparty cooperation. In contrast to existing FL techniques limited to aggregation on a single server, their approach introduces a unique disease-based grouping technique for categorizing AD models according to specific disorders. Furthermore, they introduced a Federated Time Distributed (FEDTIMEDIS) LSTM method to train the AD model. They devised a Remote Patient Monitoring use case, demonstrating its implementation leveraging edge cloudlets and Digital Twin (DT) technology. This comprehensive approach highlights the potential of FL systems in bolstering healthcare management and disease detection amidst the COVID-19 crisis, while ensuring privacy and scalability. The authors in \cite{alzubi2022cloud} introduced a novel approach to safeguarding EHR privacy by integrating DL and blockchain technology. Initially, the CNN method was employed to distinguish between normal and abnormal users within the processed dataset. Subsequently, by integrating blockchain with a cryptography-based FL module, anomalous users are detected and removed from the database, thereby restricting their access to health records. This comprehensive method ensures the protection of sensitive medical information while leveraging advanced technologies to identify and mitigate potential privacy breaches in EHR systems.
In \cite{singh2022dew}, HFL is deployed by a Dew-Cloud-based model. Dew-Cloud enhances IoMT essential application availability and provides a higher level of data privacy. The hierarchical long-term memory concept is implemented on distributed Dew servers with cloud computing as the backend.\\
In the context of 6G and IoMT, the authors in \cite{kalapaaking2022smpc} explored a CNN-based FL architecture that incorporates secure multi-party computation-based (SMPC) aggregation alongside encrypted inference techniques. Their approach accounts for various hospitals equipped with a combination of IoMT devices and edge computing clusters, where locally developed models are encrypted. Subsequently, each hospital transmits its encrypted local models to the cloud for SMPC-based encrypted aggregation, leading to the construction of the encrypted global model. Upon completion, each edge server receives the encrypted global model and conducts additional localized training to enhance model accuracy further. In addition, hospitals have the flexibility to leverage either cloud computing resources or their ESs to perform encrypted inference, ensuring the privacy of their data and models throughout the process. This comprehensive architecture provides a secure and privacy-preserving framework for collaborative model training and inference across distributed healthcare environments. 

In a tele-biomedical laboratory setup, blood tests are conducted either by patients themselves in the comfort of their homes or by biomedical technicians located at satellite clinical centers. These tests utilize IoT biomedical devices that are interconnected with Hospital Edge or Cloud systems. Through this connectivity, the test results are transmitted to physicians working at federated hospitals for validation and consultation. This pipeline facilitates efficient remote healthcare delivery that allows for timely diagnosis and treatment planning while ensuring patient convenience and accessibility to medical services. For example, the authors in \cite{romano2021towards} presented a comprehensive picture of the state of the art in the tele-biomedical laboratory while noting existing problems and upcoming difficulties. In particular, the authors urged tele-biomedical laboratories to utilize IoT, edge, and cloud technologies. They first classified the primary biomedical equipment and then described many potential tele-biomedical laboratory situations. The future directions, open areas, and accuracy of references in FL-based secured healthcare systems are listed in Table \ref{FLSHCS} with details.

\begin{table*}[t]
\footnotesize
\caption{References on FL-based Secured Healthcare Systems}
\label{FLSHCS}
\setlength{\aboverulesep}{0.05cm}
\setlength{\belowrulesep}{0.05cm}
\renewcommand{\arraystretch}{1.2}
\begin{tabularx}{\textwidth}{cccc>{\raggedright\arraybackslash}m{11.6cm}}
\specialrule{0.12em}{0pt}{0pt}
Reference& Year&Acc$>90\%$&AI/ML approach & \multicolumn{1}{c}{Open Areas and Future Challenges and Directions}\\\specialrule{0.12em}{0pt}{0pt}
\cite{shen2023efficient}& 2023 &\cmark & SVM& Assessing their online diagnostic e-healthcare scheme's resilience for Byzantine attack or lower-quality local gradients. 
\\\specialrule{0.0005em}{0pt}{0pt}
\cite{sakib2021asynchronous}& 2021   &\cmark &CNN & \textbf{---} 
\\\specialrule{0.0005em}{0pt}{0pt}
\cite{ajagbe2024deep}& 2024   &\xmark &DL & \textbf{Spotting deficiencies in current literature and employing DL methods for pandemic detection and forecasting.} 
\\\specialrule{0.0005em}{0pt}{0pt}
\cite{lim2020dynamic} & 2021   & \textbf{---}&\textbf{---} & Putting a training-based reward system on top of the dynamic contract, such as DRL. \\\specialrule{0.0005em}{0pt}{0pt}
\cite{nguyen2021federated}&2022 &\cmark &FedGAN/CNN & Federated human activity analytics by expanding the FL-blockchain model, where wearable devices with sensors cooperate together to develop a common human activity.\\\specialrule{0.0005em}{0pt}{0pt}
\cite{fourati2021federated} &2021 &\cmark &DL & Investigating of FL applications and examining their findings for the safety and performance of their fog-based smart healthcare system in the context of COVID-19. \\\specialrule{0.0005em}{0pt}{0pt}
\cite{gupta2021hierarchical}& 2021 &\cmark  & FEDTIMEDIS&  Gathering of data, complete use of the Federated Time Distributed (FEDTIMEDIS) Long Short-Term Memory (LSTM) technique and an assessment of its effectiveness. \\\specialrule{0.0005em}{0pt}{0pt}
\cite{alzubi2022cloud}& 2023 & \cmark &CNN/DL &  \textbf{---} \\\specialrule{0.0005em}{0pt}{0pt}
\cite{singh2022dew} & 2022& \cmark & hierarchical FL& The Dew-Cloud-based system may be expanded to incorporate the Gurobi optimization in order to decrease latency and improve accuracy. \\\specialrule{0.0005em}{0pt}{0pt}
\cite{kalapaaking2022smpc}& 2022 & \cmark &CNN/SMPC & Creating an encrypted inference technique based on SMPC that is more lightweight and can be executed on edge servers.\\\specialrule{0.0005em}{0pt}{0pt}
\cite{romano2021towards} & 2021 &\textbf{---}  &\textbf{---} & Arousing interest in the development of the tele-biomedical laboratory among the industrial and scientific groups. \\\specialrule{0.12em}{0pt}{0pt}
\end{tabularx}
\end{table*}

\subsection{FL in Smart Cities and Homes}
FL enhances smart cities by optimizing urban services like traffic management and energy efficiency while preserving data privacy. In smart homes, FL enables personalized automation without compromising user privacy and improves comfort and energy efficiency. FL improves collaborative ML across interconnected devices that shapes the future of efficient urban living and intelligent home automation systems. For example, in smart cities, where energy efficiency and data protection are considered top priorities, deploying a large number of sensors and data-gathering devices helps FL find many uses, including traffic management, public safety, and environment monitoring, energy optimization, urban planning, and air quality monitoring. The authors in \cite{imteaj2019distributed} listed the sensors that can be found in end-users' devices. These sensors include Gyroscope, ambient light sensor, temperature, magnetic field sensor, orientation sensor, game rotation vector, linear acceleration, relative humidity, gravity, and geomagnetic rotation vector. Given that the sensors are already built into the phones, employing them is advantageous when taking into account the complexity, effectiveness, and cost of the entire system. Designing a system that causes dispersed devices to self-activate and agree to generate all available sensor data is challenging. Additionally, since devices provide a constant stream of data, the size of the data increases. Therefore, it makes it difficult to distinguish one device's sensor data from another. To address this problem, the authors offered a distributed sensing solution to utilize a token to identify a device, activate dispersed end-user devices to send data, and store it in the cloud as needed, while retaining the appropriate format. With this method, remote data collection is made possible using existing end-user devices, and the cost of adding new sensors for autonomous IoT applications decreases. In order to provide dispersed intelligence across a network of smart devices, they expanded upon their effective sensing platform. They used the processing power of these devices for local decision-making, i.e., each smart device only interacts with a small number of nearby devices rather than broadcasting all sensing data to a centralized agent, which resolves a large-scale decision-making issue.

The authors in \cite{wang2019environmental} offered a fog computing-based architecture for monitoring the environment that uses multi-source heterogeneous data gathered from IoT sensors. They used local sub-classifiers to assess the data at each edge node and then used a DNN model to combine the results from the sub-classifiers. They modeled an FL approach to simultaneously update homologous sub-classifiers at several edge nodes through model transfer. Finally, they assessed the fog computing-based architecture using multi-source and heterogeneous data gathered in Beijing. The authors in \cite{duan2019jointrec} suggested JointRec, a framework for collaborative cloud video recommendations based on DL. JointRec enables federated training across dispersed cloud servers by integrating the Joint Cloud architecture into mobile IoT. They first designed a dual-convolutional probabilistic matrix factorization (Dual-CPMF) model to undertake video recommendation. By utilizing user profiles and the descriptions of the movies that users evaluated, each cloud proposes videos based on this model, offering more accurate video recommendation services. Next, they provided a federated recommendation approach that enables each cloud to pool its weights and jointly train a model. Eventually, to decrease uplink communication costs and network capacity, they combined the 8-bit quantization approach with low-rank matrix factorization to address the high communication costs associated with federated training. 

The FL paradigm technique was described in \cite{nguyen2021spatially} for training air pollution prediction models using environmental monitoring sensor data. In the study, the authors provided a distributed learning framework to support group training among participants from various geographic locations, such as cities and prefectures. Convolutional Recurrent Neural Networks (CRNN) are trained locally in each area to forecast the local Oxidant alert level. At the same time, an aggregated global model improves the extracted information from every part of a region. Their study shows that while CRNN's intended common elements may be fused worldwide, its predictive part's adaptive structure could capture various environmental monitoring stations' configurations in localized places. In order to enhance the accuracy of the whole FL system, certain experiment findings also pointed to strategies for maintaining the balance between local DNN training epochs and synchronous training rounds for FL.
The demand for new IoT and smart city applications is surging and these applications encompass a wide array of services and apps that manage real-time data transmission and analytics, handling vast amounts of data. To meet these demands, effective computer infrastructures are essential, enabling processing and analysis of the data generated by these interconnected devices. One AI-aided tool is using an autoencoder (AE) that stores large volumes of data by compressing input data into a lower-dimensional representation, preserving essential features while reducing storage space. This compressed representation requires less storage while retaining the ability to reconstruct the original data accurately. Therefore, autoencoders offer an efficient means of storing vast amounts of data compactly without compromising vital information \cite{akhtarshenas2022open,akhtarshenas2024CNN}. Another applicable solution for addressing the large amount of data is EC which makes this situation possible by reducing network saturation and service latency. Placing multiple smaller data centers close to the data sources facilitates this computing paradigm. The management of federated edge data centers benefit from using microgrid energy sources parameterized by user needs. Energy efficiency is a major problem in executing this scenario. For example, based on the application's required data traffic, the authors in \cite{perez2019predictive} provided an ANN predictive power model for GPU-based federated edge data centers. They verified their methodology by generating 1-hour-ahead power forecasts with a normalized root-mean-square deviation less than $7.4\%$ compared to actual measurements utilizing real traffic for a cutting-edge driving assistance application. The authors in \cite{abbasi2022flitc} offered the FL IoT Traffic Classifier (FLITC), an IoT traffic classification method based on the MultiLayer Perception (MLP) neural network that keeps local data on IoT devices intact by sending only the learned parameters to the aggregation server and ensuring the privacy of traffic data.
The authors in \cite{zhao2020privacy} devised an FL system incorporating a reputation mechanism to aid home appliance manufacturers in training ML models based on consumer data, facilitating the development of smart home systems. The system operates in two stages: initially, users train the manufacturer's base model using a mobile device and a MEC server, i.e. They collect data from household equipment and train the model locally. Subsequently, users sign and transmit their models to a blockchain, replacing the centralized aggregator in traditional FL systems, thus ensuring transparency and accountability. In the second phase, manufacturers select specific individuals or groups as miners to compute the averaged model. DP was imposed on the retrieved characteristics, and a normalization method was introduced to safeguard customer privacy while enhancing test accuracy. This framework enables manufacturers to anticipate consumer demands and consumption patterns while maintaining data privacy and integrity.

\begin{table*}[t]
\footnotesize
\caption{References on FL-based Smart Architectures}
\label{FLSCS}
\setlength{\aboverulesep}{0.05cm}
\setlength{\belowrulesep}{0.05cm}
\renewcommand{\arraystretch}{1.2}
\begin{tabularx}{\textwidth}{cccc>{\raggedright\arraybackslash}m{11.7cm}}
\specialrule{0.12em}{0pt}{0pt}
Reference& Year&Acc$>90\%$&AI/ML approach & \multicolumn{1}{c}{Open Areas and Future Challenges and Directions}\\\specialrule{0.12em}{0pt}{0pt}
\cite{imteaj2019distributed} & 2019 &\textbf{---} &\textbf{---} & Creating decentralized learning methods specifically for use cases with the autonomous internet of things. \\\specialrule{0.0005em}{0pt}{0pt}
\cite{wang2019environmental} & 2019   &\cmark &\textbf{---} & Enhancing the distributed model average method's efficiency and optimizing the consideration of variables like data dispersion and volume. \\\specialrule{0.0005em}{0pt}{0pt}
\cite{duan2019jointrec} &2020 &\xmark & Dual-CPMF& Examining a central video recommendation method that was coordinated by the edge-end-cloud, where the recommender is installed on the edge server close to the clients. \\\specialrule{0.0005em}{0pt}{0pt}
\cite{nguyen2021spatially}   &2021 &\cmark &CRNN & Personalizing clients' feedback models over a cloud server to keep local models updated with the most pertinent geographical data through neighboring update aggregation.\\\specialrule{0.0005em}{0pt}{0pt}
\cite{perez2019predictive} & 2019 &\textbf{---} & ANN& Using several frequency ranges in the GPUs for clock frequencies and memory. \\\specialrule{0.0005em}{0pt}{0pt}
\cite{abbasi2022flitc} & 2022 & \textbf{---}&FLITC & Enhancing the FLITC (FL-IoT-Traffic Classifier) by utilizing EC systems' functionalities.\\\specialrule{0.0005em}{0pt}{0pt}
\cite{zhao2020privacy} & 2021 &\cmark & \textbf{---}& Testing their block-chain method using actual datasets, and determining the ideal ratio for global and local epochs. \\\specialrule{0.12em}{0pt}{0pt}
\end{tabularx}
\end{table*}
Liu \textit{et al.} \cite{liu2022intrusion} focused on the features of IoT-based Maritime Transportation Systems (MTS) and introduced FedBatch, a CNN-MLP-based intrusion detection model trained using FL. FL preserves the confidentiality of local data on boats by conducting model training locally and updating the global model solely through the sharing of model parameters. They initially tackled the challenges of communication among multiple boats to simulate the FL process at sea. Subsequently, they developed a lightweight local model composed of CNN and MLP to minimize processing and storage overhead. Additionally, they introduced Batch Federated Aggregation, an adaptive aggregation technique designed to mitigate the oscillations of model parameters during FL, thereby addressing the straggler issue encountered in MTS. The future directions, open areas, and accuracy of references in FL-based smart architectures are listed in Table \ref{FLSCS} with details.

\section{Scalability} \label{sec:scale}
In ultra-dense networks spanning vast geographic areas with numerous clients, scalability is paramount for addressing model updating and control challenges effectively. To tackle these concerns, distributed systems employ combiners and reducers strategically distributed across the area of interest. Combiners handle partial aggregation and load balancing tasks, ensuring efficient distribution of computational load. Meanwhile, reducers manage combiner connections and oversee global model generation and refinement processes, facilitating effective coordination and enhancement of the overall model. This decentralized approach enables scalability and robustness in managing model updates and aggregation tasks across expansive network environments. The following control mechanism is applied to monitor and provide service discovery among all clients \cite{ekmefjord2022scalable}.
\begin{table*}[t]
\footnotesize
\caption{References on Scalability of FL Structures}
\label{FLScal}
\setlength{\aboverulesep}{0.05cm}
\setlength{\belowrulesep}{0.05cm}
\renewcommand{\arraystretch}{1.2}
\begin{tabularx}{\textwidth}{cccc>{\raggedright\arraybackslash}m{11.7cm}}
\specialrule{0.12em}{0pt}{0pt}
Reference& Year&Acc$>90\%$&AI/ML approach & \multicolumn{1}{c}{Open Areas and Future Challenges and Directions}\\\specialrule{0.12em}{0pt}{0pt}
\cite{ekmefjord2022scalable}& 2021 & \xmark& FEDn & Improving the reducer that spends a large amount of time downloading and loading models from the combiners. \\\specialrule{0.0005em}{0pt}{0pt}
\cite{diaz2023study} & 2023   &\xmark & \textbf{\textbf{---}} & Exploring the advantages of their medical image classifier compared with the centralized classifier. \\\specialrule{0.0005em}{0pt}{0pt}
\cite{zhang2021faithful}&2021 & \cmark& Faithful-FL(FFL)& Exploring the effect of ignoring processing and transmission cost.  \\\specialrule{0.0005em}{0pt}{0pt}
\cite{soltan2023scalable}  &2023 &\xmark & \textbf{---} & Implementing a fully-autonomous data extraction pathway through direct Electronic Healthcare Records. \\\specialrule{0.0005em}{0pt}{0pt}
\cite{de2023towards}& 2023 &\xmark & FDFL& Optimizing efficiency, security, and privacy by investigating the impact of many aggregator levels in fully decentralized FL (FDFL). \\\specialrule{0.12em}{0pt}{0pt}
\end{tabularx}
\end{table*}
An additional critical aspect of scalability involves the dynamic participation of clients within the network, significantly impacting both local and global model aggregation processes. In \cite{diaz2023study}, the authors investigates this issue through a case study involving a large number of users. They focused on analyzing medical images, particularly chest X-ray datasets, to address challenges related to accuracy, loss, complexity, time, and privacy in FL. The primary objective of their research is to evaluate the scalability of FL, examining two significant case studies. One scenario explores the involvement of intermittent clients who may occasionally join or exit the training process while still contributing to the FL framework. Another case study considers medical centers that stop their participation in the FL process, either by discontinuing the sharing of medical records or requiring enhanced facilitation to sustain collaboration effectively.
To further discuss the scalability of detection models, the authors in \cite{han2023lmca} introduced LMCA, an approach named Lightweight Model Combining Adjusted MobileNet and Coordinate Attention mechanism, for detecting anomalies in network traffic. By integrating the adjusted MobileNet model with the coordinate attention mechanism, they devised a streamlined anomaly detection model capable of efficiently capturing local, global, and spatial-temporal characteristics of traffic data. Suitable for deployment on IoT devices due to its compact size and high performance, this model offers a promising solution for anomaly detection in network environments. Additionally, they proposed a method for extracting traffic features that reduces redundancy and expedite neural network convergence. Their research addresses the challenge of optimizing the MobileNet model for small datasets, thereby expanding the applicability of anomaly detection techniques to IoT environments.
Zheng \textit{et al.} \cite{zhang2021faithful} explored the privacy and scalability enhancements for their system model, devising an economically viable strategy. To address privacy concerns, they introduced a Faithful Federated Learning (FFL) framework. This framework ensures that deviations, fixation on relevant information, message passing, and computation do not favor any specific client. FFL focuses on estimating the VCG payments, thereby ensuring faithful implementation, voluntary contribution, and optimal conditions while controlling model complexity. Subsequently, they grouped clients into clusters to approximate the scalability of the VCG structure. Finally, they developed a scalable and Differentially Private FFL scheme, enabling clients to trade off performance among three essential factors: iteration demand, payment precision loss, and privacy protection. This approach offers a comprehensive solution for addressing both privacy and scalability concerns in FL systems. COVID-19 screening check-up was federated and extended in \cite{soltan2023scalable}, where the authors evaluated and developed a quickly client-friendly and scalable FL idea applied to hospital teams in the UK. They established an E2E solution that eliminates the need for patients' data transformation. This solution involves using and analyzing medical records, including vital signs and blood tests, which may otherwise take a considerable amount of time to reach hospitals. Fully decentralized FL (FDFL) was introduced in \cite{de2023towards} based on a peer-to-peer network in order to develop the resilience and scalability of standard FL while guaranteeing proper convergence speed using global gradient methods. To this end, an aggregator-based model was designed that provides a client election process and scalability properties concerning network size cache, computation, and communication to cope with client failures caused by decentralized systems. The future directions, open areas, and accuracy of references in the scalability of FL-based structures are listed in Table \ref{FLScal} with details.

\section{Conclusion}\label{sec:conc}
As data volumes surge and privacy concerns intensify, along with the advent of large machine learning (ML) models, federated learning (FL) stands out as an opportune and significant solution. FL facilitates collaborative model training across the decentralized devices, leading to the development of robust ML models that leverage extensive data for training while simultaneously ensuring that user privacy is preserved. In this survey, we provided a comprehensive definition of FL and elucidated its distinctions from centralized and decentralized learning paradigms. Additionally, we conducted a comprehensive review of the most recent FL algorithms, systematically evaluating and categorizing them based on diverse criteria, including mathematical frameworks, privacy preservation techniques, resource allocation methodologies, and practical applications. After thoroughly reviewing each method, we summarized the recent literature and extracted the remaining gaps and open areas. The research directions were systematically tabulated separately according to different areas of research interests, including blockchain structures, privacy-preserving frameworks, resource allocation, and applications like aerial and non-terrestrial networks, healthcare systems, and smart cities. This survey provides researchers with a clear view of the existing limitations in the FL field for further investigation. We believe that the dynamic nature of ML calls for ongoing research to address new challenges and improve FL algorithms for future applications.
\section*{Conflict of Interest}\label{sec8}
On behalf of all authors, the corresponding author states that there is no conflict of interest.
\section*{Acknowledgement}\label{sec8}
This research is supported by the Generalitat Valenciana through the CIDEGENT PlaGenT, Grant CIDEXG/2022/17, Project iTENTE, and the action CNS2023-144333, financed by MCIN/AEI/10.13039/501100011033 and the European Union “NextGenerationEU”/PRTR.
\bibliographystyle{elsarticle-num}
\bibliography{rfs}





\end{document}